%% file: acl_latex.tex
\pdfoutput=1

\documentclass[11pt]{article}

\usepackage[final]{acl}

\usepackage{times}
\usepackage{latexsym}

\usepackage[T1]{fontenc}

\usepackage[utf8]{inputenc}

\usepackage{microtype}

\usepackage{inconsolata}

\usepackage{graphicx}

\input{dfns}
\input{math_commands}

\title{Can Language Models Reason about \\ \indieemoji Individualistic Human Values and Preferences?}

\author{
    \textbf{Liwei Jiang$^{\spadesuit}$\thanks{Work done while at Allen Institute for Artificial Intelligence.}} \quad
    \textbf{Taylor Sorensen$^{\spadesuit}$} \quad 
    \textbf{Sydney Levine$^{\clubsuit*}$} \quad
    \textbf{Yejin Choi$^{\heartsuit}$} \\
    \vspace{-0.5em} \\
    \textnormal{
    University of Washington$^{\spadesuit}$ \quad Google DeepMind$^{\clubsuit}$ \quad Stanford University$^{\heartsuit}$} \\
    \vspace{-0.5em} \\
    \textnormal{ \texttt{lwjiang@cs.washington.edu}} \\ 
    \vspace{1.5em}
    \textnormal{ \github \textbf{Code}: \url{https://github.com/liweijiang/indievalue}
    }
}

\begin{document}
\maketitle
\input{sections/0_abstract}
\input{sections/1_introduction}
\input{sections/2_data}
\input{sections/3_probe_results}

\input{sections/4_finetune}
\input{sections/5_discussion}

\input{sections/6_related_works}

\input{sections/7_conclusion}
\input{sections/z_acknowledgement}

\clearpage
\newpage
\input{sections/z_limitations}

\clearpage
\newpage
\bibliography{custom}

\clearpage
\newpage
\input{sections/z_appendix}
\end{document}

%% file: dfns.tex
\usepackage{colortbl}
\usepackage{wrapfig}
\usepackage{graphicx}
\usepackage{xspace}
\usepackage{multirow}
\usepackage{makecell}
\usepackage{nicematrix}
\usepackage[normalem]{ulem}
\usepackage{listings}
\usepackage{comment}
\usepackage{subfigure}
\usepackage{caption}
\usepackage{rotate}
\usepackage{float}
\usepackage{booktabs} 
\usepackage{cleveref}
\usepackage{hyperref}
\usepackage{algorithmic}
\usepackage[linesnumbered,ruled,vlined]{algorithm2e}
\usepackage[bottom]{footmisc}
\usepackage{tcolorbox}
\usepackage{tabularx}
\usepackage{arydshln}

\NewDocumentCommand{\rot}{O{45} O{1em} m}{\makebox[#2][l]{\rotatebox{#1}{#3}}}%

\definecolor{babypink}{rgb}{0.96, 0.76, 0.76}
\definecolor{coralpink}{rgb}{0.97, 0.51, 0.47}
\definecolor{aqua}{rgb}{0.0, 1.0, 1.0}
\definecolor{columbiablue}{rgb}{0.61, 0.87, 1.0}
\definecolor{cyan}{RGB}{222, 255, 255}
\definecolor{turquoiseblue}{rgb}{0.0, 1.0, 0.94}
\definecolor{realcyan}{RGB}{0, 200, 200}

\newcommand*\inlineimage[1]{\raisebox{-0.15\baselineskip}{\includegraphics[height=1\baselineskip]{#1}$\,$}}
\newcommand{\indieemoji}{\inlineimage{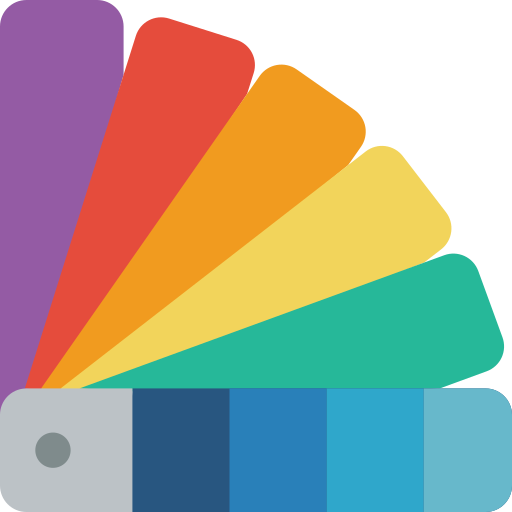}\xspace}
\newcommand{\taskemoji}{\inlineimage{figures/logos/colorful.png}\xspace}
\newcommand{\datasetemoji}{\inlineimage{figures/logos/colorful.png}\xspace}

\newcommand{\eg}{e.g.,\xspace}
\newcommand{\ie}{i.e.,\xspace}

\newcommand{\finetune}{Individualistic Value Reasoner\xspace}
\newcommand{\finetuneshort}{\textsc{IndieValueReasoner}\xspace}

\newcommand{\finetuneshorts}{\textsc{IndieValueReasoners}\xspace}

\newcommand{\llamathreeone}{Llama-3.1-8B\xspace}

\newcommand{\even}{\textsc{Value Inequity Index}\xspace}
\newcommand{\evenshort}{$\sigma$\textsc{Inequity}\xspace}

\newcommand{\evenshortmath}{\sigma \text{\textsc{Inequity}}\xspace}

\newcommand{\specialcell}[2][c]{%
  \begin{tabular}[#1]{@{}l@{}}#2\end{tabular}}

\newcommand{\taskfull}{\textsc{IndieValueCatalog}\xspace}
\newcommand{\task}{\textsc{IndieValueCatalog}\xspace}

\newcommand{\datasetfull}{\textsc{IndieValueCatalog}\xspace}
\newcommand{\dataset}{\textsc{IndieValueCatalog}\xspace}

\newcommand{\wvs}{WVS \xspace}

\newcommand{\databinary}{polar\xspace}

\newcommand{\datarefined}{refined\xspace}

\newcommand{\Databinary}{Polar\xspace}

\newcommand{\Datarefined}{Refined\xspace}

\hypersetup{
    colorlinks = true,
    linkbordercolor = {white},
    linkcolor = blue!60!black,
    citecolor = blue!60!black,
    urlcolor = blue!60!black
}

\newcommand{\github}{\raisebox{-1.5pt}{\includegraphics[height=1.05em]{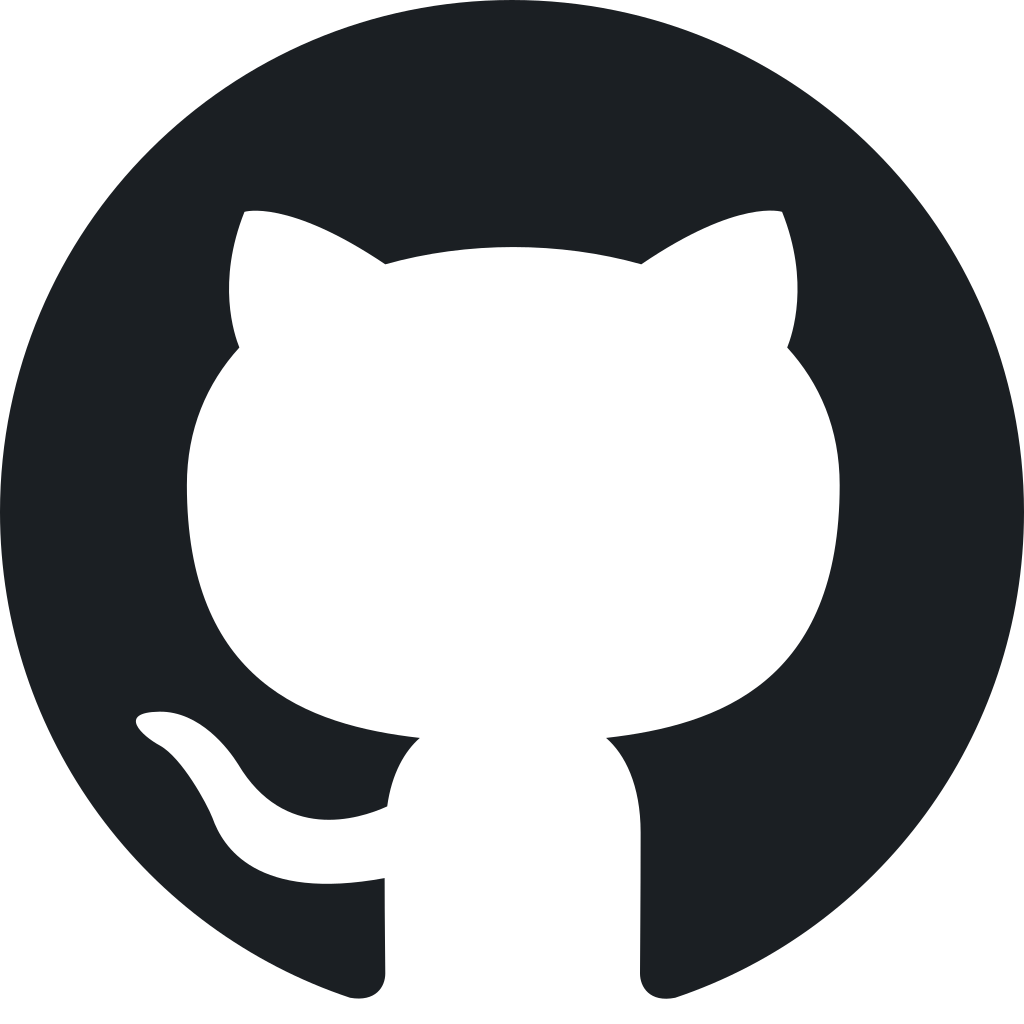}}\xspace}

%% file: math_commands.tex
\usepackage{amsmath,amsfonts,bm,bbm}

\newcommand{\indic}{\mathbbm{1}}

\def\eqref#1{equation~\ref{#1}}

\def\1{\bm{1}}

\DeclareMathAlphabet{\mathsfit}{\encodingdefault}{\sfdefault}{m}{sl}
\SetMathAlphabet{\mathsfit}{bold}{\encodingdefault}{\sfdefault}{bx}{n}



%% file: sections/0_abstract.tex
\begin{abstract}
Recent calls for pluralistic alignment emphasize that AI systems should address the diverse needs of \textit{all} people. Yet, efforts in this space often require sorting people into fixed buckets of pre-specified diversity-defining dimensions (\eg demographics), risking smoothing out individualistic variations or even stereotyping. To achieve an authentic representation of diversity that respects individuality, we propose \textit{individualistic alignment}.\footnote{We use the phrase \textit{individualistic value} to describe ``values related to one individual,'' instead of ``values about individualism, such as being independent and self-reliant.''} 
While individualistic alignment can take various forms, we introduce \indieemoji \datasetfull, a dataset transformed from the influential World Values Survey (WVS), to study language models (LMs) on the specific challenge of \textit{individualistic value reasoning}. Given a sample of an individual's value-expressing statements, models are tasked with predicting this person's value judgments in novel cases. With \dataset, we reveal critical limitations in frontier LMs, which achieve only 55 \% to  65\% accuracy in predicting individualistic values. Moreover, our results highlight that a precise description of individualistic values cannot be approximated only with demographic information. 
We also identify a partiality of LMs in reasoning about global individualistic values, as measured by our proposed \even (\evenshort). 
Finally, we train a series of \finetuneshorts to reveal new patterns and dynamics into global human values. 

\end{abstract}

%% file: sections/1_introduction.tex
\section{Introduction}
\label{sec:introduction}

\begin{figure*}[t!]
    \vspace{-0.6cm}
    \centering
    \includegraphics[width=1\textwidth]{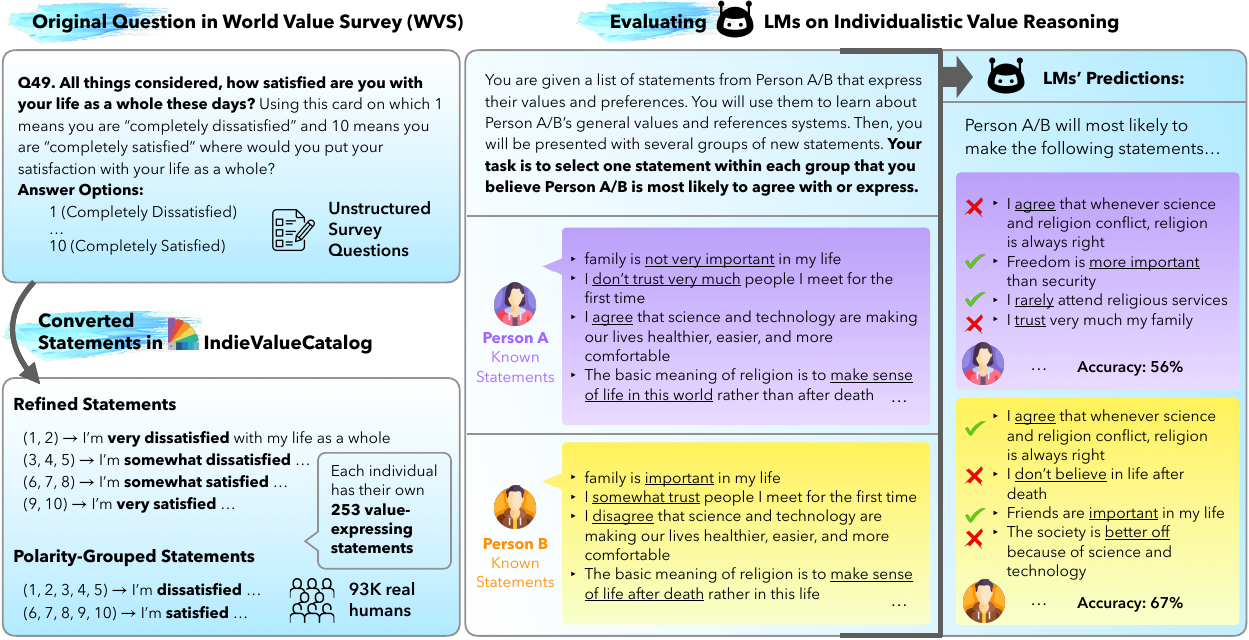}
    \vspace{-0.6cm}
    \caption{\taskemoji \dataset
    contains statements expressing individualistic human values
    from 94K real humans worldwide. With this resource, we study LMs' ability to reason about individual human values.}
    \label{fig:concept}
    \vspace{-0.5cm}
\end{figure*}

Recent advocates for pluralistic alignment underscore the importance of AI systems being geared towards the diverse perspectives and needs of \textit{all} people \citep{sorensen2024roadmappluralisticalignment}. However, existing methods and evaluation frameworks for achieving this goal face a key limitation---the diversity of people is pre-specified and coarsely categorized, papering over individuality \citep{PERSONA2024,sun2024personadb}. Pre-selected \textit{diversity-defining dimensions}, \eg demographics \citep{moon2024personabackstory,kwok2024syntheticpersonas}, personality \citep{PERSONA2024,jiang2023evaluating, zhu2024personalityalignment}, cultures \citep{chiu2024culturalbench}, writing styles \citep{han2024valueaugmentedsamplinglanguage,personalizedsoup2023}, necessitate sorting individuals into coarse buckets. These choices not only pose the risk of stereotyping \citep{kirk2024personalization}, but also inherit potentially negative biases from the specific choice of the diversity dimensions. While some evaluations exist for assessing value representations among more fine-grained demographic groups \citep{GlobalOpinionQA2024,lmopinion2023}, these efforts still rely on group-level distributional inferences, and do not capture individual variations. 

As a bottom-up alternative to addressing these challenges, we propose \textit{individualistic value alignment}, a maximal version of pluralistic alignment that models diversity at the individual level. This framework infers individual preferences from the ground up, avoiding predefined categories and thereby authentically representing diversity by honoring individuality. As a crucial step towards this goal, we propose and study \textit{individualistic value reasoning}---a task for inferring a person's general value system based on descriptive evidence of their preferences and applying this inference to predict their value preferences in new situations.

One key challenge in studying individual human values lies in the difficulty of acquiring multi-faceted data that is sufficiently representative of an individual's overall value preferences. To this end, we present \indieemoji \taskfull, a dataset specifically designed to evaluate and advance LMs' ability to reason about an individual's value preferences in novel situations. \task transforms unstructured survey questions from the influential social science study of World Value Survey (WVS) into 929 standardized natural language statements describing one's value preferences (\eg ``I don't believe in life after death''). Our data conversion results in a rich repository of value-expressing statements from 93K unique \textit{real} humans across the globe. 
Each person has, on average, 242 (max 253) value-expressing statements, along with 31 demographics-declaring statements. In sum, \task presents the first application of WVS for studying individualistic human values with LMs in a unified, configurable, and easy-to-measure schema. 

With \dataset, we first expose the lack of proficiency of frontier LMs in interpreting and predicting individualistic human values, as demonstrated by zero-shot accuracies ranging between 55\% to 65\%. We also introduce \even (\evenshort), a unified metric for assessing the degree of \textit{equity} and \textit{impartiality} of LMs on this task, revealing critical inequity of LMs in handling individualistic values across population groups. We also discover that adding demographic specifications alongside value-expressing statements has a marginal impact on improving individualistic value predictions for strong LMs. This highlights the risks of over-relying on demographic factors to define the identities and values of individuals and stresses the importance of addressing values from a granular and descriptive perspective.

Finally, we train a collection of \finetune{s} (\finetuneshort) with \dataset, achieving improved proficiency and \evenshort on the individualistic value reasoning task, as measured by held-out evaluation data. We conduct extensive experimentation involving different training configurations with \dataset, \eg the number of value-expressing demonstration statements, the granularity of these statements, and the regional origins of the training individuals. Our findings reveal novel dynamics and characteristics of global human values captured in the classical social science resource. We hope our study inspires further research into \textit{individualistic value alignment} and \textit{reasoning}, and we outline key challenges and opportunities.

%% file: sections/2_data.tex
\section{\datasetemoji \datasetfull: A Real-World Dataset for Individualistic Human Value Reasoning}
\label{sec:data}

Credible, real-world cross-cultural data that captures diverse human values and preferences is difficult to obtain at scale \citep{PERSONA2024}. The influential World Value Survey (WVS) addresses this challenge by collecting global responses on social, political, economic, religious, and cultural values \citep{wvswave72020}, and has been used to assess LMs' biases across demographic groups \citep{worldvaluesbench2024,GlobalOpinionQA2024}. However, for the first time, individual respondent data sequences of WVS are used to evaluate LMs' reasoning on individualistic values and preferences.

\subsection{Data Transformation}

\paragraph{Question Unification.} 
The original WVS is composed of questions with varying answer formats (\eg multiple-choice, Likert scale) and fragmented language descriptions. We thus standardized all questions by converting them into unified natural language statements reflecting value preferences. For instance, we morph questions (\eg WVS Q131: ``Could you tell me how secure you feel these days?'') and answers (\eg 1. ``very secure,'' 2. ``quite secure'' ...) into sets of statements like ``I feel very secure these days.'' Figure \ref{fig:concept} and Table \ref{tab:example_converted_statements} show example converted statements in two distinct granularity forms, \ie polarity-grouped (\textit{\databinary}) and \textit{\datarefined} statements. Demographic questions (31 in total) were similarly converted into identity-declaring statements (\eg ``I'm an immigrant to this country''). See Table \ref{tab:demographics_1}-\ref{tab:demographics_5} for demographics questions and Appendix \S \ref{asec:dataset_details} for full details.

\paragraph{Dataset statistics.} 253 original questions are converted to 929 and 567 possible statements for \textit{\datarefined} and \textit{\databinary} setups, respectively, across 93K  survey respondents from 70+ countries. For each WVS question, one statement is chosen by each person (unless a question was chosen to be omitted). The combinatorial answer space for all 253 questions is extremely large: the \textit{\datarefined} setup has $1.65 \times 10^{139}$ and the \textit{\databinary} setup has $3.94 \times 10^{86}$ answer combinations, making predicting the exact value choices of a person highly difficult. Table \ref{tab:dataset_stats} in Appendix \S \ref{asec:dataset_details} shows the dataset statistics.

\input{tables/dataset_statistics}

\subsection{Evaluation Setups}

\begin{figure*}[t!]
    \centering
    \begin{minipage}[b]{0.76\textwidth}
        \centering
        \includegraphics[width=1\textwidth]{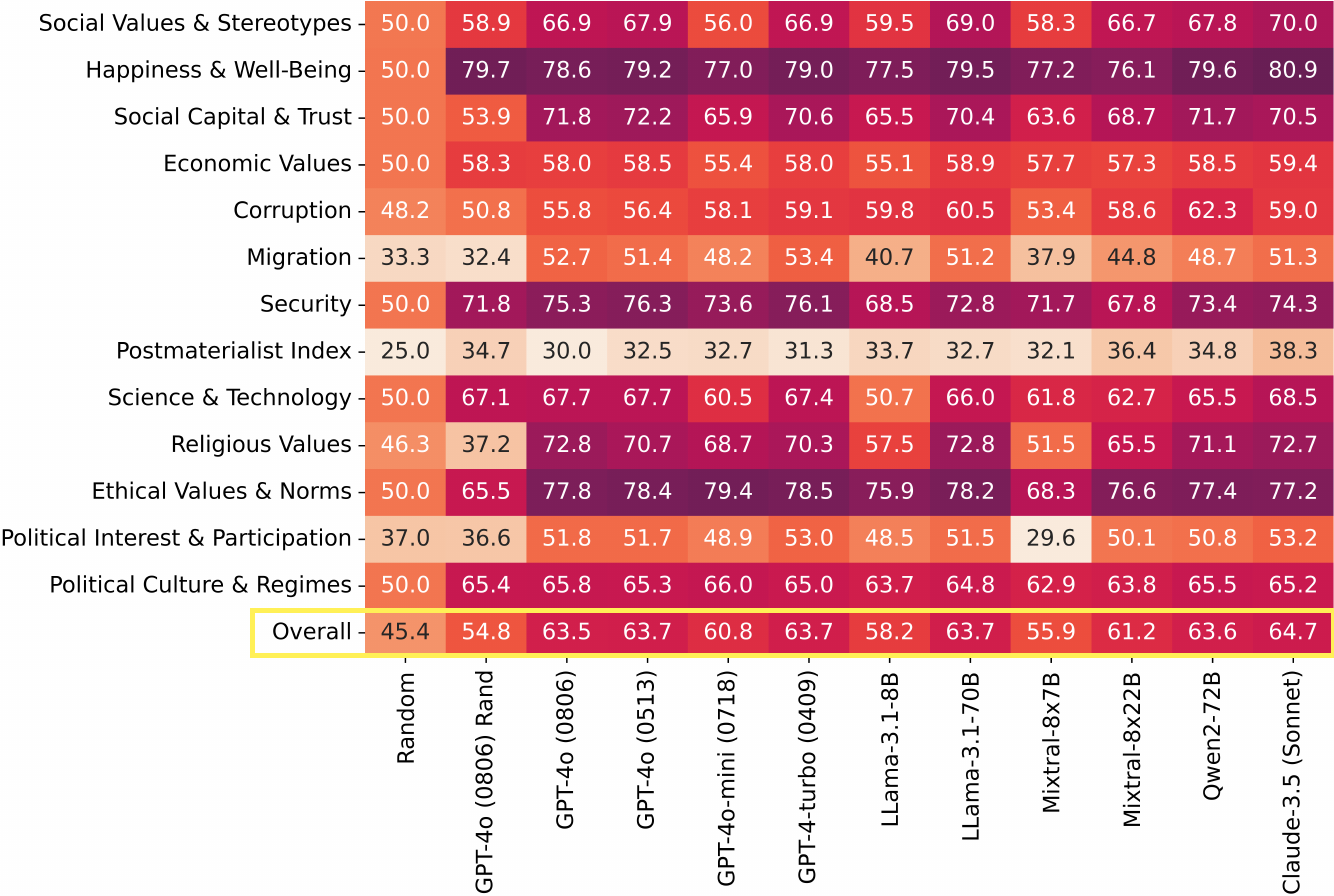}
        \vspace{-0.8cm}
        \captionof{figure}{Evaluation of LMs' individualistic human value reasoning capability using \dataset. 
        \texttt{Random} 
        randomly chooses a statement candidate. \texttt{GPT-4o (0806) Rand} lets GPT-4o randomly guess a statement without demonstration statements.}
        \label{fig:model_comparisons}
    \end{minipage}
    \hfill
    \begin{minipage}[b]{0.2\textwidth}
        \centering
        \small
        
        \begin{tabular}{l@{\hspace{0.2\tabcolsep}}
        r@{\hspace{0.6\tabcolsep}}}
        \toprule
        
        \scriptsize{\textbf{Model}} & \hspace{-0.95cm} \scriptsize{\textbf{\evenshort} $\downarrow$} \\
        \midrule

        \texttt{\scriptsize{GPT-4o(0806)}}       & 3.03 \\
        \texttt{\scriptsize{GPT-4o(0513)}}       & 2.87 \\
        \texttt{\scriptsize{GPT-4o-mini(0718)}}  & 2.55 \\
        \texttt{\scriptsize{GPT-4-turbo(0409)}}  & 2.83 \\
        \texttt{\scriptsize{LLama-3.1-8B}}        & 2.97 \\
        \texttt{\scriptsize{LLama-3.1-70B}}       & \textbf{1.94} \\
        \texttt{\scriptsize{Mixtral-8x7B}}        & 3.19 \\
        \texttt{\scriptsize{Mixtral-8x22B}}       & 3.06 \\
        \texttt{\scriptsize{Qwen2-72B}}           & 3.24 \\
        \texttt{\scriptsize{Claude-3.5(Sonnet)}} & 3.14 \\
        \bottomrule
        \end{tabular}
        \captionof{table}{\evenshort, \ie \even, measures the level of \textit{partiality} or \textit{inequity} of LMs in reasoning about individualistic human values across diverse population groups averaged by 13 demographic dimensions.}
        \label{tab:equity_index_compare_models}
    \end{minipage}
    \vspace{-0.4cm}
\end{figure*}

\paragraph{Evaluation setups.}
As illustrated in Figure \ref{fig:concept}, each individual's statements are divided into a \textit{demonstration} (50 to 200 statements) and a \textit{probing} subset (39 statements across 13 WVS question categories; see Table \ref{tab:probing_question_splits} of Appendix \S \ref{assec:probing_setups} for data split details). LMs are tasked with selecting a statement most likely to align with an individual's values among an unseen \textit{probing} set according to \textit{demonstration} value statements and, optionally, demographic statements. 
We adopt a cross-validation setup with three splits of 200 demonstrations and 39 probes; reporting averaged results to prevent overfitting specific probing set choices. Finally, we sample 800 individuals from \dataset as the held-out probing and evaluation set, ensuring a balanced demographic representation. 

Formally, $\mathbb{Q}$ is the full set of 253 value-inquiring questions and $\mathbb{I}$ represents all individuals in \dataset, which is split into a held-out evaluation set with 800 individuals ($\mathbb{I}_{\text{eval}}$) and a training set ($\mathbb{I}_{\text{train}}$). Each question $q \in \mathbb{Q}$ has a set of statements $S_q$ expressing varying opinions regarding $q$. For each individual $I_i \in \mathbb{I}$, with each question $q \in \mathbb{Q}$, $I_i$ chooses one of the statements in $S_q$, \ie $s_q^{I_i} = S_q(I_i), s_q^{I_i} \in S_q$, which best represents their opinions regarding $q$.

Each probing setup, $P_j \in \{P_0, P_1, P_2\}$, splits $\mathbb{Q}$ into a \textit{probing} set of $39$ questions ($\mathbb{Q}_{P_j}^{\text{probe}}$)
and a remaining \textit{demonstration} set ($\mathbb{Q}_{P_j}^{\text{demo}}$).
For each $I_i \in \mathbb{I}_{\text{eval}}$  we sample $d$ demonstration questions, \ie $\mathbb{Q}_{P_j}^{\text{demo}}(I_i, d) \subseteq \mathbb{Q}_{P_j}^{\text{demo}}$, and gather the chosen statements of $I_i$ of these questions, \ie $\mathbb{S}_{P_j}^{\text{demo}}(I_i, d) = \{s_q^{I_i} | \forall q \in  \mathbb{Q}_{P_j}^{\text{demo}}(I_i, d)\}$. 
During probing, we present a model, $M$, with $\mathbb{S}_{P_j}^{\text{demo}}(I_i, d)$ along with statement options of all probing questions, $\mathbb{S}_{P_j}^{\text{probe}} = \{S_q | \forall q \in \mathbb{Q}_{P_j}^{\text{probe}}\}$. Finally, we conclude $M$'s choice of value statements for $I_i$ by sampling with \verb|temperature=0|\footnote{We aim to elicit deterministic predictions from the model to avoid randomness in the evaluation resulting from temperature-scaled sampling.} and \verb|top_p=1|, \ie $\{\hat{s}_{M,q}^{I_i} \sim M(S_q | \mathbb{S}_{P_j}^{\text{demo}}(I_i, d)) | \forall q \in \mathbb{Q}_{P_j}^{\text{probe}}\}$.

\paragraph{Measuring \textit{proficiency}.}
The average accuracy of $M$ for each individual across three probing setups and the overall accuracy are calculated as follows.

\vspace{-0.3cm}
{\small
\begin{align}
Acc_M^{I_i} = \frac{1}{3 \times |\mathbb{Q}_{P_j}^{\text{probe}}|} \sum_{P_j \in \{P_0, P_1, P_2\}} \sum_{q \in \mathbb{Q}_{P_j}^{\text{probe}}} \indic \big[ \hat{s}_{M,q}^{I_i} = s_q^{I_i} \big] 
\end{align}
}
\vspace{-0.8cm}
{\small
\begin{align}
\tiny
Acc_M = \frac{1}{|\mathbb{I}_{\text{eval}}|} \sum_{I_i \in \mathbb{I}_{\text{eval}}} Acc_M^{I_i}
\end{align}
}
\vspace{-0.5cm}

\paragraph{Measuring \textit{impartiality} and \textit{equity}.}
We introduce \even (\evenshort), a metric for measuring the \textit{impartiality} or \textit{equity} level of a LM in individualistic value reasoning. In essence, we measure how much performance \textit{variance} a LM shows in reasoning about individuals across demographic groups---a lower variance means more impartial understanding. We consider 13 demographic dimensions ($\mathcal{D}^k \in \mathbb{D}$, \eg income level; see Appendix \S \ref{assec:probing_setups} for details). 
Each demographic dimension is broken into numbers of groups, $g_{k_t} \in \mathcal{D}^k$, \eg low/middle/high for $\mathcal{D}^{\text{income level}}$. Every individual belongs to one of the groups for each demographic dimension, \ie $\mathcal{D}^k(I_i) = g_{k_t}^{I_i}$. We denote all individuals belong to $g_{k_t}$ as $\mathbb{I}_{\text{evel}}^{g_{k_t}} = \{I_i \; | \; \forall I_i \in \mathbb{I}_{\text{evel}}, \mathcal{D}^k(I_i) = g_{k_t}\}$. We define \evenshort of a LM, $M$, as follows.

\vspace{-0.5cm}
{\small
\begin{align}
\evenshortmath_M = \frac{1}{|\mathbb{D}|} \sum_{\mathcal{D}^k \in \mathbb{D}} \sigma( \{Acc_M^{\mathbb{I}_{\text{evel}}^{g_{k_t}}} \; | \; \forall g_{k_t} \in \mathcal{D}^k\} )
\end{align}
}
\vspace{-0.5cm}

where $Acc_M^{\mathbb{I}_{\text{probe}}^{g_{k_t}}}$ is the accuracy among population of the $g_{k_t}$ demographic group for model $M$. $\sigma$ denotes standard deviation. \textbf{The lower $\evenshortmath_M$, the more impartial $M$ is.}

%% file: tables/dataset_statistics.tex
\begin{table}[t]
\vspace{-0.4cm}
\centering
\scriptsize

\begin{tabularx}{7.5cm}{c|c|c|c}
\toprule
\multicolumn{4}{l}{\textsc{\textbf{Data Conversion}}} \\
\midrule

\texttt{\#Questions (Q)} & \texttt{\#Refined Stmt} & \texttt{\#Polar Stmt} & \texttt{\#Person (P)} \\
 253 & 929 & 567 & 93,279 \\ 
\end{tabularx}

\begin{tabularx}{7.5cm}{c|c|c}
\toprule
\multicolumn{3}{l}{\textsc{\textbf{Data with Valid Labels}}} \\
\midrule

\texttt{Total \#Valid Q} & \texttt{\#Valid Q / P} & \texttt{\#P w/ Full Q Set} \\
22.6M & 242.03 ($\sigma=$17.31) & 15,819 \\ 
\bottomrule
\end{tabularx}
\vspace{-0.3cm}
\caption{Statistics of \dataset.}
\label{tab:dataset_stats}
\vspace{-0.6cm}
\end{table}

%% file: sections/3_probe_results.tex
\section{Can LMs Reason about Individualistic Human Values and Preferences?}
\label{sec:probe_results}

\paragraph{\textit{Proficiency} in individualistic value reasoning.}
Figure \ref{fig:model_comparisons} shows that all models outperform the \texttt{Random} baseline, where a statement is chosen randomly. The \texttt{GPT-4o (0806) Rand} baseline, in which GPT-4o is given no demonstration, achieves higher accuracy than \texttt{Random}, suggesting that GPT-4o has systematic preferences over statements, allowing it to align with broader preferences without demonstrations. Notably, GPT-4o with 200 demonstrations considerably outperforms the model without any (63.5 vs. 54.8), indicating that demonstrations can effectively guide LMs in interpreting an individual's general value preferences. Yet, no model achieves high performance, with average accuracies ranging between 55\% to 65\%. Lastly, certain categories of statements (\eg Happiness \& Well-being, Ethical Values \& Norms) are easier to predict than others (\eg Economic Values, Postmaterial Index). Figure \ref{fig:probe_by_subcategory} in Appendix \S \ref{assec:probing_results} shows how each value statements type influences the prediction of others.

\begin{figure}[t]
    \centering
    \includegraphics[width=0.4\textwidth]{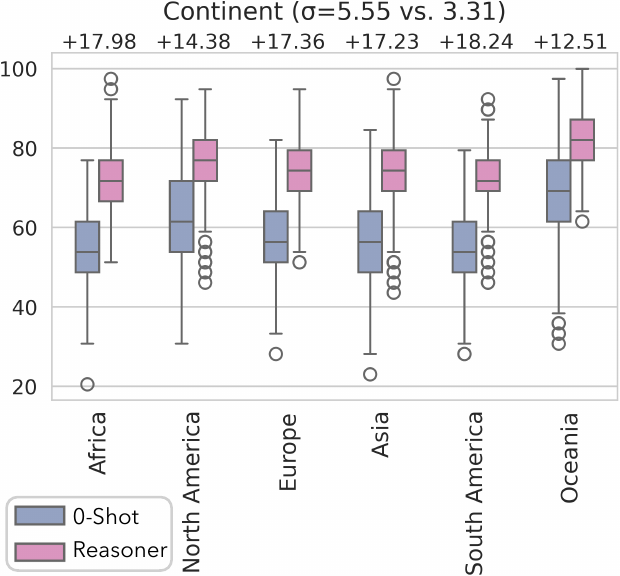}
    \vspace{-0.2cm}
    \captionof{figure}{Comparing \llamathreeone zero-shot vs. \finetuneshort performances broken down by \textit{Continent}. Lower $\sigma$ indicates more impartial model.}
    \label{fig:variances_demographics_dimensions}
    \vspace{-0.6cm}
\end{figure}

\paragraph{Whose values are easier for LMs to predict?}
LMs are subject to varying difficulty levels in predicting values across populations, \eg \llamathreeone is most accurate at predicting values of individuals from Oceania, as shown in Figure \ref{fig:variances_demographics_dimensions} (blue boxes). The disparity across sub-populations aligns with prior research that compares models' distributional outputs to human labels in WVS \citep{GlobalOpinionQA2024}. Refer to Figure \ref{fig:demographics_breakdown_full} in Appendix \S \ref{sec:probe_results} for full results of other demographics groups for GPT-4o, and Figure \ref{fig:evenness_age} to \ref{fig:evenness_sex} for \llamathreeone.

\paragraph{\textit{Equity} in individualistic value reasoning.} 
Table \ref{tab:equity_index_compare_models} shows the \even (\evenshort) of various frontier LMs. Notably, models with similar proficiency in individualistic value reasoning (indicated by accuracies in Figure \ref{fig:model_comparisons}) may have drastically different \evenshort, revealing discrepant equity levels across populations. For instance, both \texttt{GPT-4o (0513)} and \texttt{Llama-3.1-70B} have an accuracy of 63.7. However, \texttt{GPT-4o (0513)} has higher \evenshort (2.87), compared to \texttt{Llama-3.1-70B} (1.94), indicating a less equitable value representation. We introduce \evenshort as a new quantifiable measure of the equity of LMs. \evenshort presents complementary metrics to proficiency for assessing LMs in individualistic value reasoning.

\begin{figure}[t]
    \centering
    \includegraphics[width=0.48\textwidth]{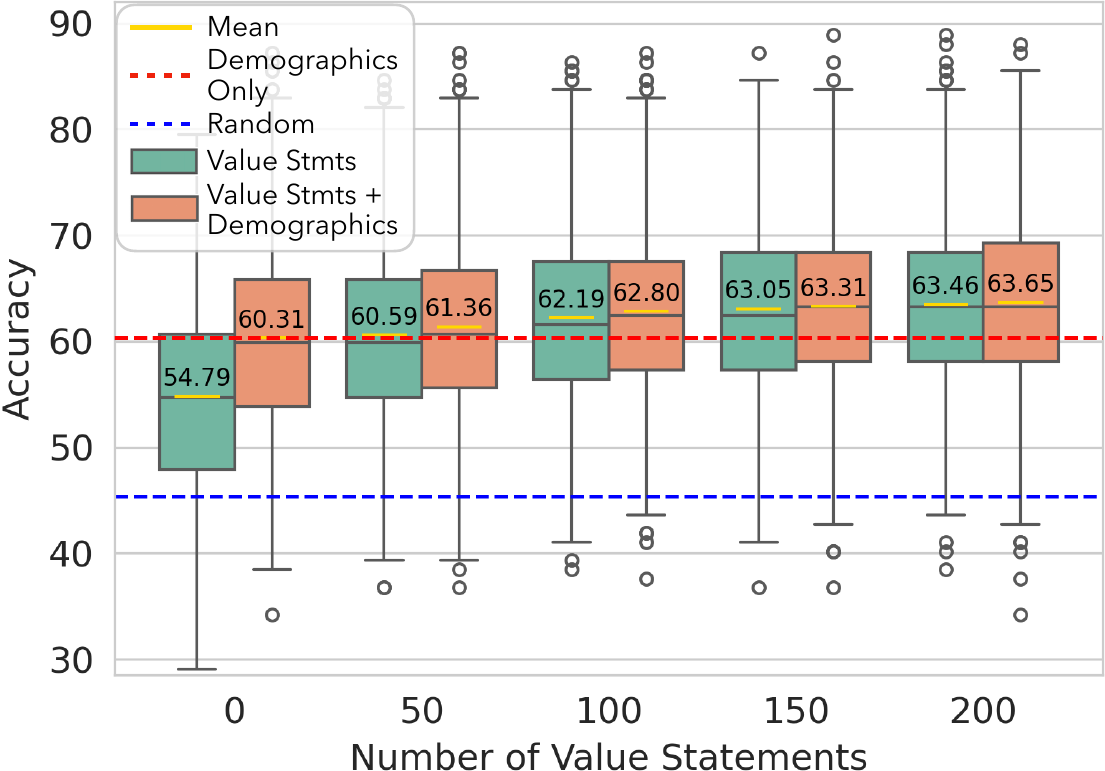}
    \vspace{-0.6cm}
    \captionof{figure}{The effect of different numbers of demonstration statements, and with or without demographics statements on GPT-4o's performance.}
    \label{fig:probe_ablations}
    \vspace{-0.5cm}
\end{figure}

\paragraph{Impact of the number of demonstration statements.}
Figure \ref{fig:probe_ablations} shows that increasing the number of demonstration statements improves GPT-4o's accuracy. Notably, even 50 statements boost accuracy from 54.79 to 60.59, demonstrating the impact of a small set of value expressions in guiding models to understand individualistic values.

\paragraph{Demographics descriptions do not teach LMs accurate sense of individualistic values.} 
Figure \ref{fig:probe_ablations} shows that when only demographic statements are provided (leftmost orange box), GPT-4o achieves 60.31\% accuracy, slightly lower than the 60.59\% achieved with 50 value-expressing statements. Adding demographic statements alongside value statements consistently yields marginally higher performance, though not statistically significant. Notably, when GPT-4o is given more value demonstrations, its accuracy improves compared to scenarios with fewer value statements and demographic data. This indicates that value statements encapsulate key latent information essential for approximating individual uniqueness. For weaker models like GPT-4o-mini, including demographics significantly improves predictions compared to value statements alone, likely due to challenges in interpreting descriptive value expressions (see Figure \ref{fig:probe_ablations_mini} in Appendix \ref{assec:probing_results}). Importantly, relying solely on demographics risks reinforcing stereotypical group-based interpretations, undermining nuanced understanding of individual values.

Please refer to \S \ref{assec:probing_results} for full probing experiments.

%% file: sections/4_finetune.tex
\section{Training Models to Capture Patterns and Dynamics of Individualistic Values}
\label{ssec:supervised_reasoner}

\subsection{Method}
\label{ssec:finetune_method}

The rich data in \dataset allows us to train a series of \finetune models (\finetuneshort) based on \llamathreeone.
We form the training data using value statements from $\mathbb{I}_{\text{train}}$. 
Each training data contains $d$ demonstration statements (\texttt{demo})\footnote{$d=$\texttt{200} or \texttt{mixed} stands for drawing 200 or randomly between 50-200 demonstrations, respectively.} and statement candidates of a probing question (\texttt{probe}), all from the same individual. The model takes in the \texttt{demo} statements and outputs a choice among the \texttt{probe} candidates. Both \texttt{demo} and \texttt{probe} can take either \databinary (\texttt{p}) or \datarefined (\texttt{r}) forms. For each of the 253 questions ($q$), we sample \texttt{N} individuals from $\mathbb{I}_{\text{train}}$ to form different demonstration sets for $q$, and use each individual's statement choice of $q$ as the gold label, forming $253 \times$\texttt{N} training data. Full training details are shown in Appendix \S \ref{assec:finetune_setups}.

Our goal in training the \finetuneshort is not to \textit{solve} the individualistic value reasoning mission but to explore how data and LMs can be combined to reveal meaningful patterns in human values and assess the data-driven performance upper-bound for this task.
We include both statistics and LM-based baselines. For statistics-based baselines, we consider selecting the statement for $I_i \in \mathbb{I}_{\text{eval}}$ based on (1) \texttt{Global (majority vote)}: the majority vote across the global pool of individuals ($\mathbb{I}_{\text{train}}$); (2) \texttt{Resemble (1600/all, top 1)}: the statement choice of $I_j \in \mathbb{I}_{\text{train}}$ who shares the most number of common demonstration statements with $I_i$, among 1,600 randomly selected or the overall pool of individuals; (3) \texttt{Resemble (1600/all, top cluster)}: the majority vote among the top cluster of training individuals who share the most number of common demonstration statements with $I_i$, among 1,600 randomly selected or the overall pool of individuals. For LM-based baselines, we consider (1) \texttt{GPT-4o (no demo.)}: GPT-4o without demonstrations; (2) \texttt{GPT-4o (only demographics)}: GPT-4o with only demographics information; (3) \texttt{GPT-4o (200 demo.)}: GPT-4o with 200 demonstrations; (4) \texttt{\llamathreeone (200 demo.)}: \llamathreeone with 200 demonstrations. Details of baseline are shown in Appendix \S \ref{assec:finetune_setups}. 
Finally, we evaluate models using both \databinary and \datarefined setups for both demonstrations and probes.

\input{tables/improved_reasoning_results}

\subsection{Results}

\paragraph{\finetuneshort{s} improves over individualistic value reasoning.}
Table \ref{tab:improved_reasoning_results} shows that the best-performing \finetuneshort, \texttt{[probe=p+r,demo=mix:200,N=1600]}, achieves 46.6\% relative improvements compared to the zero-shot setting, \texttt{[Llama-3.1-8B (200 demo.)]}.
Compared to the best-performing GPT-4o configuration, \texttt{[GPT-4o (only demographics)]}, the best trained model achieves 34.0\% relative improvement, showing that smaller and less capable models can greatly improve over more capable models with supervision of individualistic values data.

We compare models trained to predict \datarefined and \databinary question forms. The \databinary-specialized model, \texttt{[probe=p,demo=mixed,N=800]}, does well only on \databinary questions without extrapolating to \datarefined questions. The \datarefined-specialized model, \texttt{[probe=r,demo=mixed,N=800]}, improves on \datarefined questions, while maintaining performance on \databinary questions, despite not as high as the \databinary-specialized model. We choose to combine both \datarefined and \databinary probes for training to have a balanced performance between the two forms.

We show that using a mixed number of demonstrations (\texttt{[probe=p+r,demo=mixed,N=800]}) improves performance (66.49) over a fixed 200 demonstrations (\texttt{[probe=p+r,demo=200,N=800]}) when tested on 200-demonstration examples (66.14).
This shows that despite we seemingly provide less information during training (\ie less total number of demonstration statements for \texttt{[probe=p+r,demo=mixed,N=800]}), the diversity brought by the mixed number of demonstrations provides richer variety of information for the model to gain stronger generalizability. Even better, combining data with 200 and mixed demonstrations produces the best-performing model, \texttt{[probe=p+r,demo=mixed+200,N=800]}.

\begin{figure}[t]
    \centering
    \includegraphics[width=0.48\textwidth]{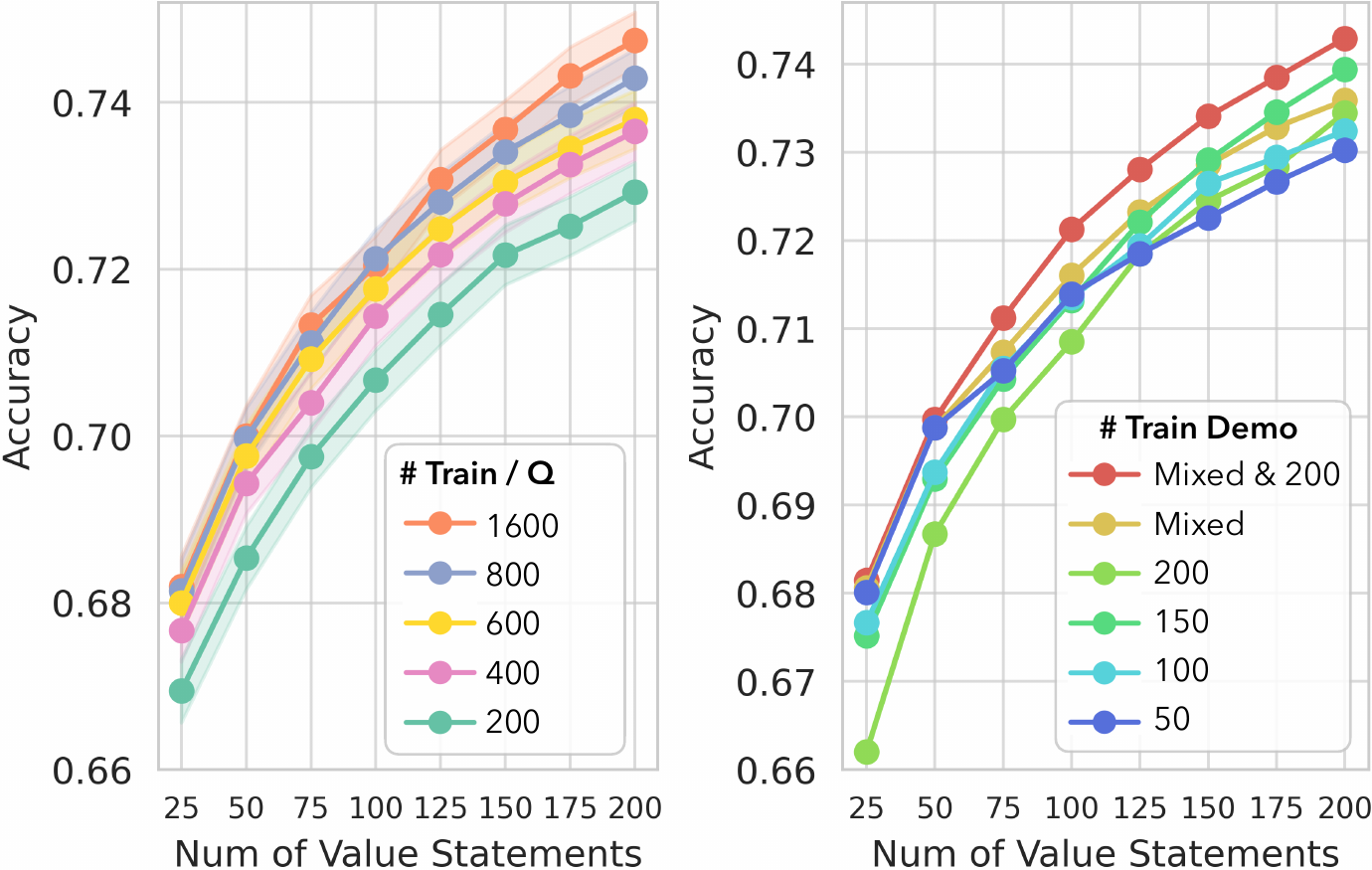}
    \captionof{figure}{(Left) The effect of training data size. (Right) The impact of varied numbers of training demonstration statements on the performances of models trained with data of different mixtures of demonstrations.}
    \label{fig:finetune_ablations}
    \vspace{-0.4cm}
\end{figure}

Finally, scaling up the training data to 1.6K individuals (\texttt{[probe=p+r, demo=mixed+200, N=1600]}) further enhances performance. Figure \ref{fig:finetune_ablations} (Left) shows that increasing the data size consistently improves \finetuneshort's performance across varying numbers of demonstrations.

\paragraph{Similar value demonstration trajectories can help predict an individual's value choices.}  Statistics-based baselines rely on oracle access to data from all individuals, allowing for strong predictive power by aggregating value choices from similar individuals, as demonstrated by \texttt{[Resemble (top cluster)]}. These baselines significantly outperform zero-shot LM-based approaches, which risk guessing individual value choices without explicit teaching.
However, the best-performing trained model, \texttt{[probe=p+r,demo=mix:200,N=1600]} (67.67), outperforms \texttt{[Resemble (top cluster)]} (66.60), despite using data from only 1.6K individuals per question---substantially fewer than the 92K used by statistics-based baselines. When compared against a statistics-based baseline using the same 1.6K randomly sampled individuals, the trained model achieves a significantly higher accuracy (67.67 vs. 62.38). This shows the superior sample efficiency and generalizability of LMs in capturing individual value patterns.


\paragraph{Improved \evenshort.}
The best \finetuneshort achieves improved \evenshort (2.22) compared to zero-shot \llamathreeone (2.97). 
Figure \ref{fig:variances_demographics_dimensions} highlights the performance improves greatly among previously \textit{underperforming} demographic groups. For example, \finetuneshort achieves an +18.24\% absolute performance gain in South America, the lowest-performing region, compared to Oceania (+12.51\%) and North America (+14.38\%). This shows that training models on diverse, globally representative data alleviates the partiality of off-the-shelf \llamathreeone in reasoning about demographic differences. Breakdowns of all demographic dimensions are shown in Figure \ref{fig:evenness_age}-\ref{fig:evenness_sex} and Table \ref{tab:variances_demographics_dimensions} in Appendix \S \ref{assec:finetune_results}.

\begin{figure}[t]
    \centering
    \includegraphics[width=0.48\textwidth]{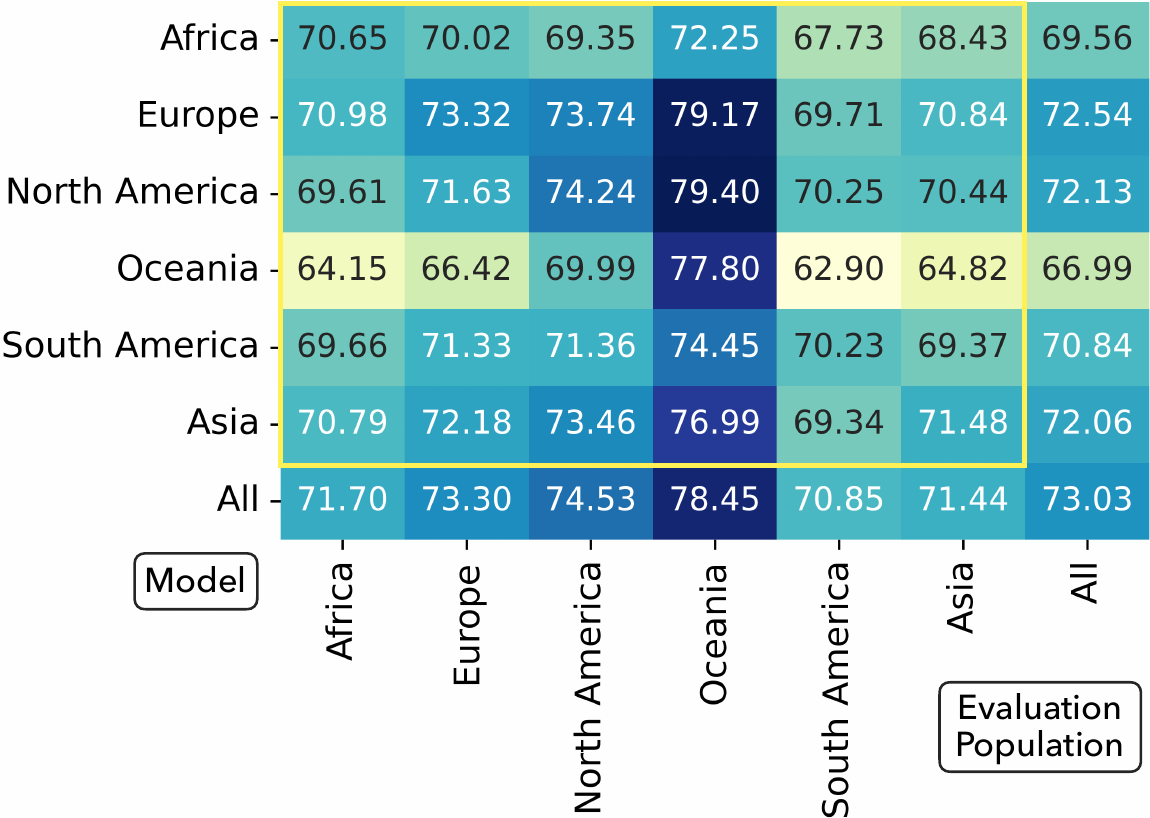}
    \vspace{-0.7cm}
    \caption{Continent-specific \finetuneshort evaluated with continent-specific test sets.}
    \label{fig:continent_models}
    \vspace{-0.4cm}
\end{figure}

\paragraph{A hybrid number of demonstrations improves reasoning generalizability.}
Figure \ref{fig:finetune_ablations} (Right) shows that increasing the number of test demonstrations enhances models' performances. Notably, training \finetuneshort with a random mix of 50 to 200 demonstrations outperforms training with any fixed number of statements. Yet, a model trained with a maximum of 200 demonstrations performs only moderately well but struggles with fewer test demonstrations, where stronger extrapolation is needed. Conversely, a model trained on 50 demonstrations excels with limited evidence but struggles to generalize with more. Training with a random number of demonstrations (50 to 200) yields strong overall performance but underperforms when tested with 150 or 200 demonstrations. Thus, we trained a model on both mixed demonstrations and the full 200, achieving the best results. These findings highlight the synergy between diverse demonstration configurations for improving individualistic value reasoning across both abstract and specific evidence.

\input{tables/train_mix_demographics}

\paragraph{Models trained on different global regions show discrepant predictive power over cross-region individuals.}
To evaluate how regional data impacts a model's ability to reason about diverse populations, we trained models using data from each of six continents (Figure \ref{fig:continent_models}). These continent-specific models showed significantly varying performances. They typically achieve the best (Europe, North America, Asia) or second-best (South America, Africa) performance for the corresponding continent's test population (except Oceania), emphasizing the strong impact of regional data on performance for the corresponding populations. Sometimes, we also observe a particularly strong performance of some content-specific models on other populations. For instance, \textit{North America} model achieves the best performance on the \textit{South America} test data; \textit{European} model achieves the best performance on \textit{Africa} test set. This trend aligns with geographical proximity and the commonly held impression of a close influence between the source and test continents.

Oceania was an exception: while most models (except the Africa model) performed well on Oceania’s test set, the model trained on Oceania data underperformed on all test sets except Oceania and North America. We hypothesize this is due to the limited diversity of the Oceania dataset, which consists entirely of responses from New Zealand.

To quantitatively confirm the homogeneity of the Oceania data, we assess whether respondents' answers are skewed across survey questions (i.e., disproportionately clustered around certain options). To measure this, we compute the \textbf{Shannon’s Evenness Index} ($J'$). This metric captures how evenly distributed responses are across answer choices, analogous to species evenness in ecology. The index ranges from 0 (completely uneven) to 1 (perfectly even), with lower values indicating more homogeneous response patterns.

$$J' = \frac{H'}{H'_{max}} = \frac{H'}{\ln(S)}$$ where $H' = -\sum_{i=1}^{S} p_i \ln(p_i)$

\begin{figure}[t]
    \centering
    \includegraphics[width=0.42\textwidth]{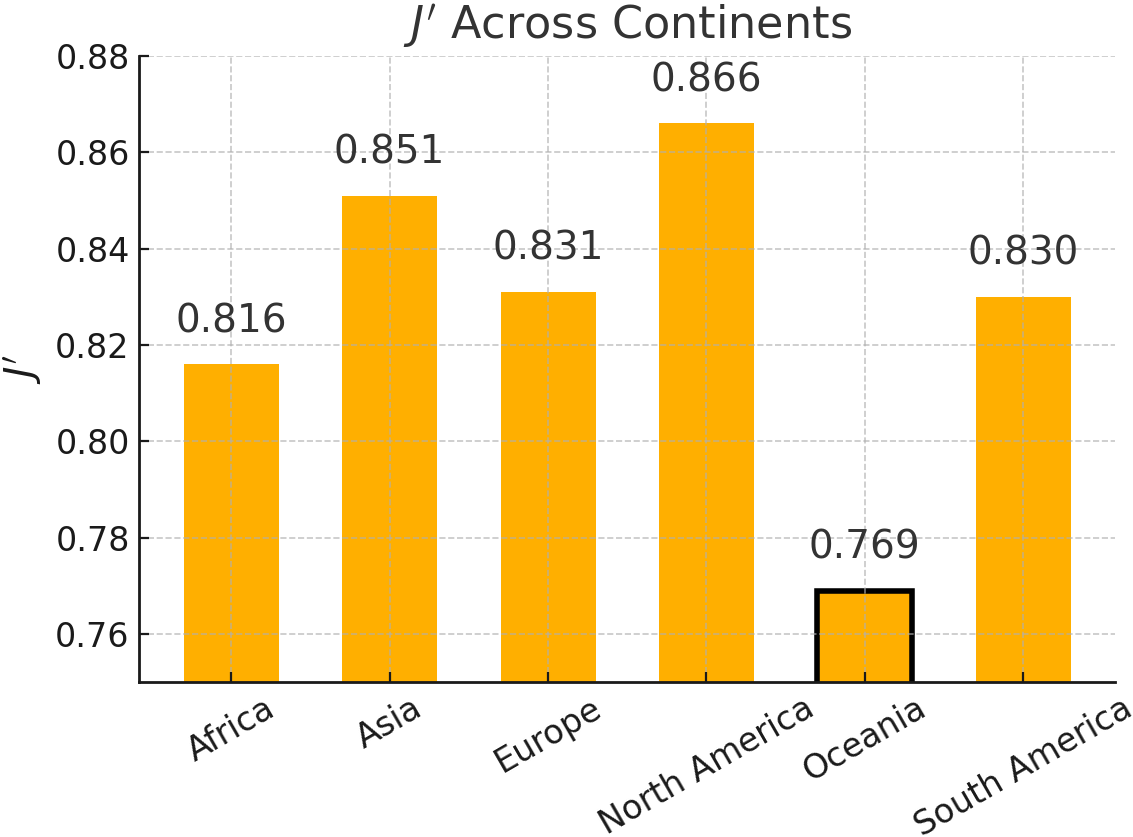}
    \vspace{-0.3cm}
    \caption{The Shannon’s Evenness Index ($J'$) across different continents. Lower $J'$ values indicate a more uneven distribution across all answer choices.}
    \label{fig:shannon_entropy_Oceania}
    \vspace{-0.5cm}
\end{figure}

As we can see from Table \ref{fig:shannon_entropy_Oceania}, Oceania indeed has a lower $J'$ score compared to other continents, meaning that it has a more uneven distribution across all answer choices. This exactly confirms that Ocreania data cluster around certain answer choices, thus proving to be less diverse. This show that a homogeneous training set limits the model's ability to capture diverse value patterns; correspondingly, a homogeneous test set is easier to predict, even for regional models. 

Finally, the globally trained model consistently matched or outperformed continent-specific models across all regions, emphasizing the importance of diverse, cross-regional data for robust reasoning about global human value patterns.

\paragraph{Training on demographic descriptions does not generalize to test cases with value-expressing statements.}

We train a \finetuneshort using only demographic descriptions (\eg ``I'm 25-34 years old'') instead of value-expressing statements. As shown in Table \ref{tab:main_results_finetune}, this model fails to generalize to test cases using value-expressing demonstrations. Conversely, models trained on value-expressing statements struggle with demographic-based demonstrations but perform slightly better overall (68.14 vs. 65.62). Training with a mix of demographic and value-expressing data improves in both test cases (70.88), suggesting a mutually reinforcing relationship between the two data types.

%% file: tables/improved_reasoning_results.tex
\begin{table}[t!]
\centering
\small
\begin{tabular}{l@{\hspace{0.9\tabcolsep}}
r@{\hspace{0.95\tabcolsep}}
r@{\hspace{0.95\tabcolsep}}
r@{\hspace{0.95\tabcolsep}}
}

\toprule

& \textbf{\Databinary} & \textbf{\Datarefined} & \textbf{All} \\ 

\midrule

\scriptsize{\texttt{Random}} & 45.37 & 27.87 & 36.62 \\

\scriptsize{\texttt{Global (majority vote)}} & 64.95 & 48.70 & 56.83 \\

\scriptsize{\texttt{Resemble (1600, top 1)}} & 65.65 & 47.55 & 56.60 \\ 

\scriptsize{\texttt{Resemble (1600, top cluster)}} & 70.39 & 54.36 & 62.38 \\

\scriptsize{\texttt{Resemble (all, top 1)}} & 69.83 & 53.51 & 61.67  \\

\scriptsize{\texttt{Resemble (all, top cluster)}} & 73.73 & 59.47 & 66.60 \\

\midrule

\scriptsize{\texttt{GPT-4o (no demo.)}} & 54.79 & 33.06 & 43.93 \\

\scriptsize{\texttt{GPT-4o (only demographics)}} & 60.31 & 40.69 & 50.50 \\

\scriptsize{\texttt{GPT-4o (200 demo.)}} & 63.46 & 35.59 & 49.52 \\

\scriptsize{\texttt{Llama-3.1-8B
(200 demo.)}} & 54.34 & 37.97 & 46.16 \\

\midrule

\scriptsize{\texttt{[probe=p,demo=mixed,N=800]}} & \underline{73.59} & 44.08 & 58.84 \\

\scriptsize{\texttt{[probe=r,demo=mixed,N=800]}} & 73.25 & \underline{59.94} & \underline{66.59} \\

\scriptsize{\texttt{[probe=p+r,demo=200,N=800]}} & 73.45 & 58.84 & 66.14 \\

\scriptsize{\texttt{[probe=p+r,demo=mixed,N=800]}} & \underline{73.59} & 58.27 & 66.49 \\

\scriptsize{\texttt{[probe=p+r,demo=mixed+200,N=800]}} & \textbf{74.29} & \textbf{60.23} & \textbf{67.26} \\

\midrule

\rowcolor{cyan}  
\scriptsize{\texttt{[probe=p+r,demo=mixed+200,N=1600]}} & \textbf{74.74} & \textbf{60.60} & \textbf{67.67} \\

\bottomrule
\end{tabular}
\caption{Results of \finetuneshort models for improved individualistic value reasoning for both the \databinary and \datarefined evaluation setups. For the middle section of ablation models, the best performances are \textbf{bolded}, and the second best performances are \underline{underlined}. All results in this table are obtained by giving 200 demonstration value-expressing statements during test time. For naming convention, \texttt{[probe=p+r,demo=mixed+200,N=800]} refers to a model trained using both \textit{mixed} and fixed \textit{200} demonstration statements, and evaluated with probing statements in both \textit{polar} and \textit{refined} forms. Each of the 253 value questions has 200 examples for each setup---(mixed, refined), (mixed, polar), (200, refined), and (200, polar)---totaling 800 examples. Please refer to Table~\ref{tab:training_config_baselines} for details on the baseline configurations.}
\label{tab:improved_reasoning_results}
\vspace{-0.3cm}
\end{table}

%% file: tables/train_mix_demographics.tex
\begin{table}[t]
\small
\centering
\begin{tabular}{l@{\hspace{0.9\tabcolsep}}
l@{\hspace{0.9\tabcolsep}}
l@{\hspace{0.9\tabcolsep}}
l@{\hspace{0.95\tabcolsep}}
l@{\hspace{0.95\tabcolsep}}
l@{\hspace{0.95\tabcolsep}}}
\toprule

 \multicolumn{3}{c}{\textbf{\# Train Per Q}} & \multicolumn{3}{c}{\textbf{Evaluation}} \\

 \cmidrule(r){1-3} \cmidrule(r){4-6}
 
\small{Demogr.} & \specialcell{\small{Stmts}} & \small{Total} & \specialcell{\small{Stmts}} & \small{Demogr.} & \small{Avg.} \\
\midrule

 400 & 400 & 800 & \textbf{73.74} & \textbf{68.02} & \textbf{70.88} \\
 800 & 0  & 800 & 63.81 & 67.42 & 65.62 \\
 0 & 800 & 800 & 73.45 & 62.84 & 68.14 \\

\bottomrule
\end{tabular}
\vspace{-0.2cm}
\captionof{table}{Models trained with value-expressing statements, demographics descriptions, or both.}
\label{tab:main_results_finetune}
\vspace{-0.3cm}
\end{table}

%% file: sections/5_discussion.tex
\section{Discussion and Future Directions}
\label{sec:discussion}

\paragraph{Impact.} Predicting individual value statements in NLP has significant practical value, particularly for applications requiring a deep understanding of human behavior. For instance, mental health chatbots can offer more empathetic and context-aware support, while personalized recommendation systems can improve content relevance by aligning with user values. Moreover, understanding individual values is essential for developing culturally sensitive and inclusive AI, enabling more nuanced human-AI interactions.

Predicting individualistic human values also intersects with existing computational social science research (CSS) that simulates human behavior via LLMs. Recent work explores using LLMs to simulate social interactions and cultural dynamics \cite{ziems-etal-2024-large, zhou2024sotopia, social_simulacra}. Finally, computationally reasoning through individual human values also potentially add novel insights to other disciplines, such as moral philosophy or psychology. By bridging technical innovation into humanity research, value prediction not only refines AI systems for real-world alignment but also offers ethically scalable methods to study decision-making, enriching both technology development and social scientific inquiry.

\paragraph{Future.} Studying individualistic values is challenging due to the scarcity of rich, individual-level data that accurately represents personal value systems. While our adaptation of the WVS addresses this partially, it is limited by its reliance on static, abstract questions that lack the complexity of real-world human interactions.
Gathering ecologically valid data from dynamic, real-world interactions with humans is a critical next step for advancing individualistic alignment. Given the time and cost constraints, sample-efficient methods (\eg active learning or interactive questioning) are promising avenues.
Exploring low-dimensional representations of human values to increase tractability while maintaining fidelity will also be important. 
Despite the complexity of human decisions, underlying structures may explain much of their variation, making this an ideal focus for interdisciplinary work in statistics, cognitive science, and decision theory.
Finally, even given a good model of individual values and preferences, applying these representations to system behavior is non-trivial. Future research must address computational and data tradeoffs while accounting for the non-stationary and context-dependent nature of human preferences.

%% file: sections/6_related_works.tex
\section{Related Work}
\label{sec:related_work}

\paragraph{Pluralistic value alignment.}
Recent research in value alignment has significantly advanced the utility and safety of LMs \citep{ouyang2022rlhf,schulman2017ppo,rafailov2024dpo,bai2022hhrlhf}. However, general value alignment risks promoting a monolithic value representation \citep{ryan-etal-2024-unintended}. In response, recent calls for \textit{pluralistic alignment} highlights the need for AI systems to cater to the diverse needs of a broad population \citep{sorensen2024roadmappluralisticalignment}, encouraging methods \citep{modularpluralism2024,lake2024distributionalovertonpluralisminvestigating,chen2024palpluralisticalignmentframework}, benchmarks \citep{PERSONA2024}, and training data \citep{kirk2024prismalignment} developed to support this vision. Additionally, methods have been proposed for improving diversity by leveraging the collaboration of multiple LMs \citep{modularpluralism2024,chen2024reconcile,verga2024replacingjudgesjuries} and system messages \citep{lee2024aligningthousandspreferencesmessage}. Meanwhile, existing works measure the cultural disparity of LMs \citep{chiu2024culturalteaming,Rao2024NORMADAB} and improves models' cultural representations \citep{shi2024culturebank,Li2024CultureLLMIC,Fung2024MassivelyMK,Myung2024BLEnDAB}. However, most existing work in pluralistic alignment rely on pre-selected \textit{diversity-defining dimensions} for capturing variances among population, such as demographics \citep{moon2024personabackstory,kwok2024syntheticpersonas}, personality \citep{PERSONA2024,jiang2023evaluating,serapiogarcía2023personalitytraits, zhu2024personalityalignment}, writing styles \citep{han2024valueaugmentedsamplinglanguage,personalizedsoup2023}, and cultures \citep{Myung2024BLEnDAB}, forcing individuals into predefined buckets.

\paragraph{Individualistic value alignment and reasoning.} Related to individualistic value learning are the tasks of personalization and preference elicitation. Work on personalizing LMs aims to provide customized, user-specific responses across applications, such as summarization \citep{han2024valueaugmentedsamplinglanguage}, 
chatbot interactions \citep{xu-etal-2022-beyond}, movie tagging \citep{liu2024llmspersonaplug}, 
open-text generation \citep{zhu2024personalityalignment}, survey questions \citep{li-etal-2024-steerability}, simulated control tasks \citep{poddar2024personalizing}, and writing assistant \citep{mysore2023pearlpersonalizinglargelanguage}. To understand users' needs in specific tasks, active learning is applied to interactively and efficiently investigate people' preferences
\citep{keswani2024prosconsactivelearning,zhang2024selfexploring,ji2024activequeries,mehta2023sampleefficientreinforcementlearning,muldrew2024activepreferencelearning,piriyakulkij2024activepreference}. Uniquely, \cite{zhu2024personalityalignment} introduces \textit{personality alignment}, which is closely related to \textit{individualistic alignment} but emphasizing aligning models with psychometric dimensions capturing people's personalities. Our work differs from prior works by focusing on modeling and reasoning about individualistic human values rather than personality traits or application-driven personalization.

%% file: sections/7_conclusion.tex
\section{Conclusion}
\label{sec:conclusion}

We propose a bottom-up approach to pluralistic value learning by inducing individualistic values bottom-up. We harvest the well-established social science resource of WVS in a novel way to highlight frontier LMs' limitations in individualistic value reasoning. By fine-tuning models on WVS data, we uncover insights into global human values while exposing significant gaps in modeling individual value systems. Our work highlights key challenges in \textit{individualistic value reasoning} and the broader pursuit of \textit{individualistic alignment}.

%% file: sections/z_acknowledgement.tex
\section*{Acknowledgement}

We thank helpful discussions from Wenting Zhao, Ronan Le Bras, Jared Moore, Zaid Harchaoui, Pang Wei Koh, and Jena D. Hwang during various exploratory stages of the project. We thank Jacob Morrison and Jing-Jing Li for helping set up the training codebase with Open-Instruct. This work is supported by DARPA under the ITM program (FA8650-23-C-7316) and the Office of Naval Research (N00014-24-1-2207), Templeton World Charity Foundation (TWCF-2023-32585), and Allen Institute for Artificial Intelligence.

%% file: sections/z_limitations.tex
\section*{Limitations}
\label{sec:limitations}

The findings presented in this paper come with several important limitations that warrant careful consideration.

\paragraph{Data representation.} 
While \dataset represents a comprehensive dataset of human value expressions transformed from WVS, it is inherently limited by its reliance on static, survey-based questions. These questions may not fully encapsulate the complexity and dynamic nature of real-world human interactions and values. The transformation of the data set into standardized statements, while useful for model training and evaluation, may oversimplify nuanced expressions of real-world values.

\paragraph{Prompt design and generalizability.}
Due to computational constraints, we evaluated models using a single, carefully crafted prompt for probing (see \ref{assec:probing_setups}). Although using multiple prompts could potentially provide more robust and nuanced insights, this study focused on a single, well-crafted prompt to rigorously evaluate the models' performance in individualistic value reasoning. Furthermore, the cross-validation setup with three distinct evaluation configurations strengthens the reliability of the results. Despite this limitation, the consistency and clarity of the results across different configurations suggest that the conclusions about model performance remain sufficiently reliable. Future work should explore the impact of diverse prompt designs to further enhance the robustness of the findings.

\paragraph{Training data scale.}
Due to computational resource limitations, we trained the \finetune models using data from only 200 individuals per survey question. While increasing the training data from 100 to 200 individuals led to noticeable performance improvements, it remains unclear whether further increases in training data size could yield additional gains. We hope that future work can explore training with larger datasets, potentially uncovering new dynamics and further enhancing the models’ ability to reason about individualistic values.

\section*{Ethical Considerations}
\label{sec:ethical_considerations}

Individual alignment brings up several ethical considerations around the societal implications of building AI tailored towards individual values (for a thorough discussion, see \citet{kirk2024personalization}).

\paragraph{Privacy infringement.}
Individualistic value alignment naturally requires access to data that contains deeply private information about individual values and preferences. This concern is amplified when users anthropomorphize models tailored to their values, potentially leading to the disclosure of even more sensitive information. Additionally, using real-world data to understand individualistic values must be transparent to participants and users, who should provide informed consent.

\paragraph{Bias reinforcement.} 
A primary motivation for individualistic alignment is to bypass the popular need to put people into buckets while exploring the diversity space. Thus, it should be less prone to bias compared to typical alignment frameworks. However, other types of biases may occur if misleading features and evidence are used to draw conclusions about people's values, \eg confirmation biases (\ie overlooking evidence that addresses opposite or unexpected value conclusions), anchoring bias (\ie drawing conclusions of one's value choices by an uneven weighing of different evidence), and framing effect (\ie the interpretation of values is influenced by how they are described). Researchers must proactively consider these bias sources when developing technical solutions for individualistic value alignment.

\paragraph{Misuse or over-reliance on individualized AI.} By tailoring AI systems to align closely with personal values, there is a danger that these systems could be exploited for manipulative purposes, such as influencing people's political views and social behaviors. Such hyper-individualized human-AI interaction can also reduce users' autonomy, jeopardizing independent thought. To mitigate these risks, safeguards should be in place to ensure that AI systems empower users rather than manipulate them based on their personal values, maintaining fairness and diversity in the process.

%% file: sections/z_appendix.tex
\clearpage
\newpage
\appendix

\input{sections/appendix/related_work}

\input{sections/appendix/dataset}

\newpage
\clearpage

\input{sections/appendix/probing}
\newpage
\clearpage

\input{sections/appendix/finetune}

\section{Acknowledgement of AI Assistance}
\label{asec:ai_assistance}

We use AI solely for assistance with the language of the paper.

%% file: sections/appendix/related_work.tex
\section{Additional Related Works}
\label{asec:additional_related_work}

\paragraph{How are human values studied across scholarly fields?} 
Despite the extensive studies and debates over human values across scholarly fields, it remains a mystery how to best represent them. One famous social psychology theory, Schwartz's Theory of Basic Values \citep{schwartz2012overview}, strives to define top-down categories of fundamental human values. Other empirical psychometric instruments such as self-report questionnaires \citep{1a35f96dcc55415ca791831dced159cd,MAIO20101,curry2019mapping},
behavioral observations \citep{KALIMERI2019428}, and controlled experiments \citep{curry2019good} are also commonly used in the attempt to describe people's value systems. Philosophers hold distinct views towards the meaning and scope of human values. For instance, distinctions had been made between intrinsic vs. extrinsic values \citep{sep-value-intrinsic-extrinsic}, value monism \citep{sep-monism} vs. pluralism \citep{sep-value-pluralism} that debate about whether there are one or more fundamental values, and whether there exist human values that are incommensurable (\ie cannot be traded-off; \citep{sep-value-incommensurable}). Social science research like Pew Public Opinion Polling \citep{pew_research_center} and World Value Survey \citep{worldvaluesurveywave7} conducts large-scale empirical surveys to collect people's value opinions across regions.

%% file: sections/appendix/dataset.tex
\section{Details of the \dataset Dataset}
\label{asec:dataset_details}

\paragraph{Dataset License}

\href{https://www.worldvaluessurvey.org/WVSContents.jsp?CMSID=DataDissemination&CMSID=DataDissemination#:~:text=The%20World%20Values%20Survey%20data,tools%20developed%20for%20online%20analysis.}{The World Values Survey data} is publicly available for free under a non-redistribution data use license for research purposes. Our use of this resource complies with these licensing requirements.

\paragraph{Dataset Statistics}
The complete details of the statistics of the \dataset is shown in Table \ref{tab:dataset_stats_full}. The set of considered demographics-related WVS questions are shown in Table \ref{tab:demographics_1}, \ref{tab:demographics_2}, and \ref{tab:demographics_3}.

\input{tables/dataset_statistics_full}

\paragraph{Data Conversion Details}

The original World Value Survey contains unstructured questions with varying numbers of answer options or scales. Previous works have adopted the original questions formats as-is \citep{GlobalOpinionQA2024} or converting all questions to Likert scale format \citep{worldvaluesbench2024} for evaluating language models' distributional knowledge of values across global population groups. However, we identify the unnatural multiple-choice question formats and somewhat fragmented language descriptions may impair the nuanced understanding of pragmatics compared to what natural language statements can convey.

Thus, we standardized all questions with multiple answer choices or ratings onto a Likert scale by converting them into independent sets of unified natural language statements that reflect people's value preferences. To do so, we morph the survey question descriptions (\eg Q131 of WVS: ``Could you tell me how secure do you feel these days?'') and the answer options (\eg 1. ``very secure;'' 2. ``quite secure;'' 3. ``not very secure;'' 4. ``not at all secure.'') together into self-contained statements that express a person's value preference (\eg ``I feel very secure/quite secure/not very secure/not at all secure these days.''). Some questions of WVS have Likert scale answer space (\eg Q158: From scale 1 (completely disagree) to 10 (completely agree), select how much you agree that ``science and technology are making our lives healthier, easier, and more comfortable.'') since the granularity of the answer space makes it noisy to calibrate with language statements that precisely captures the fine-grained scaled ratings, we map the scales to four answer choices that capture the broad extent and polarity of scaled answers to reduce the variability and noises caused by overly fine-grained answer options. To further reduce the noised variations introduced by fine-grained answer options, we create another variation of the dataset by grouping statements sharing the same polarity together, \eg ``agree strongly'' and ``agree'' are grouped into ``agree''; ``disagree strongly,'' and ``disagree'' are grouped into ``disagree;'' ``neither agree nor disagree'' is kept as a neural answer choice. 
Figure \ref{fig:concept} shows an example conversion of original questions in WVS to our value statement format. More example converted statements of \dataset are shown in Table \ref{tab:example_converted_statements}.

Finally, we also convert questions related to the demographic background of people into identity-declaring statements, \eg I'm currently in Andorra; I'm an immigrant to this country (see Table \ref{tab:demographics_1}-\ref{tab:demographics_5} for the considered set of demographics questions).

\input{tables/demographics}

\input{tables/example_converted_statements}

\paragraph{Motivation of Data Format Transformation}

Although the original survey questions for WVS seem to be directly usable for evaluating LMs, they are in practice hard to use as-is as those questions take hybrid questions forms, spanning across multiple choice questions (with different numbers of answer choices), ranking questions, and rating questions (with different scaler scales). In addition, some questions in the original WVS have very fine-grained answer space (e.g., rating between 1 to 10), which makes the task too nuanced and difficult for LMs to elicit meaningful model performance rather than prediction noises. Also, some question descriptions have fragmented language forms, potentially introducing unnecessary noises for LMs to interpret the meaning of the questions.







To address the challenges posed by the heterogeneous formats in WVS, we made a deliberate design choice to convert all questions into sets of natural language, value-expressing statements with a unified format. We further introduce two types of statements, \textit{polarity-grouped} and \textit{refined}, to support evaluation and training at varying levels of granularity. This unified design simplifies the formulation of evaluation tasks and the definition of evaluation metrics. It also facilitates model training by ensuring that all demonstration statements follow a consistent format, eliminating the need for models to handle diverse question types as a confounding factor.

Although our current work does not apply this dataset for direct model alignment via RLHF, we see strong potential for its use in future alignment research. For example, pairs of statements from our dataset can be repurposed as preference data to explore novel approaches to individualistic value alignment. Overall, we believe our data conversion provides a flexible and practical resource for advancing alignment techniques grounded in individual-level human values.

In summary, our data conversion is motivated by the goal of enabling more effective evaluation and training. That said, we acknowledge that alternative processing choices may also be valid, depending on the specific objectives of other research efforts.

%% file: tables/dataset_statistics_full.tex
\begin{table*}[t]
\centering
\begin{tabular}{l|r|rr|rr}
\toprule
\textbf{} & \textbf{} & \multicolumn{2}{c|}{\textbf{\Databinary}} & \multicolumn{2}{c}{\textbf{\Datarefined}} \\ 
\midrule

\textbf{Question Category} & \textbf{\#Q} & \textbf{\#S} & \textbf{\#S / \#Q} & \textbf{\#S} & \textbf{\#S / \#Q} \\
\midrule

Social Values, Attitudes \& Stereotypes & 45 & 103 & 2.29 & 145 & 3.22 \\
Happiness and Well-Being & 11 & 23 & 2.09 & 44 & 4.00 \\
Social Capital, Trust \& Organizational Membership & 44 & 88 & 2.00 & 163 & 3.70 \\ 
Economic Values & 6 & 12 & 2.00 & 22 & 3.67 \\ 
Corruption & 9 & 19 & 2.11 & 37 & 4.11 \\
Migration & 10 & 29 & 2.90 & 33 & 3.30 \\ 
Security & 21 & 42 & 2.00 & 68 & 3.24 \\ 
Postmaterialist Index & 6 & 24 & 4.00 & 24 & 4.00 \\ 
Science \& Technology & 6 & 12 & 2.00 & 24 & 4.00 \\
Religious Values & 12 & 27 & 2.25 & 42 & 3.50 \\ 
Ethical Values and Norms & 23 & 46 & 2.00 & 92 & 4.00 \\
Political Interest \& Political Participation & 35 & 92 & 2.63 & 135 & 3.86 \\
Political Culture \& Political Regimes & 25 & 50 & 2.00 & 100 & 4.00 \\
\midrule

\textbf{Total} & \textbf{253} & \textbf{567} & \textbf{2.24} & \textbf{929} & \textbf{3.67} \\
\bottomrule
\end{tabular}
\caption{Number of questions (\#Q), statements (\#S), and the average number statements per question (\#S / \#Q) counts broken down by question category.}
\label{tab:dataset_stats_full}
\end{table*}

%% file: tables/demographics.tex
\begin{table*}[t]
\centering
\begin{tabular}{p{2cm}p{2cm}p{1.6cm}lp{3.2cm}}
\toprule

\textbf{Dimension} & \textbf{QID} & \textbf{Answer Type} & \textbf{Demographics Var} & \textbf{Conversion Template} \\
\midrule

Country & B\_COUNTRY & Code & text & I am currently in \{var\} \\
\midrule

Continent & \specialcell{B\_COUNTRY\\ \_to\_continent} & Code & text & I am currently in \{var\} \\
\midrule

Sex & Q260 & MC & \specialcell{- ``male" \\ - ``female"} & I am a \{var\} \\
\midrule

Age & X003R & MC & \specialcell{- ``16-24" \\
- ``25-34" \\
- ``35-44" \\
- ``45-54" \\
- ``55-64" \\
- ``65+"} & I am \{var\} years old \\
\midrule

Immigrant & Q263 & MC & \specialcell{- ``born in" \\ - ``an immigrant to"} & I am \{var\} this country \\
\midrule

\specialcell{Immigrant \\ (mother)} & Q264 & MC & \specialcell{- ``born in" \\ - ``an immigrant to"} & My mother is \{var\} this country \\
\midrule

\specialcell{Immigrant \\ (father)} & Q265 & MC & \specialcell{- ``born in" \\ - ``an immigrant to"} & My father is \{var\} this country \\
\midrule

\specialcell{Country \\ of birth} & Q266 & Code & text & I was born in \{var\} \\
\midrule

\specialcell{Country \\ of birth \\ (mother)} & Q267 & Code & text & My mother was born in \{var\} \\
\midrule

\specialcell{Country \\ of birth \\ (father)} & Q268 & Code & text & My father was born in \{var\} \\
\midrule

Citizen & Q269 & MC & \specialcell{- ``citizen" \\ - ``not a citizen"} & I am \{var\} of this country \\
\midrule

Number of people in household & Q270 & Numerical & number & There are \{var\} people in my household \\
\midrule

Live with parents & Q271 & MC & \specialcell{- ``do not live" \\ - ``live"} & I \{var\} with my parents or parents-in-law \\
\midrule

Language at home & Q272 & Code & text & I normally speak \{var\} at home \\
\bottomrule

\end{tabular}
\caption{Demographics dimensions, corresponding question ID (QIDs) in the original \wvs, the question type, the demographics variables, and the conversion templates for converting the raw questions from \wvs to statements in \dataset. (Part 1)}
\label{tab:demographics_1}
\end{table*}

\begin{table*}[t]
\centering
\begin{tabular}{p{2.8cm}p{0.7cm}p{2.2cm}lp{3.8cm}}
\toprule

\textbf{Dimension} & \textbf{QID} & \textbf{Answer Type} & \textbf{Demographics Var} & \textbf{Conversion Template} \\
\midrule

Marital status & Q273 & MC & \specialcell{- ``married" \\ - ``living together as married" \\ - ``divorced" \\ - ``separated" \\ - ``widowed" \\ - ``single"} & I am \{var\} \\
\midrule

Number of children & Q274 & Numerical & number & I have \{var\} children \\
\midrule

Highest educational level & Q275 & MC & \specialcell{- ``early childhood education or \\ no education" \\ - ``primary education" \\ - ``lower secondary education" \\ - ``upper secondary education" \\ - ``post-secondary non-tertiary \\ education" \\ - ``short-cycle tertiary education" \\ - ``bachelor or equivalent" \\ - ``master or equivalent" \\ - ``doctoral or equivalent"} & The highest educational level that I have attained is \{var\} \\
\midrule

Highest educational level (spouse or partner) & Q276 & MC & \specialcell{- ``early childhood education or \\ no education" \\ - ``primary education" \\ - ``lower secondary education" \\ - ``upper secondary education" \\ - ``post-secondary non-tertiary \\ education" \\ - ``short-cycle tertiary education" \\ - ``bachelor or equivalent" \\ - ``master or equivalent" \\ - ``doctoral or equivalent"} & The highest educational level that my spouse or partner has attained is \{var\} \\
\midrule

Highest educational level (mother) & Q277 & MC & \specialcell{- ``early childhood education or \\ no education" \\ - ``primary education" \\ - ``lower secondary education" \\ - ``upper secondary education" \\ - ``post-secondary non-tertiary \\ education" \\ - ``short-cycle tertiary education" \\ - ``bachelor or equivalent" \\ - ``master or equivalent" \\ - ``doctoral or equivalent"} & The highest educational level that my mother has attained is \{var\} \\

\bottomrule

\end{tabular}
\caption{Demographics dimensions, corresponding question ID (QIDs) in the original \wvs, the question type, the demographics variables, and the conversion templates for converting the raw questions from \wvs to statements in \dataset. (Part 2)}
\label{tab:demographics_2}
\end{table*}

\begin{table*}[t]
\centering
\begin{tabular}{p{1.8cm}p{0.7cm}p{1cm}p{8.5cm}p{1.8cm}}
\toprule

\textbf{Dimension} & \textbf{QID} & \textbf{Answer Type} & \textbf{Demographics Var} & \textbf{Conversion Template} \\
\midrule

Highest educational level (father) & Q278 & MC & \specialcell{- ``early childhood education or no education" \\ - ``primary education" \\ - ``lower secondary education" \\ - ``upper secondary education" \\ - ``post-secondary non-tertiary education" \\ - ``short-cycle tertiary education" \\ - ``bachelor or equivalent" \\ - ``master or equivalent" \\ - ``doctoral or equivalent"} & The highest educational level that my father has attained is \{var\} \\
\midrule

Employment status & Q279 & MC & \specialcell{- ``employed full time" \\ - ``employed part time" \\ - ``self employed" \\ - ``retired or pensioned" \\ - ``a housewife and not otherwise employed" \\ - ``a student" \\ - ``unemployed"} & I am \{var\} \\
\midrule

Employment status (spouse or partner) & Q280 & MC & \specialcell{- ``employed full time" \\ - ``employed part time" \\ - ``self employed" \\ - ``retired or pensioned" \\ - ``a housewife and not otherwise employed" \\ - ``a student" \\ - ``unemployed"} & My spouse or partner is \{var\} \\
\midrule

Occupational group & Q281 & MC & 
\specialcell{
- ``never had a job" \\ 
- ``a professional and technical job, e.g., \\doctor, teacher, engineer, artist, accountant, nurse" \\
- ``a higher administrative job, e.g., banker, \\ executive in big business, high government  \\ official, union official" \\
- ``a clerical job, e.g., secretary, clerk, \\ office manager, civil servant, bookkeeper" \\
- ``a sales job, e.g., sales manager, shop \\ owner, shop assistant, insurance agent, buyer" \\
- ``a service job, e.g., restaurant owner, \\ police officer, waitress, barber, caretaker" \\
- ``a skilled worker job, e.g., foreman, motor \\ mechanic, printer, seamstress, tool and die \\ maker, electrician" \\
- ``a semi-skilled worker job, e.g., bricklayer, \\ bus driver, cannery worker, carpenter, sheet \\ metal worker, baker" \\
- ``an unskilled worker job, e.g., labourer, \\ porter, unskilled factory worker, cleaner" \\
- ``a farm worker job, e.g., farm laborer, tractor driver" \\
- ``a farm owner or farm manager job"} & I have \{var\} \\
\bottomrule

\end{tabular}
\caption{Demographics dimensions, corresponding question ID (QIDs) in the original \wvs, the question type, the demographics variables, and the conversion templates for converting the raw questions from \wvs to statements in \dataset. (Part 3)}
\label{tab:demographics_3}
\end{table*}

\begin{table*}[t]
\centering
\begin{tabular}{p{2.5cm}p{0.7cm}p{1cm}lp{2.5cm}}
\toprule

\textbf{Dimension} & \textbf{QID} & \textbf{Answer Type} & \textbf{Demographics Var} & \textbf{Conversion Template} \\
\midrule

Occupational group (spouse or partner) & Q282 & MC & 
\specialcell{
- ``never had a job" \\ 
- ``a professional and technical job, e.g., \\doctor, teacher, engineer, artist, accountant,  \\ nurse" \\
- ``a higher administrative job, e.g., banker, \\ executive in big business, high government  \\ official, union official" \\
- ``a clerical job, e.g., secretary, clerk, \\ office manager, civil servant, bookkeeper" \\
- ``a sales job, e.g., sales manager, shop \\ owner, shop assistant, insurance agent, buyer" \\
- ``a service job, e.g., restaurant owner, \\ police officer, waitress, barber, caretaker" \\
- ``a skilled worker job, e.g., foreman, motor \\ mechanic, printer, seamstress, tool and die \\ maker, electrician" \\
- ``a semi-skilled worker job, e.g., bricklayer, \\ bus driver, cannery worker, carpenter, sheet \\ metal worker, baker" \\
- ``an unskilled worker job, e.g., labourer, \\ porter, unskilled factory worker, cleaner" \\
- ``a farm worker job, e.g., farm laborer, \\ tractor driver" \\
- ``a farm owner or farm manager job"} & I have \{var\} \\
\midrule

Sector of employment & Q284 & MC & \specialcell{- ``government or public institution" \\ - ``private business or industry" \\ - ``private non-profit organization"} & I am working for or have worked for \{var\} \\
\midrule

Chief wage earner & Q285 & MC & \specialcell{- ``I am" \\ - ``I am not"} & \{var\} the chief wage earner in my household \\
\midrule

Family savings & Q286 & MC & \specialcell{- ``was able" \\ - ``was not able"} & During the past year, my family \{var\} to save money \\
\bottomrule

\end{tabular}
\caption{Demographics dimensions, corresponding question ID (QIDs) in the original \wvs, the question type, the demographics variables, and the conversion templates for converting the raw questions from \wvs to statements in \dataset. (Part 4)}
\label{tab:demographics_4}
\end{table*}

\begin{table*}[t]
\centering
\begin{tabular}{p{2.5cm}p{0.7cm}p{1cm}lp{2.5cm}}
\toprule

\textbf{Dimension} & \textbf{QID} & \textbf{Answer Type} & \textbf{Demographics Var} & \textbf{Conversion Template} \\
\midrule

Social class (subjective) & Q287 & MC & \specialcell{- ``upper class" \\ - ``upper middle class" \\ - ``lower middle class" \\ - ``working class" \\ - ``lower class"} & I would describe myself as belonging to the \{var\} \\
\midrule

Scale of incomes & Q288R & MC & \specialcell{- ``low" \\- ``middle" \\ - ``high"} & My household is among the \{var\} income households in my country \\
\midrule

Religious denominations & Q289 & MC & \specialcell{- ``no religion or religious \\ denomination" \\ - ``the Roman Catholic religion" \\
- ``the Protestant religion" \\
- ``the Orthodox (Russian/Greek/\\etc.) religion" \\
- ``the Jewish religion" \\
- ``the Muslim religion" \\
- ``the Hindu religion" \\
- ``the Buddhist religion" \\
- ``some other Christian (Evangelical\\/Pentecostal/etc.) religion" \\
- ``some other religion or religious \\ denomination"} & I belong to \{var\} \\
\midrule

Racial belonging / ethnic group & Q290 & Code & text & I belong to the \{var\} ethnic group \\ 
\bottomrule

\end{tabular}
\caption{Demographics dimensions, corresponding question ID (QIDs) in the original \wvs, the question type, the demographics variables, and the conversion templates for converting the raw questions from \wvs to statements in \dataset. (Part 5)}
\label{tab:demographics_5}
\end{table*}

%% file: tables/example_converted_statements.tex
\begin{table*}[ht]
\centering
\begin{tabular}{p{1cm}p{6.2cm}p{5.8cm}}
\toprule

\textbf{QID} & \textbf{\Databinary} & \textbf{\Datarefined} \\
\midrule

Q51 & \specialcell{- My family and I have \textbf{often or} \\ {sometimes} gone without enough food \\ to eat \\ - My family and I have \textbf{rarely or never} \\ gone without enough food to eat} & \specialcell{- My family and I have \textbf{often} \\ gone without enough food to eat \\
- My family and I have \textbf{sometimes} \\ gone without enough food to eat \\
- My family and I have \textbf{rarely} \\ gone without enough food to eat \\
- My family and I have \textbf{never} \\ gone without enough food to eat} \\
\midrule

Q142 & \specialcell{- \textbf{I worry} about losing my \\ job or not finding a job \\
- \textbf{I'm not worried} about losing \\ my job or not finding a job \\
} & \specialcell{- \textbf{I very much worry} about losing \\ my job or not finding a job \\
- \textbf{I worry a good deal} about losing \\ my job or not finding a job \\
- \textbf{I'm not much worried} about losing \\ myjob or not finding a job \\
- \textbf{I'm not at all worried} about losing \\ my job or not finding a job
} \\
\midrule

Q253 & \specialcell{- My country \textbf{is} respectful for \\ individual human rights nowadays \\ 
- My country \textbf{is not} respectful for \\ individual human rights nowadays } & \specialcell{- My country has \textbf{a great deal of} respect \\ for individual human rights nowadays \\

- My country has \textbf{fairly much} respect for \\ individual human rights nowadays \\

- My country has \textbf{not much} respect for \\ individual human rights nowadays \\

- My country has \textbf{no respect at all} for \\ individual human rights nowadays
} \\

\midrule

Q171 & \specialcell{
- Apart from weddings and funerals, \\ I \textbf{often} attend religious services \\
- Apart from weddings and funerals, \\ I \textbf{do not often} attend religious services \\
- Apart from weddings and funerals, I \\ \textbf{never or practically never} attend \\ religious services
} & \specialcell{- Apart from weddings and funerals, \\ I attend religious services \textbf{more than}\\ \textbf{once a week} \\
- Apart from weddings and funerals, \\ I attend religious services \textbf{once a week} \\
- Apart from weddings and funerals, \\ I attend religious services \textbf{once a month} \\
- Apart from weddings and funerals, \\ I attend religious services \textbf{only on} \\ \textbf{special holy days} \\
- Apart from weddings and funerals, \\ I attend religious services \textbf{once a year} \\
- Apart from weddings and funerals, \\ I attend religious services \textbf{less often} \\
- Apart from weddings and funerals, \\ I \textbf{never or practically never} attend \\ religious services} \\

\bottomrule

\end{tabular}
\caption{Example converted value-describing statements in \dataset.}
\label{tab:example_converted_statements}

\end{table*}

%% file: sections/appendix/probing.tex
\section{Probing Off-the-Shelf Language Models with \dataset}
\label{asec:probing}

\subsection{Probing Setups}
\label{assec:probing_setups}


\paragraph{Probing models.} We consider a list of representative state-of-the-art instruction-tuned language models with different sizes and from different model families in our probing experiment. Since the demonstration statements have long sequence lengths (200 demonstration value-expressing statements combined with the probing instruction/template requires the model to have $>8$k of context window), we also pick models that do support long context window length. We consider both open-source (Llama-3.1-8B-Instruct\footnote{We refer to Llama-3.1-8B-Instruct by Llama-3.1-8B in this paper to save space.}, Llama-3.1-70B-Instruct, Mixtral-8x7B, Mixtral-8x22B, Qwen2-72B) and closed-source (GPT-4o, GPT-4o-mini, GPT-4-turbo, Claude-3.5-sonnet) models for holistic understanding of different model families. Figure \ref{fig:model_comparisons} shows the comparisons of all models with the \dataset probing setups.

\input{tables/probing_splits}

\input{tables/lm_probing_prompt}

\subsection{Probing Results}
\label{assec:probing_results}

\paragraph{\Datarefined vs. \Databinary value-expressing statements.}
We experiment with using \datarefined value-expressing statements (\eg ``I \textit{strongly} agree...'' vs. ``I \textit{somewhat} agree...'') instead of \databinary statements (\eg ``I \textit{agree}...'' vs. ``I \textit{disagree}...'') as demonstrations to LMs. Table \ref{tab:refined_vs_not_refined} shows that \datarefined statements prove more effective in aiding language models to make predictions, underscoring the importance of precise and nuanced value expressions.

\paragraph{Probing results broken down by three probe setups.}
Table \ref{tab:lm_probe_breakdown} shows the results of the probing experiments under the \databinary evaluation scheme broken down by the three probing sets, corresponding to the main probe results in Figure \ref{fig:model_comparisons}.

\paragraph{Breakdown \evenshort scores of all probed models.}
Full results of \evenshort of all probed models per each of the considered demographics dimensions are shown in Table 
\ref{tab:evenness_indices_full}.

\begin{figure*}[ht]
    \centering
    \includegraphics[width=0.9\textwidth]{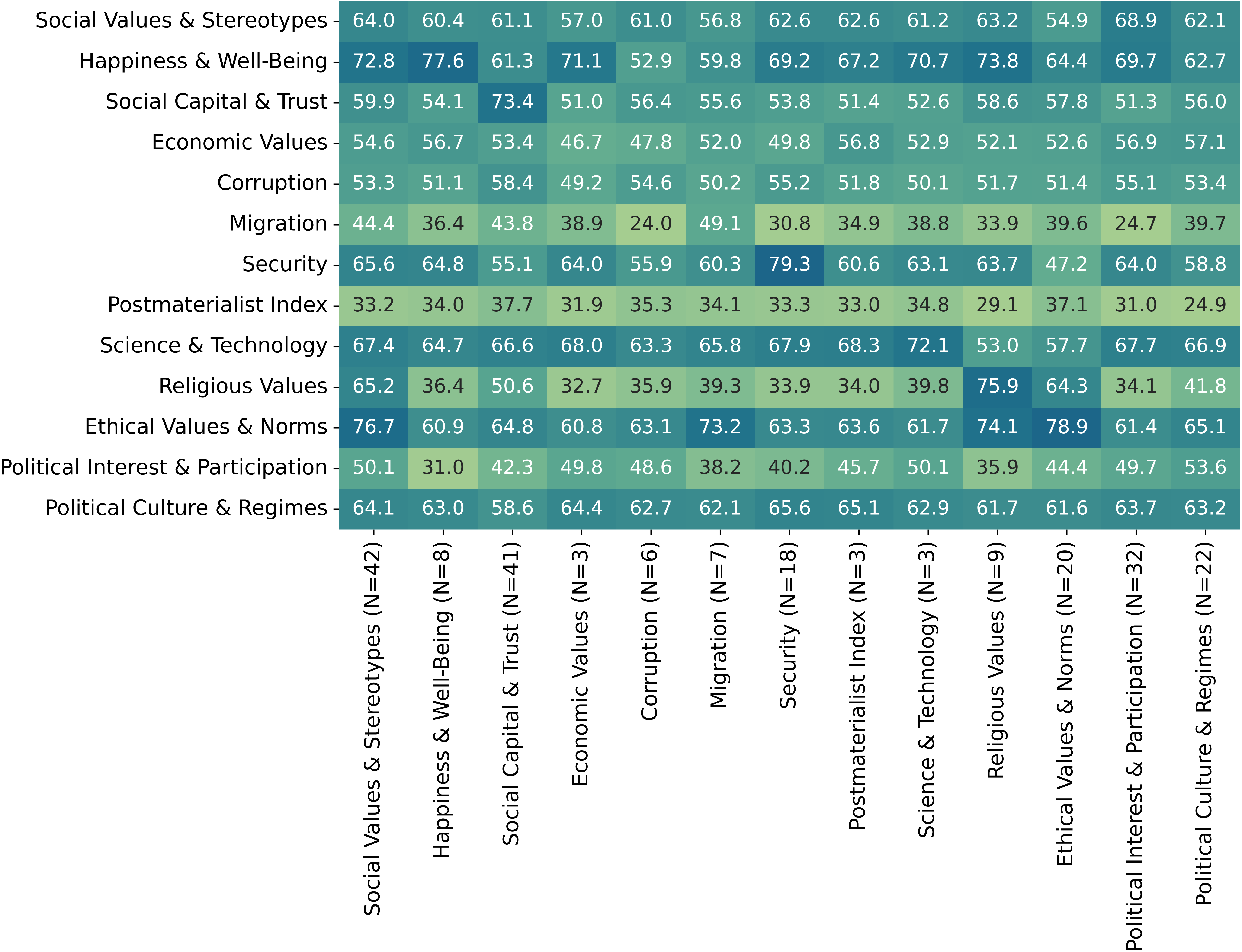}
    \caption{Results across statement categories of providing GPT-4o with different categories of demonstration examples.}
    \label{fig:probe_by_subcategory}
\end{figure*}

\paragraph{How do different types of statement influence the prediction of the other types?}
Figure \ref{fig:probe_by_subcategory} illustrates how using different categories of value statements as demonstrations affects the prediction of other categories. Our results indicate that value statements are not limited to strongly predicting only within their own category; sometimes, other categories can perform surprisingly well in predicting different types of value choices. This finding highlights intriguing dynamics and connections between various categories of value statements.

\begin{figure*}[ht]
    \centering
    \includegraphics[width=0.75\textwidth]{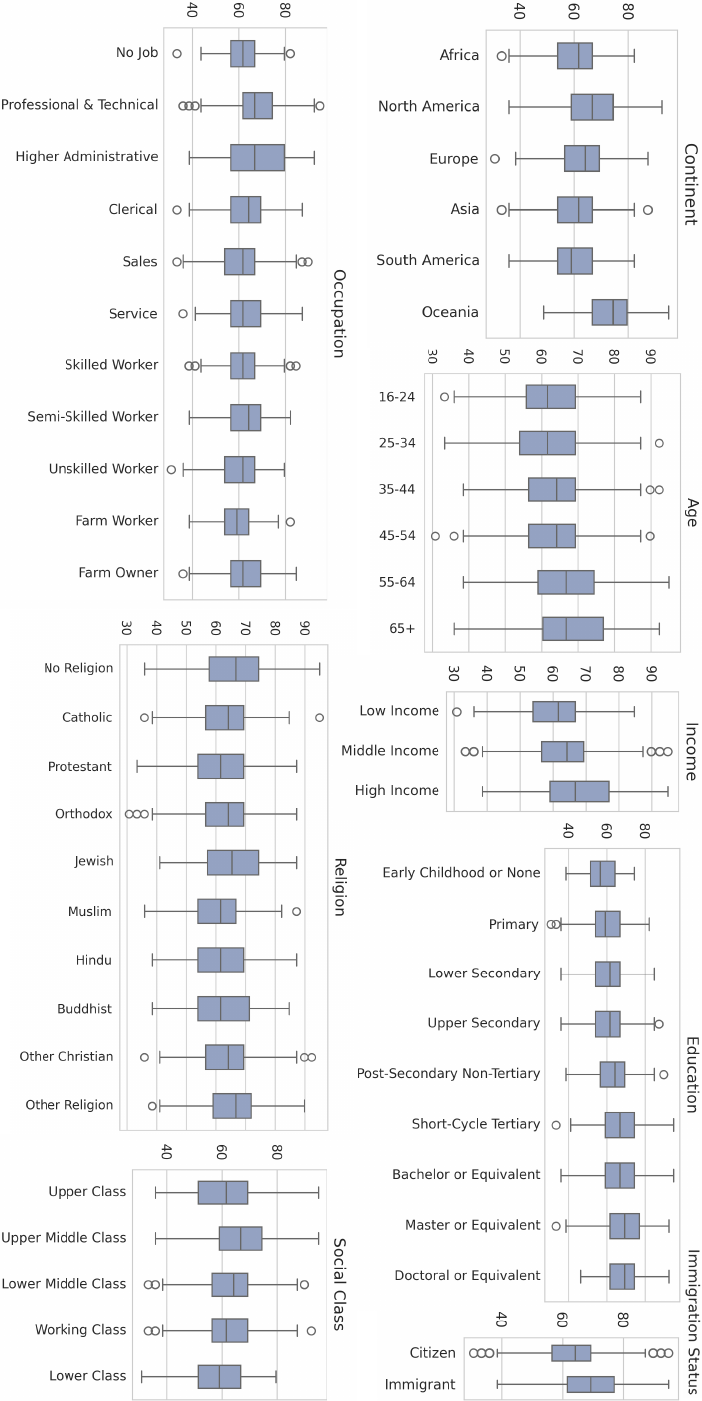}
    \caption{GPT-4o (0806) shows uneven performance within subgroups broken down by different demographics dimensions.}
    \label{fig:demographics_breakdown_full}
\end{figure*}

\paragraph{The uneven individualistic value reasoning ability of GPT-4o across demographics groups.}
Figure \ref{fig:demographics_breakdown_full} shows the performance disparity across demographic groups of different demographic dimensions.

\begin{figure*}[ht]
    \centering
    \includegraphics[width=0.65\textwidth]{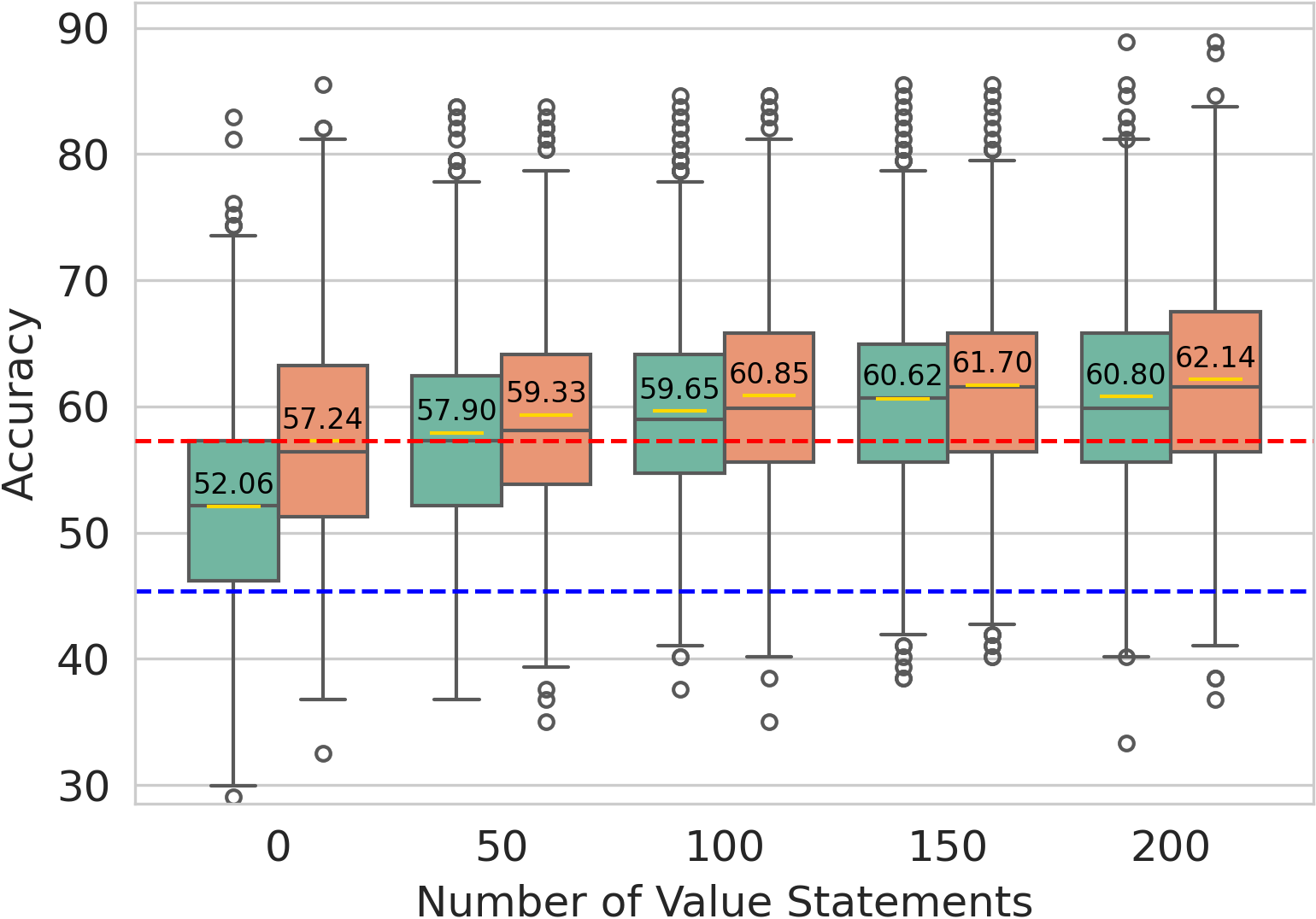}
    \caption{The effect of different numbers of demonstration statements, and with or without demographics statements on GPT-4o-mini's performance measured by \dataset.}
    \label{fig:probe_ablations_mini}
\end{figure*}

\paragraph{How do demographic statements impact weak models like GPT-4o-mini in individualistic value reasoning?}
Figure \ref{fig:probe_ablations_mini} compares probing setups with and without demographic information with GPT-4o-mini. For such a weaker model, including demographics leads to significantly better predictions than providing value statements alone, as the model is likely to struggle to interpret nuanced descriptive value statements compared to direct demographic identity deceleration.

\input{tables/refined_vs_not_refined}

\input{tables/lm_probe_breakdown}

\input{tables/evenness_indices_full}

%% file: tables/probing_splits.tex
\begin{table*}[b!]
\centering
\small
\begin{tabular}{l|l|l|l}
\toprule

\textbf{Question Category} & \multicolumn{1}{c|}{\textbf{Probe 1}} & \multicolumn{1}{c|}{\textbf{Probe 2}} & \multicolumn{1}{c}{\textbf{Probe 3}} \\ 
\midrule

Social Values, Attitudes \& Stereotypes & 1, 2, 3 & 4, 5, 6 & 7, 8, 9 \\
Happiness and Well-Being & 46, 47, 48 & 49, 50, 51 & 52, 53, 54 \\
Social Capital, Trust \& Organizational Membership & 57, 58, 59 & 60, 61, 62 & 63, 64, 65 \\ 
Economic Values & 106, 107, 108 & 109, 110, 111 & 106, 107, 108 \\ 
Corruption & 112, 113, 114 & 115, 116, 117 & 118, 119, 120 \\
Migration & 121, 122, 123 & 124, 125, 126 & 127, 128, 129 \\ 
Security & 131, 132, 133 & 134, 135, 136 & 137, 138, 139 \\ 
Postmaterialist Index & 152, 153, 154 & 155, 156, 157 & 152, 153, 154 \\ 
Science \& Technology & 158, 159, 160 & 161, 162, 163 & 158, 159, 160 \\ 
Religious Values & 164, 165, 166 & 167, 168, 169 & 170, 171, 172 \\ 
Ethical Values and Norms & 176, 177, 178 & 179, 180, 181 & 182, 183, 184 \\ 
Political Interest \& Political Participation & 199, 200, 201 & 202, 203, 204 & 205, 206, 207 \\ 
Political Culture \& Political Regimes & 235, 236, 237 & 238, 239, 240 & 241, 242, 243 \\ 
\midrule

\textbf{Total \# Probing Questions} & \multicolumn{3}{c}{\textbf{39}} \\
\bottomrule
\end{tabular}
\caption{World Value Survey question IDs (QIDs) of the three cross-validation probing setups.}
\label{tab:probing_question_splits}

\end{table*}

%% file: tables/lm_probing_prompt.tex
\begin{figure*}[ht]
\centering
\begin{tcolorbox}[colback=white,colframe=black!75!white,colbacktitle=black!75!white,width=1.0\textwidth,title=Prompt for 
Evaluating LMs' Capability for Reasoning about Individualistic Human Values]
You are an assistant helping researchers analyze an individual's value system. You will be provided with a list of statements that reflect a person's values and preferences. Your task is to interpret these statements to understand the person's underlying value system and use this understanding to predict their likely responses to additional statements.
Instructions: 

1. Review Known Statements: You will first receive a list of known statements from Person A. These statements illustrate Person A's values and preferences. Examples of such statements include:

\# I somewhat trust people I meet for the first time. \\
\# I disagree that work is a duty towards society. \\
\# I disagree that adult children have the duty to provide long-term care for their parents. \\
\# It's especially important to encourage children to learn a sense of responsibility at home. \\

This is the format of known statements that you will see:

[Known Statements of Person A]:
\begin{verbatim}
# known statement 1
# known statement 2
# known statement 3 ...
\end{verbatim}

2. Analyze and Predict: After reviewing the known statements, you will be presented with several groups of new statements. For each group, your task is to select the one statement that you believe Person A is most likely to agree with or express. Only one statement should be selected per group. 

This is the format of new statement groups that you will see:

[New Groups of Statements]:
\begin{verbatim}
{"new statement group 1 (NSG1)": [
    {"NSG1_s1": "statement 1 in NSG1"},
    {"NSG1_s2": "statement 2 in NSG1"},
    {"NSG1_s3": "statement 3 in NSG1"},
    ...],
 "new statement group 2 (NSG2)": [
    {"NSG2_s1": "statement 1 in NSG2"},
    {"NSG2_s2": "statement 2 in NSG2"},
    {"NSG2_s3": "statement 3 in NSG2"},
    ...], ...}
\end{verbatim}

3. Format Your Response: Please provide your response in the following format:

[Your Response]:
\begin{verbatim}
{"NSG1": {
    "rationale": "reason of why you choose NSG1_s2",
    "choice": "NSG1_s2"}
 "NSG2": {
    "rationale": "reason of why you choose NSG2_s1",
    "choice": "NSG2_s1"} ...}
\end{verbatim}

Now, let’s begin the task! Make sure to follow the format requirement. Only reply with the dictionary; do not include any other text; use double quotes for all string values.

[Known Statements of Person A]:
\begin{verbatim}
{known_statements}
\end{verbatim}

[New Groups of Statements]:
\begin{verbatim}
{new_statement_groups}
\end{verbatim}
[Your Response]:
\end{tcolorbox}

\end{figure*}

%% file: tables/refined_vs_not_refined.tex
\begin{table*}[t]
    \centering
    \begin{tabular}{l@{\hspace{0.95\tabcolsep}}
    l@{\hspace{0.95\tabcolsep}}
    l@{\hspace{0.95\tabcolsep}}
    l@{\hspace{0.95\tabcolsep}}
    l@{\hspace{0.95\tabcolsep}}}
    \toprule
    
    \textbf{Demo.}    & \textbf{Probe 0} & \textbf{Probe 1} & \textbf{Probe 2} & \textbf{Average} \\
    \midrule
    
    \textbf{\Datarefined} & 64.96 & \textbf{64.97} & \textbf{60.91} & \textbf{63.61} \\
    \textbf{\Databinary} & \textbf{65.21} & 64.77 & 60.39 & 63.46 \\
    
    \bottomrule
    \end{tabular}
    \vspace{-0.1cm}
    \captionof{table}{Comparing using \textit{\datarefined} and \textit{\databinary} forms of statements as value demonstrations, and evaluate with \textit{\databinary} probing statements. \datarefined are more informative for reconstructing one's value preferences compared to \databinary statements.
    }
    \label{tab:refined_vs_not_refined}
\end{table*}

%% file: tables/lm_probe_breakdown.tex
\begin{table*}[h]
\centering
\begin{tabular}{llllr}
\toprule
\textbf{Model} & \textbf{Probe 1}       & \textbf{Probe 2} & \textbf{Probe 3} & \textbf{Overall} \\
\midrule

GPT-4o (0806)       & 65.21 & 64.77 & 60.39 & 63.46                                             \\
GPT-4-turbo (0409)  & 65.08 & 65.73 & 60.41 & 63.74                                             \\
GPT-4o (0513)       & 65.66 & 64.85 & 60.61 & 63.71                                             \\
GPT-4o-mini (0718)  & 60.05 & 64.13 & 58.21 & 60.80                                             \\
LLama-3.1-8B        & 58.72 & 62.09 & 53.80 & 58.20                                             \\
LLama-3.1-70B       & 65.41 & 66.53 & 59.20 & 63.71                                             \\
Mixtral-8x7B        & 59.18 & 58.03 & 51.58 & 56.26                                             \\
Mixtral-8x22B       & 62.91 & 63.47 & 57.10 & 61.16                                             \\
Qwen2-72B           & 65.10 & 65.16 & 60.58 & 63.61                                             \\
Claude-3.5 (Sonnet) & 65.74 & 66.48 & 61.76 & 64.66 \\

\bottomrule

\end{tabular}
\caption{Main probing results with the \databinary evaluation setup of all models, broken down by three probing setups.}
\label{tab:lm_probe_breakdown}
\end{table*}

%% file: tables/evenness_indices_full.tex
\begin{table*}[t]
\small
\centering
\begin{tabular}{l@{\hspace{0.7\tabcolsep}}
r@{\hspace{0.9\tabcolsep}}
r@{\hspace{0.9\tabcolsep}}
r@{\hspace{0.9\tabcolsep}}
r@{\hspace{0.9\tabcolsep}}
r@{\hspace{0.9\tabcolsep}}
r@{\hspace{0.9\tabcolsep}}
r@{\hspace{0.9\tabcolsep}}
r@{\hspace{0.9\tabcolsep}}
r@{\hspace{0.9\tabcolsep}}
r@{\hspace{0.9\tabcolsep}}}

\toprule

\textbf{Dimension} & \specialcell{LLama\\-3.1\\-8B} & \specialcell{GPT-4o\\(0806)} & \specialcell{GPT-4\\-turbo\\(0409)} & \specialcell{GPT-4o\\(0513)} & \specialcell{GPT-4o\\-mini\\(0718)} & \specialcell{LLama\\-3.1\\-70B} & \specialcell{Mixtral\\-8x7B} & \specialcell{Mixtral\\-8x22B} & \specialcell{Qwen2\\-72B} & \specialcell{Claude\\-3.5\\(Sonnet)} \\

\midrule

Country            & 3.47                                      & 3.97                                       & 3.79                                            & 3.88                                       & 3.67                                            & 2.94                                       & 4.14                                      & 3.98                                       & 4.24                                   & 4.14                                             \\
Continent          & 5.55                                      & 5.67                                       & 5.43                                            & 5.37                                       & 5.09                                            & 3.85                                       & 5.64                                      & 5.95                                       & 5.85                                   & 5.72                                             \\
Sex                & 0.98                                      & 0.50                                       & 0.27                                            & 0.52                                       & 0.42                                            & 0.14                                       & 0.45                                      & 0.54                                       & 0.35                                   & 0.18                                             \\
Age                & 2.33                                      & 2.31                                       & 2.17                                            & 2.13                                       & 2.18                                            & 1.36                                       & 2.18                                      & 2.50                                       & 2.63                                   & 2.19                                             \\
Immigration Status & 4.58                                      & 4.62                                       & 4.22                                            & 4.41                                       & 4.20                                            & 2.90                                       & 4.29                                      & 5.04                                       & 4.54                                   & 4.71                                             \\
Birth Country      & 4.96                                      & 5.10                                       & 4.74                                            & 4.92                                       & 4.50                                            & 3.63                                       & 6.23                                      & 5.86                                       & 5.49                                   & 5.43                                             \\
Citizenship        & 2.44                                      & 3.22                                       & 3.48                                            & 2.92                                       & 2.51                                            & 0.38                                       & 3.97                                      & 2.87                                       & 4.16                                   & 4.18                                             \\
Marital Status     & 1.10                                      & 1.36                                       & 1.55                                            & 1.39                                       & 0.97                                            & 0.58                                       & 1.45                                      & 1.47                                       & 1.86                                   & 1.95                                             \\
Education          & 3.73                                      & 4.06                                       & 3.31                                            & 3.69                                       & 2.87                                            & 2.92                                       & 4.37                                      & 3.39                                       & 3.98                                   & 3.81                                             \\
Employment Status  & 2.73                                      & 2.65                                       & 2.53                                            & 2.62                                       & 2.07                                            & 1.54                                       & 2.76                                      & 2.58                                       & 2.66                                   & 2.77                                             \\
Occupation         & 2.44                                      & 2.66                                       & 2.29                                            & 2.48                                       & 2.19                                            & 1.90                                       & 2.47                                      & 2.58                                       & 2.69                                   & 2.66                                             \\
Employment Sector  & 1.19                                      & 1.33                                       & 1.01                                            & 1.08                                       & 1.07                                            & 0.92                                       & 1.10                                      & 0.78                                       & 1.24                                   & 1.05                                             \\
Family Saving      & 3.23                                      & 3.18                                       & 3.06                                            & 2.99                                       & 2.73                                            & 2.04                                       & 3.09                                      & 3.25                                       & 3.51                                   & 3.22                                             \\
Social Class       & 2.97                                      & 2.83                                       & 2.50                                            & 2.57                                       & 1.95                                            & 1.96                                       & 2.86                                      & 2.75                                       & 2.78                                   & 2.99                                             \\
Income             & 4.05                                      & 3.39                                       & 2.94                                            & 3.33                                       & 2.65                                            & 2.68                                       & 3.99                                      & 3.58                                       & 3.80                                   & 3.57                                             \\
Religion           & 1.76                                      & 1.69                                       & 1.95                                            & 1.66                                       & 1.77                                            & 1.30                                       & 2.02                                      & 1.87                                       & 2.09                                   & 1.73                                             \\

\midrule
\textbf{Average}   & 2.97                                      & 3.03                                       & 2.83                                            & 2.87                                       & 2.55                                            & 1.94                                       & 3.19                                      & 3.06                                       & 3.24                                   & 3.14 \\           
\bottomrule           
\end{tabular}
\caption{The \even (\evenshort) of models by demographic dimensions.}
\label{tab:evenness_indices_full}
\end{table*}

%% file: sections/appendix/finetune.tex
\section{Details of the \finetune}
\label{asec:finetune_details}

\subsection{Training Setups}
\label{assec:finetune_setups}

To train the \finetuneshort, we sequentially finetune the \llamathreeone using the  \href{https://github.com/allenai/open-instruct}{Open-Instruct codebase}. All models are trained on a single node of 8 NVIDIA H100 80GB HBM3 GPUs. Table \ref{tab:training_hyper_parameters} includes particular hyperparameters we adopt in our experiments. Training on 1 batch of training data takes roughly 0.9 seconds. All evaluations use the checkpoint at the end of epoch 2.

\input{tables/training_hyper_parameters}

Table \ref{tab:training_config_baselines} shows the detailed specification of baselines and \finetuneshort variations used in Table \ref{tab:improved_reasoning_results_full} of the main paper.

Below is an example of training data for the \finetuneshort.

\input{tables/finetune_prompt}

\input{tables/reasoner_data_composition}

\subsection{Additional Results}
\label{assec:finetune_results}

Table \ref{tab:variances_demographics_dimensions} shows the comparison of \evenshort between zero-shot \llamathreeone vs. trained \finetuneshort across varied demographics dimensions. Figure \ref{fig:evenness_age}-\ref{fig:evenness_sex} show a breakdown of the relative performance improvement of \finetuneshort compared to zero-short \llamathreeone for each demographics category within different demographic dimensions.

\input{tables/improved_reasoning_results_full}

\input{tables/variances_demographics_dimensions}

\input{tables/variances_demographics_dimensions_full}

\input{tables/resemble_cluster}

%% file: tables/training_hyper_parameters.tex
\begin{table*}[t!]
  \centering
  \small
  \begin{tabular}{l  l}
    \toprule
    Base Model & \verb|meta-llama/Meta-Llama-3.1-8B-Instruct| \\
    Precision & BFloat16 \\
    Epochs & 2 \\
    Weight decay & 0 \\
    Warmup ratio & 0.03 \\
    Learning rate & 5e-6 \\
    Learning rate scheduler & linear \\
    Max. seq. length & 4096 \\
    Batch size & 8 \\
    \bottomrule
  \end{tabular}
  \caption{Hyperparameters used for training the \finetuneshort.}
\label{tab:training_hyper_parameters}
\end{table*}

%% file: tables/finetune_prompt.tex
\begin{figure*}[ht]
    \centering
    \begin{tcolorbox}[colback=white,colframe=black!75!white,colbacktitle=black!75!white,title=An Example Training Data for the \finetune]
    You will first receive a list of known statements from Person A, illustrating Person A's values and preferences. You will then be presented with a group of new statements. Your task is to select the one statement you believe Person A is most likely to agree with or express.
    \\
    
    [Known statements]:
    \begin{verbatim}
    # I am not an active member of any women's group
    # I believe in hell
    # I do not have confidence in banks
    # I believe that suicide is not justifiable
    # I do not trust people I meet for the first time
    # I would not like to have drug addicts as neighbors
    # Friends are important in my life
    \end{verbatim}
    [New statements options]:
    \begin{verbatim}
    Option 1: I believe that claiming government benefits to 
    which you are not entitled is not justifiable
    
    Option 2: I believe that claiming government benefits
    to which you are not entitled is justifiable
    \end{verbatim}
    [Person A most likely agrees with]:
    \tcblower
    \begin{verbatim}
    Option 2: I believe that claiming government benefits
    to which you are not entitled is justifiable
    \end{verbatim}
    \end{tcolorbox}
\end{figure*}

%% file: tables/reasoner_data_composition.tex
\begin{table*}[t!]
  \centering
  \small
  \begin{tabular}{p{6cm} p{7.2cm}}
  \toprule
  \textbf{Model or Baseline} & \textbf{Details} \\
  \midrule

\texttt{Random} & Randomly selecting a candidate statement choice. \\
\texttt{Global (majority vote)} & Selecting the statement choice based on the majority vote across the entirety of $\mathbb{I}_{\text{train}}$. \\
\texttt{Resemble (top 1)} & Selecting the statement choice based on the choice of the individual who shares the most number of common demonstration statements with $I_i \in \mathbb{I}_{\text{eval}}$. \\
\texttt{Resemble (top cluster)} & Selecting the statement choice based on the majority choice among a cluster of the top $N$ individuals who shared the most number of common demonstration statements with $I_i \in \mathbb{I}_{\text{eval}}$. Since the different sizes of the cluster may result in different prediction accuracy---in general, too small or too large of the cluster can both lead to noisy prediction. Table \ref{tab:top_cluster_breakdown} shows the breakdown performance of different cluster size, $N$. We pick the best-performing setting with $N=24$ to report in Table \ref{tab:improved_reasoning_results}. \\
\midrule

\texttt{GPT-4o (no demo.)} & Giving GPT-4o no demonstration statements when predicting an individual $I_i$'s value statement selection. \\
\texttt{GPT-4o (only demographics)} & Giving GPT-4o only demographics-declaring statements when predicting an individual $I_i$'s value statement selection. \\
\texttt{GPT-4o (200 demo.)} & Giving GPT-4o 200 value-expressing statements when predicting an individual $I_i$'s value statement selection. \\
\texttt{Llama-3.1-8B (200 demo.)} & Giving \llamathreeone-Instruct 200 value-expressing statements when predicting an individual $I_i$'s value statement selection. \\
\midrule

\texttt{[probe=p,demo=mixed,N=800]} & \finetuneshort trained with a \textit{mixed} number of demonstration statements, and with probing statements in \databinary form. Each of the 253 value questions has 800 data. \\
\texttt{[probe=r,demo=mixed,N=800]} & \finetuneshort trained with a \textit{mixed} number of demonstration statements, and with probing statements in \datarefined form. Each of the 253 value questions has 800 data. \\
\texttt{[probe=p+r,demo=200,N=800]} & \finetuneshort trained with a fixed number of \textit{200} demonstration statements, and with probing statements in both \datarefined and \databinary forms. Each of the 253 value questions has 400 data for \datarefined and \databinary probing question forms, respectively, with a total of 800 data. \\
\texttt{[probe=p+r,demo=mixed,N=800]} & \finetuneshort trained with a \textit{mixed} number of demonstration statements, and with probing statements in both \datarefined and \databinary forms. Each of the 253 value questions has 400 data for \datarefined and \databinary probing question forms, respectively, with a total of 800 data. \\
\texttt{[probe=p+r,demo=mixed+200,N=800]} & \finetuneshort trained with both \textit{mixed} number of demonstration statements and a fixed number of 200 demonstration statements, and with probing statements in both \datarefined and \databinary forms. Each of the 253 value questions has 200 data for (mixed, \datarefined), (mixed, \databinary), (200, \datarefined), (200, \databinary) setups, respectively, with a total of 800 data. \\
\texttt{[probe=p+r,demo=mixed+200,N=1600]} & \finetuneshort trained with both \textit{mixed} number of demonstration statements and a fixed number of 200 demonstration statements, and with probing statements in both \datarefined and \databinary forms. Each of the 253 value questions has 400 data for (mixed, \datarefined), (mixed, \databinary), (200, \datarefined), (200, \databinary) setups, respectively, with a total of 1600 data. \\

    \bottomrule
  \end{tabular}
  \caption{Training data composition for different versions of \finetuneshort and specifications of baselines in Table \ref{tab:improved_reasoning_results}.}
\label{tab:training_config_baselines}
\end{table*}

%% file: tables/improved_reasoning_results_full.tex
\begin{table*}[t!]
\vspace{-0.4cm}
\centering
\small
\begin{tabular}{l@{\hspace{0.9\tabcolsep}}
r@{\hspace{0.95\tabcolsep}}
r@{\hspace{0.95\tabcolsep}}
r@{\hspace{0.95\tabcolsep}}
r@{\hspace{0.95\tabcolsep}}
r@{\hspace{0.95\tabcolsep}}
r@{\hspace{0.95\tabcolsep}}
r@{\hspace{0.95\tabcolsep}}
r@{\hspace{0.95\tabcolsep}}
r@{\hspace{0.95\tabcolsep}}
}

\toprule

 & \multicolumn{4}{c}{\textbf{\Databinary}} & \multicolumn{4}{c}{\textbf{\Datarefined}} & \multicolumn{1}{c}{\textbf{All}} \\ 

 \cmidrule(r){2-5}  \cmidrule(r){6-9} \cmidrule(r){10-10}

\textbf{Method} & \textbf{\scriptsize{Probe 1}} & \textbf{\scriptsize{Probe 2}} & \textbf{\scriptsize{Probe 3}} & \textbf{\scriptsize{Avg.}}  & \textbf{\scriptsize{Probe 1}} & \textbf{\scriptsize{Probe 2}} & \textbf{\scriptsize{Probe 3}} & \textbf{\scriptsize{Avg.}} & \textbf{\scriptsize{Avg.}} \\

\midrule

\scriptsize{\texttt{Random}} & 46.37 & 45.51 & 44.23 & 45.37 & 29.16 &	29.03 & 25.43 & 27.87 & 36.62 \\

\scriptsize{\texttt{Global (majority vote)}} & 66.60 & 65.98 & 62.28 & 64.95 & 49.82 & 49.08 & 47.20 & 48.70 & 56.83 \\

\scriptsize{\texttt{Resemble (1600, top 1)}} & 66.41 & 65.95 & 64.59 & 65.65 & 47.03 & 48.07 & 47.55 & 47.55 & 56.60 \\ 

\scriptsize{\texttt{Resemble (1600, top cluster)}} & 71.79 & 71.36 & 68.01 & 70.39 & 54.02 & 55.79 & 53.27 & 54.36 & 62.38 \\

\scriptsize{\texttt{Resemble (all, top 1)}} & 70.31 & 70.15 & 69.02 & 69.83 & 53.26 & 54.01 & 53.27 & 53.51 & 61.67  \\

\scriptsize{\texttt{Resemble (all, top cluster)}} 
& 74.74 & 74.87 & 71.60 & 73.73 & 59.32 & 60.78 & 58.32 & 59.47 & 66.60 \\

\midrule

\scriptsize{\texttt{GPT-4o (no demo.)}} & 58.80 & 57.60 & 47.98 & 54.79 & 35.50 & 32.92 & 30.76 & 33.06 & 43.93 \\

\scriptsize{\texttt{GPT-4o (only demographics)}} & 62.13 & 62.67 & 56.13 & 60.31 & 41.57 & 43.10 & 37.40 & 40.69 & 50.50 \\

\scriptsize{\texttt{GPT-4o (200 demo.)}} & 65.21 & 64.77 & 60.39 & 63.46 & 36.12 & 38.70 & 31.94 & 35.59 & 49.52 \\

\scriptsize{\texttt{Llama-3.1-8B
(200 demo.)}} & 53.06 & 56.16 & 53.82 & 54.34 & 35.64 & 39.32 & 38.94 & 37.97 & 46.16 \\

\midrule

\scriptsize{\texttt{[probe=p,demo=mixed,N=800]}} & 74.03 & \underline{75.45} & \underline{71.28} & \underline{73.59} & 43.22 & 48.42 & 40.61 & 44.08 & 58.84 \\

\scriptsize{\texttt{[probe=r,demo=mixed,N=800]}} & 73.23 & 75.24 & 71.27 & 73.25 & \underline{58.82} & \textbf{62.31} & \underline{58.67} & \underline{59.94} & \underline{66.59} \\

\scriptsize{\texttt{[probe=p+r,demo=200,N=800]}} & 73.96 & 75.13 & 71.25 & 73.45 & 57.52 & 61.38 & 57.61 & 58.84 & 66.14 \\

\scriptsize{\texttt{[probe=p+r,demo=mixed,N=800]}} & \underline{74.21} & 75.32 & 71.24 & \underline{73.59} & 58.27 & \underline{61.71} & 58.21 & 59.40 & 66.49 \\

\scriptsize{\texttt{[probe=p+r,demo=mixed+200,N=800]}} & \textbf{74.65} & \textbf{75.94} & \textbf{72.28} & \textbf{74.29} & \textbf{59.20} & \textbf{62.31} & \textbf{59.18} & \textbf{60.23} & \textbf{67.26} \\

\midrule

\rowcolor{cyan}  
\scriptsize{\texttt{[probe=p+r,demo=mixed+200,N=1600]}} & \textbf{75.05} & \textbf{76.42} & \textbf{72.76} & \textbf{74.74} & \textbf{59.42} & \textbf{62.68} & \textbf{59.72} & \textbf{60.60} & \textbf{67.67} \\

\bottomrule
\end{tabular}
\caption{Results of \finetuneshort models for improved individualistic value reasoning for both the \databinary and \datarefined evaluation setups. For the middle section of ablation models, the best performances are \textbf{bolded}, and the second best performances are \underline{underlined}. All results in this table are obtained by giving 200 demonstration value-expressing statements during test time.}
\label{tab:improved_reasoning_results_full}
\vspace{-0.4cm}
\end{table*}
\vspace{-0.2cm}

%% file: tables/variances_demographics_dimensions.tex
\begin{table*}[h]
\centering
\small

\begin{tabular}{l@{\hspace{0.9\tabcolsep}}
r
r
}
\toprule

\textbf{Dimension} & \texttt{0-Shot} & \texttt{probe=p+r,d=mix:200,N=1600} \finetuneshort \\
\midrule

Country            & 3.47          & \textbf{3.03} \\
Continent          & 5.55          & \textbf{3.31} \\
Sex                & 0.98          & \textbf{0.35} \\
Age                & 2.33          & \textbf{1.64} \\
Immigration Status & 4.58          & \textbf{3.28} \\
Birth Country      & 4.96          & \textbf{3.84} \\
Citizenship        & \textbf{2.44} & 3.51          \\
Marital Status     & 1.10          & \textbf{0.72} \\
Education          & 3.73          & \textbf{2.18} \\
Employment Status  & 2.73          & \textbf{2.03} \\
Occupation         & 2.44          & \textbf{1.81} \\
Employment Sector  & \textbf{1.19} & 1.34          \\
Family Saving      & 3.23          & \textbf{2.27} \\
Social Class       & 2.97          & \textbf{2.16} \\
Income             & 4.05          & \textbf{2.83} \\
Religion           & 1.76          & \textbf{1.16} \\
\midrule
\textbf{Average}   & 2.97          & \textbf{2.22} \\

\bottomrule
\end{tabular}
\caption{The \evenshort of \llamathreeone-based 0-shot and \finetuneshort performances across different demographics groups for different demographics dimensions. The lower $\sigma$, the more even performance the model is in reasoning about individualistic values across populations with different demographics groups.}
\label{tab:variances_demographics_dimensions}
\end{table*}

%% file: tables/variances_demographics_dimensions_full.tex
\begin{figure*}[ht]
    \vspace{-0.4cm}
    \centering
    \includegraphics[width=0.7\textwidth]{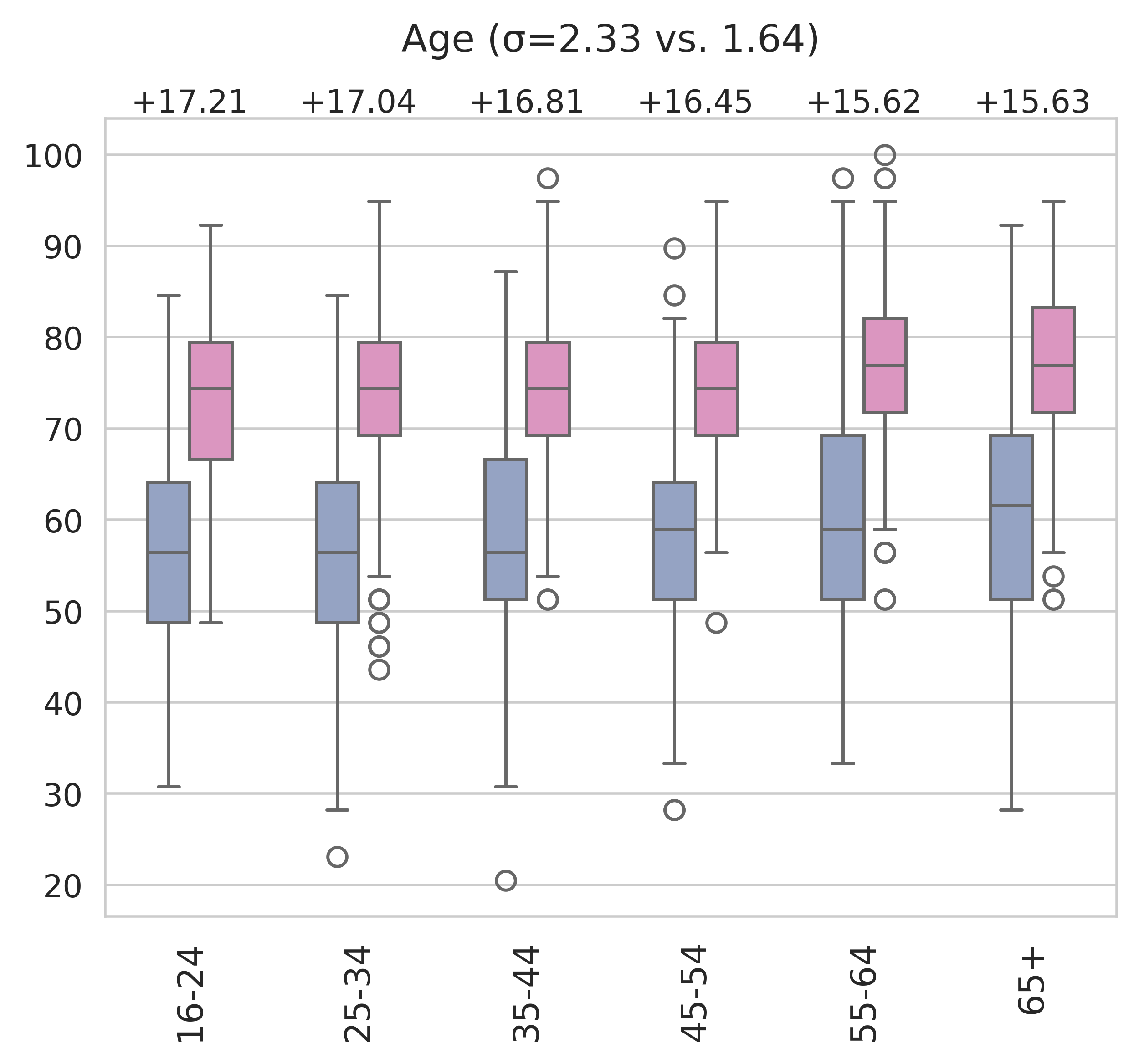}
    \captionof{figure}{The breakdown of the relative performance improvement of \finetuneshort compared to zero-short \llamathreeone for each demographics category within the \textit{Age} dimension.}
    \label{fig:evenness_age}
\end{figure*}

\begin{figure*}[ht]
    \vspace{-0.4cm}
    \centering
    \includegraphics[width=0.38\textwidth]{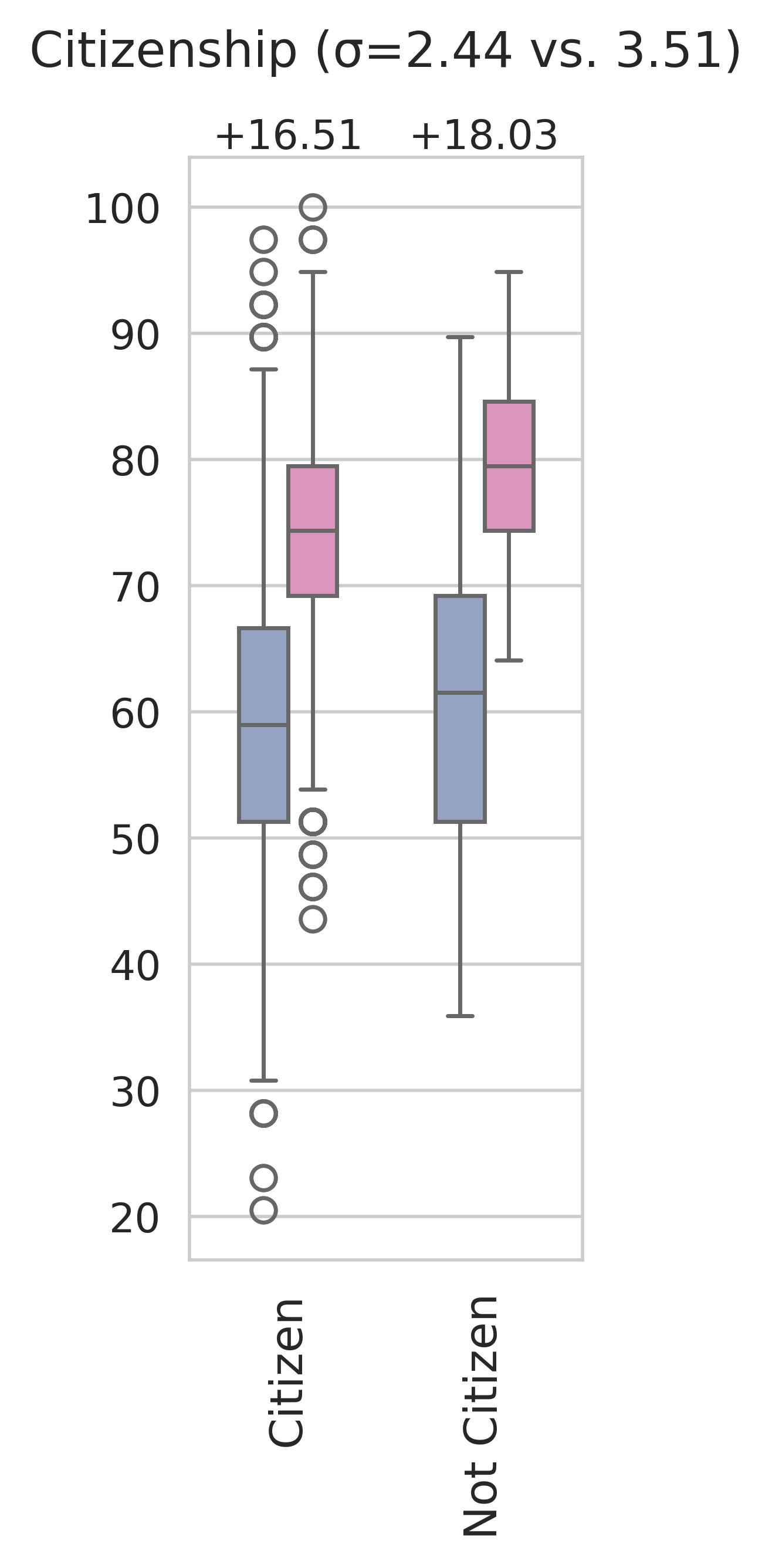}
    \captionof{figure}{The breakdown of the relative performance improvement of \finetuneshort compared to zero-short \llamathreeone for each demographics category within the \textit{Citizenship} dimension.}
    \label{fig:evenness_citizenship}
\end{figure*}

\begin{figure*}[ht]
    \vspace{-0.4cm}
    \centering
    \includegraphics[width=1\textwidth]{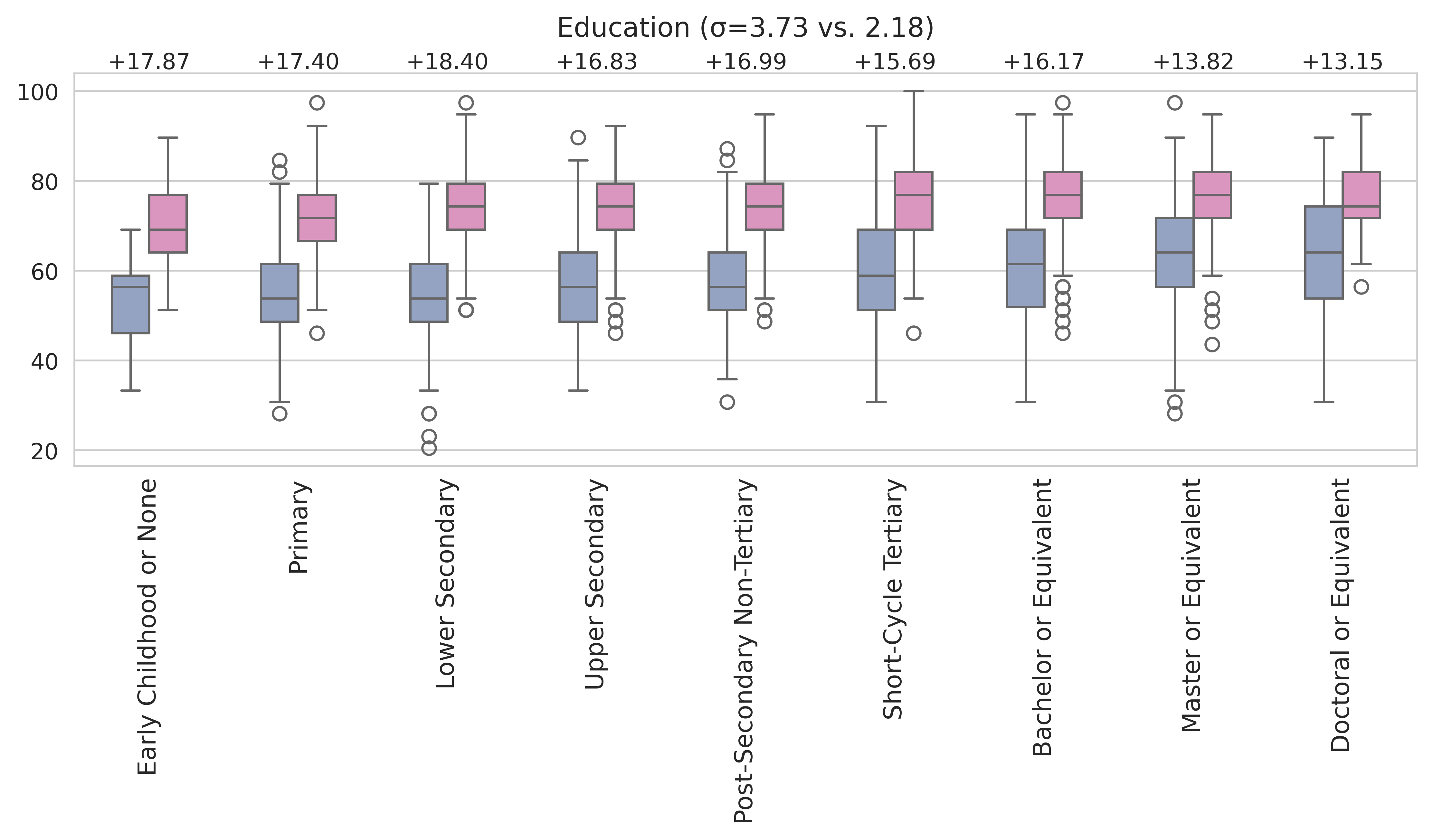}
    \captionof{figure}{The breakdown of the relative performance improvement of \finetuneshort compared to zero-short \llamathreeone for each demographics category within the \textit{Education} dimension.}
    \label{fig:evenness_education}
\end{figure*}

\begin{figure*}[ht]
    \vspace{-0.4cm}
    \centering
    \includegraphics[width=0.45\textwidth]{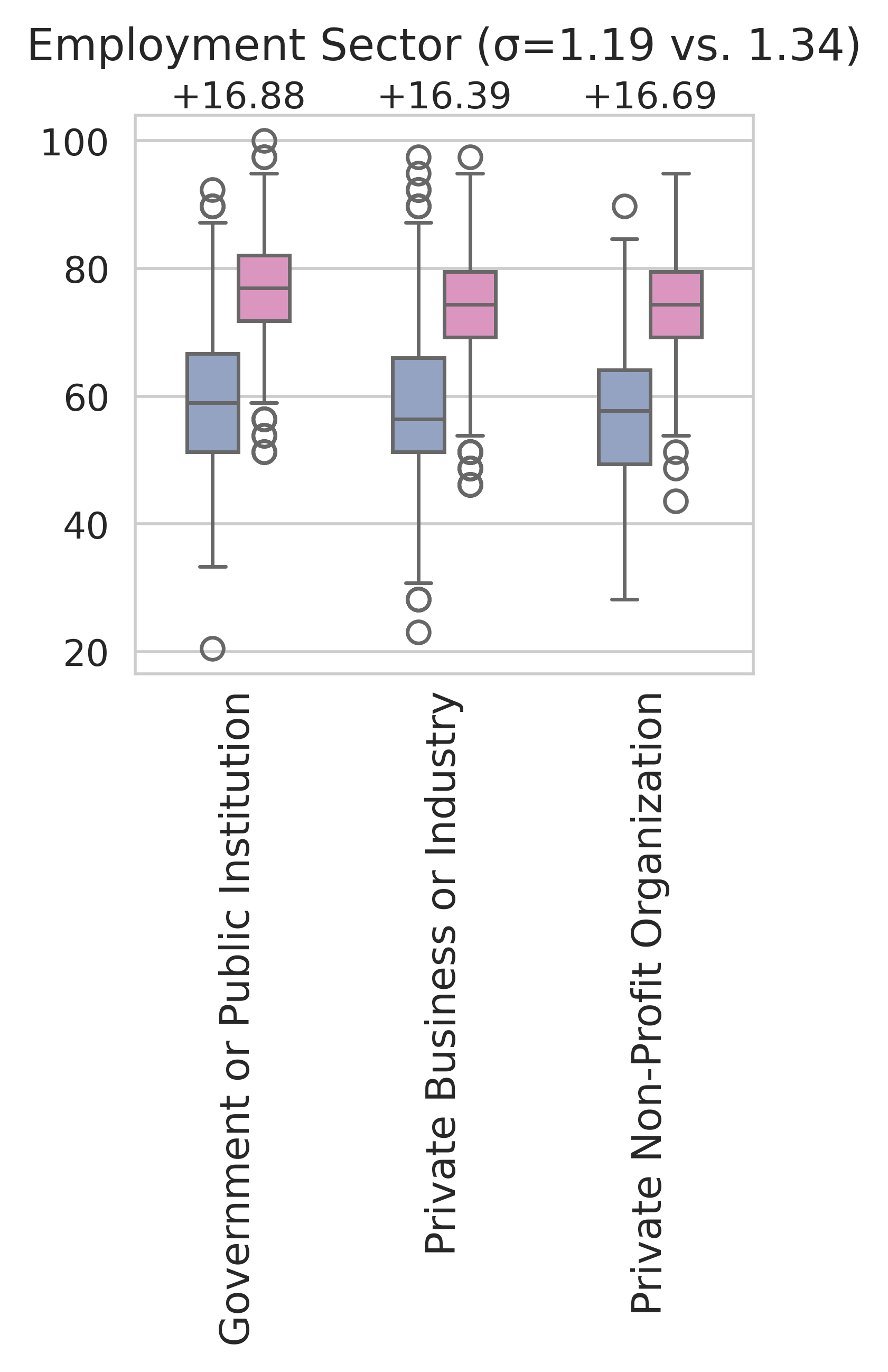}
    \captionof{figure}{The breakdown of the relative performance improvement of \finetuneshort compared to zero-short \llamathreeone for each demographics category within the \textit{Employment Sector} dimension.}
    \label{fig:evenness_employment_sector}
\end{figure*}

\begin{figure*}[ht]
    \vspace{-0.4cm}
    \centering
    \includegraphics[width=0.8\textwidth]{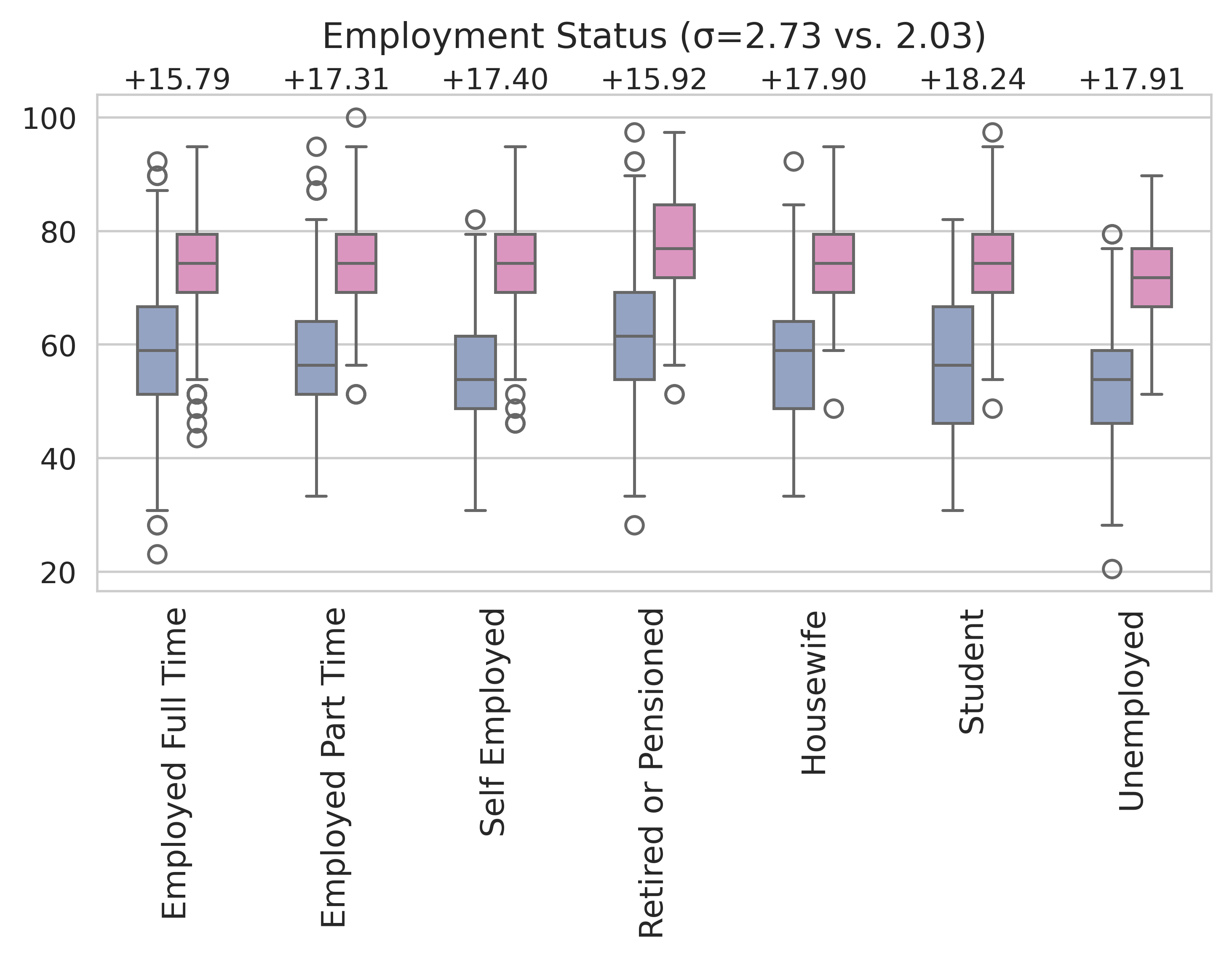}
    \captionof{figure}{The breakdown of the relative performance improvement of \finetuneshort compared to zero-short \llamathreeone for each demographics category within the \textit{Employment Status} dimension.}
    \label{fig:evenness_employment_status}
\end{figure*}

\begin{figure*}[ht]
    \vspace{-0.4cm}
    \centering
    \includegraphics[width=0.5\textwidth]{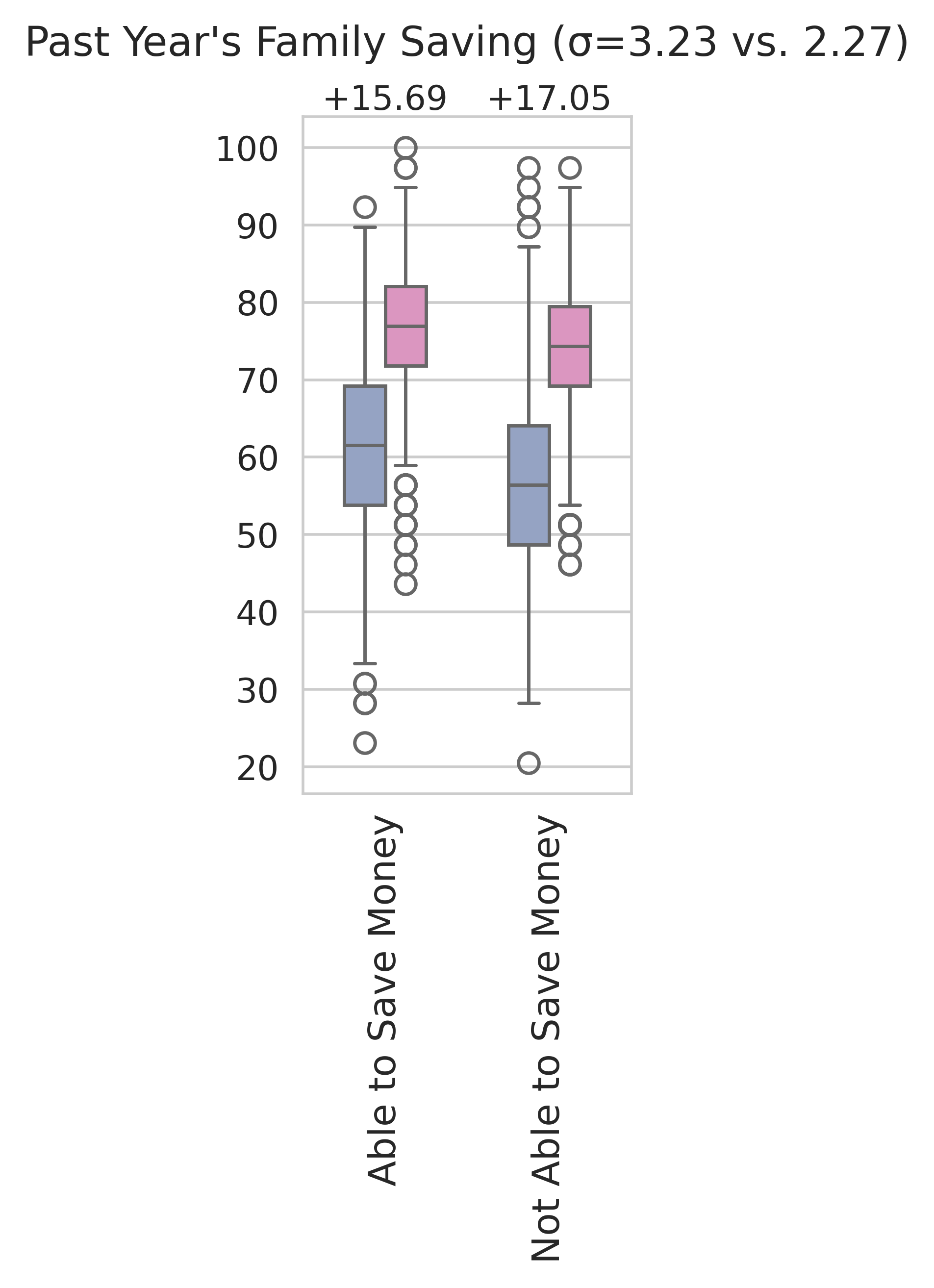}
    \captionof{figure}{The breakdown of the relative performance improvement of \finetuneshort compared to zero-short \llamathreeone for each demographics category within the \textit{Family Saving} dimension.}
    \label{fig:evenness_family_saving}
\end{figure*}

\begin{figure*}[ht]
    \vspace{-0.4cm}
    \centering
    \includegraphics[width=0.45\textwidth]{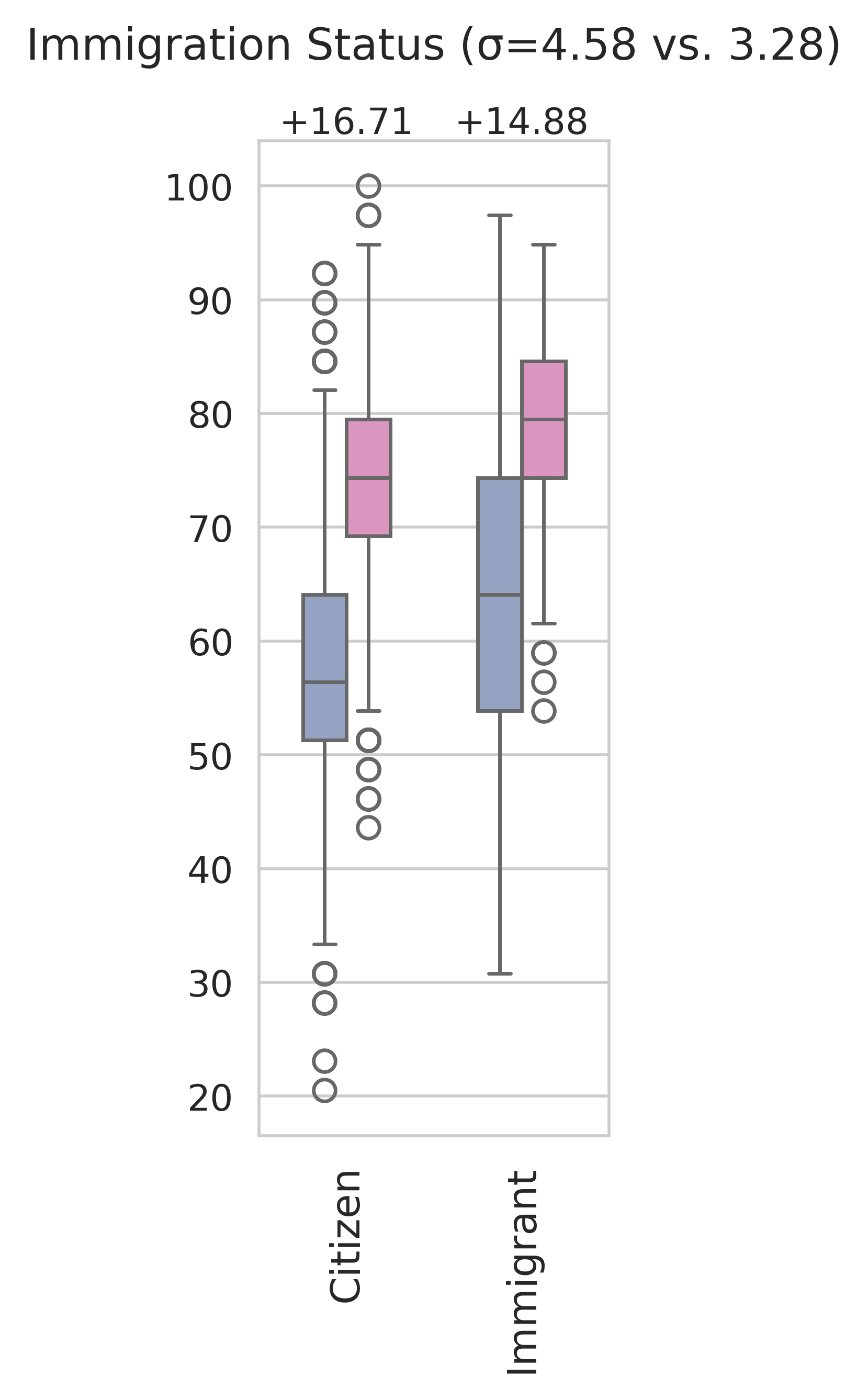}
    \captionof{figure}{The breakdown of the relative performance improvement of \finetuneshort compared to zero-short \llamathreeone for each demographics category within the \textit{Immigration Status} dimension.}
    \label{fig:evenness_immigration_status}
\end{figure*}

\begin{figure*}[ht]
    \vspace{-0.4cm}
    \centering
    \includegraphics[width=0.7\textwidth]{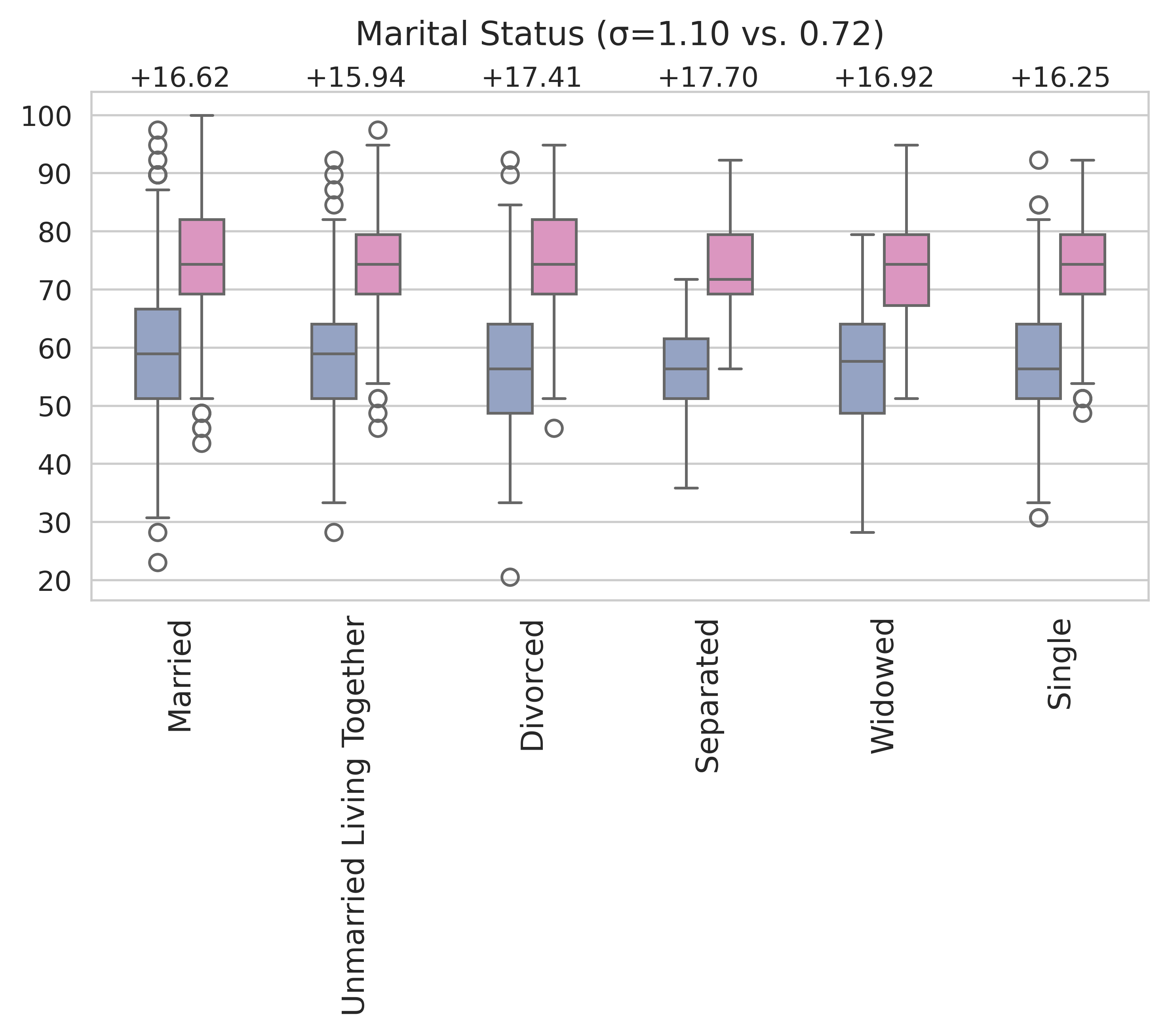}
    \captionof{figure}{The breakdown of the relative performance improvement of \finetuneshort compared to zero-short \llamathreeone for each demographics category within the \textit{Marital Status} dimension.}
    \label{fig:evenness_marital_status}
\end{figure*}

\begin{figure*}[ht]
    \vspace{-0.4cm}
    \centering
    \includegraphics[width=1\textwidth]{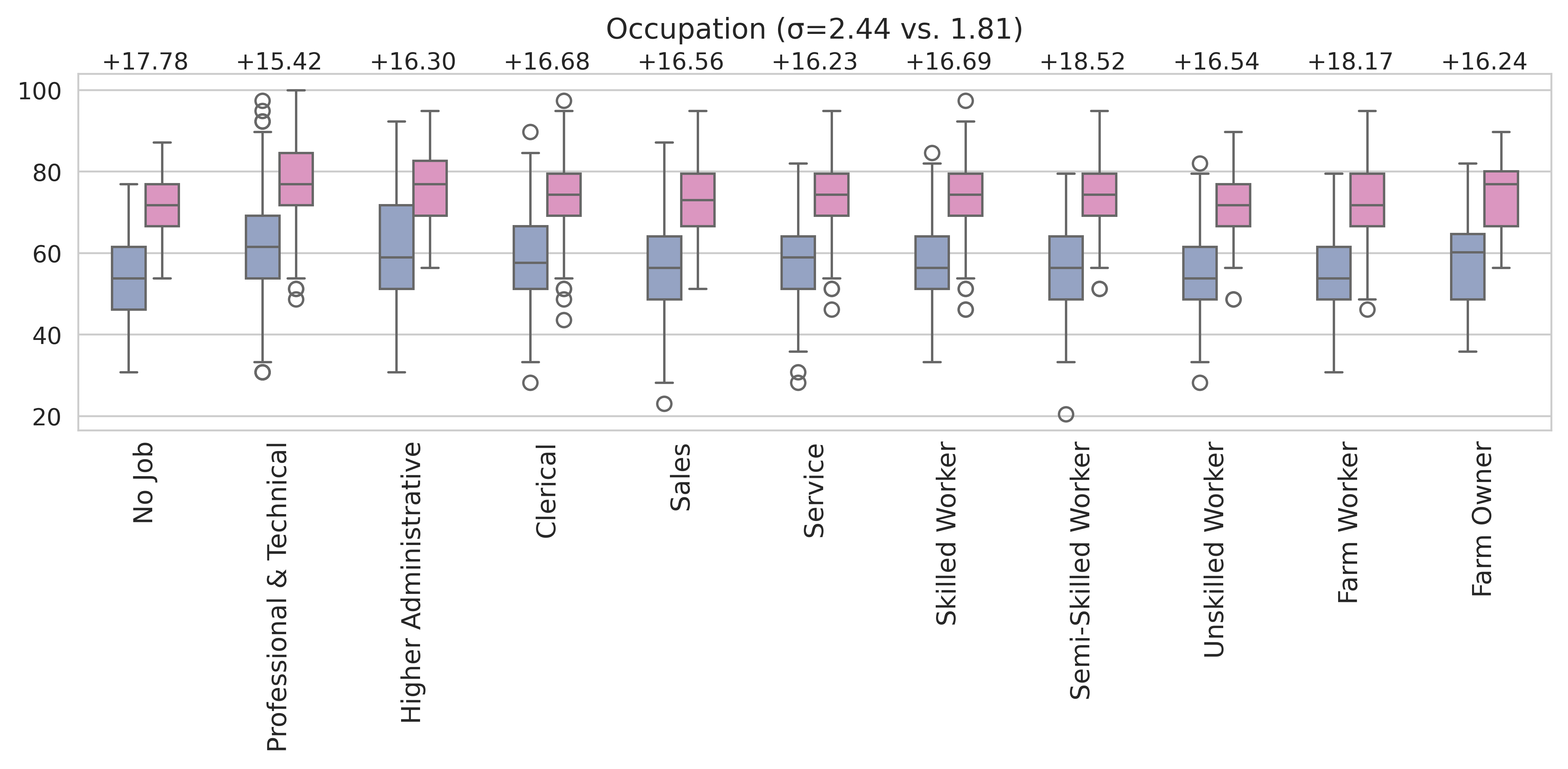}
    \captionof{figure}{The breakdown of the relative performance improvement of \finetuneshort compared to zero-short \llamathreeone for each demographics category within the \textit{Occupation} dimension.}
    \label{fig:evenness_occupation}
\end{figure*}

\begin{figure*}[ht]
    \vspace{-0.4cm}
    \centering
    \includegraphics[width=1\textwidth]{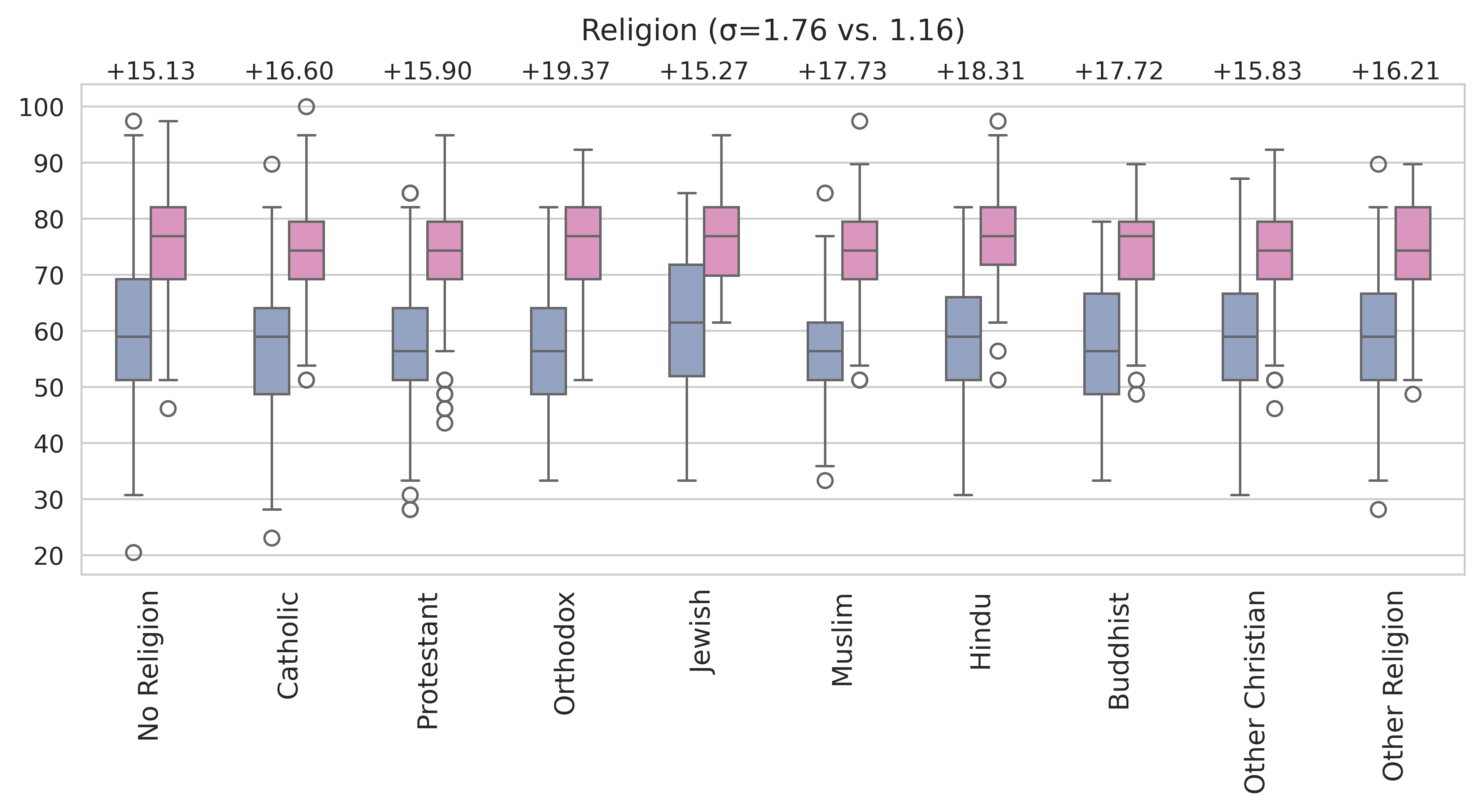}
    \captionof{figure}{The breakdown of the relative performance improvement of \finetuneshort compared to zero-short \llamathreeone for each demographics category within the \textit{Religion} dimension.}
    \label{fig:evenness_religion}
\end{figure*}

\begin{figure*}[ht]
    \vspace{-0.4cm}
    \centering
    \includegraphics[width=0.26\textwidth]{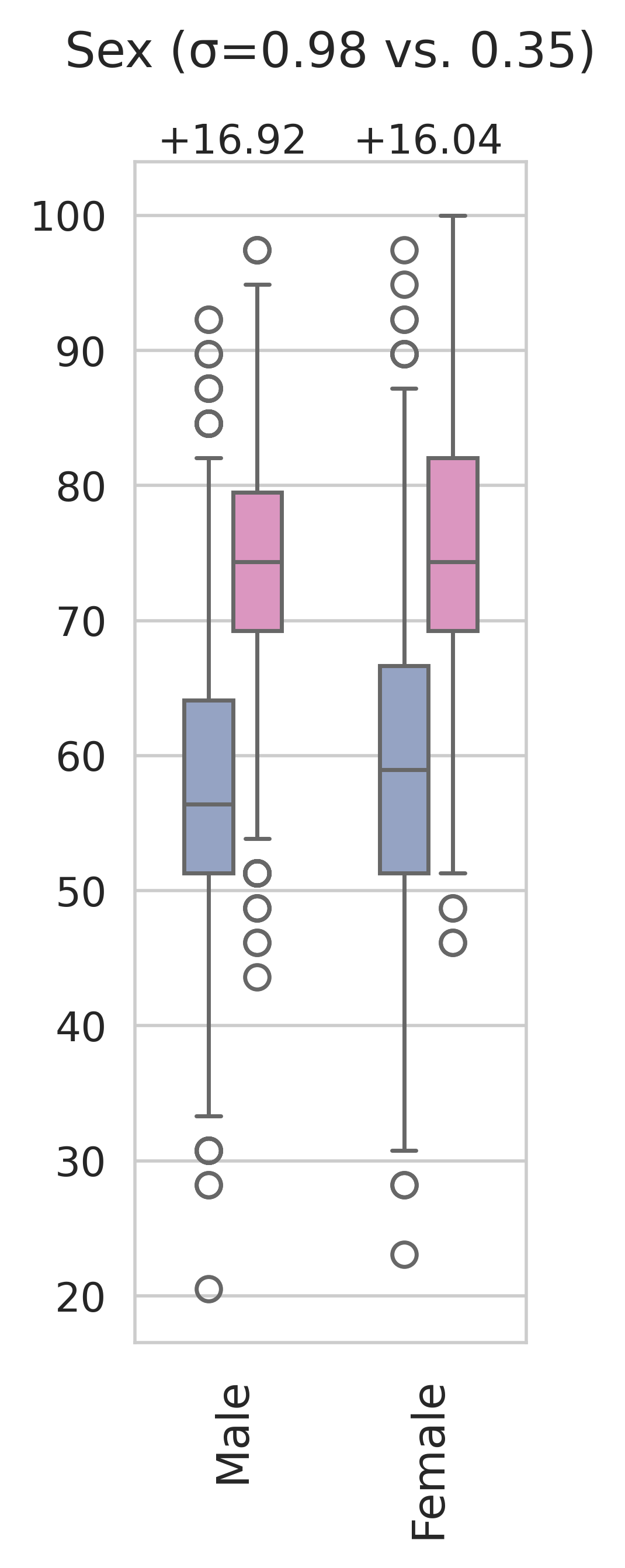}
    \captionof{figure}{The breakdown of the relative performance improvement of \finetuneshort compared to zero-short \llamathreeone for each demographics category within the \textit{Sex} dimension.}
    \label{fig:evenness_sex}
\end{figure*}

%% file: tables/resemble_cluster.tex
\begin{table*}[t!]
\centering

\begin{tabular}{r | rrrr|rrrr|r}
\toprule
\multicolumn{1}{c}{} & \multicolumn{4}{c}{\textbf{Polar}}                                                              & \multicolumn{4}{c}{\textbf{Refined}}                                                            & \multicolumn{1}{c}{\textbf{Overall}} \\

\midrule

\multicolumn{1}{r}{\textbf{N}} & \multicolumn{1}{r}{Probe 1} & \multicolumn{1}{r}{Probe 2} & \multicolumn{1}{r}{Probe 3} & \multicolumn{1}{r}{Avg} & \multicolumn{1}{r}{Probe 1} & \multicolumn{1}{r}{Probe 2} & \multicolumn{1}{r}{Probe 3} & \multicolumn{1}{r}{Avg} & \multicolumn{1}{r}{Avg}              \\
\midrule

1                    & 70.30              & 70.09              & 66.76              & 69.05                & 53.25              & 54.77              & 51.84              & 53.29                & 61.17                             \\
2                    & 70.54              & 70.92              & 66.56              & 69.34                & 52.38              & 55.48              & 50.98              & 52.94                & 61.14                             \\
3                    & 72.78              & 73.06              & 69.37              & 71.74                & 55.43              & 57.26              & 54.28              & 55.66                & 63.70                             \\
4                    & 72.90              & 73.23              & 69.23              & 71.79                & 56.30              & 58.15              & 55.13              & 56.53                & 64.16                             \\
5                    & 73.63              & 74.07              & 70.47              & 72.72                & 57.36              & 58.81              & 55.98              & 57.38                & 65.05                             \\
6                    & 73.86              & 74.11              & 70.45              & 72.81                & 57.27              & 58.90              & 56.45              & 57.54                & 65.17                             \\
7                    & 74.25              & 74.74              & 70.95              & 73.31                & 57.87              & 59.45              & 56.75              & 58.02                & 65.67                             \\
8                    & 74.18              & 74.59              & 70.78              & 73.19                & 58.27              & 59.78              & 57.13              & 58.39                & 65.79                             \\
9                    & 74.47              & 74.82              & 71.16              & 73.48                & 58.33              & 59.87              & 57.24              & 58.48                & 65.98                             \\
10                   & 74.43              & 74.72              & 71.20              & 73.45                & 58.22              & 60.24              & 57.62              & 58.69                & 66.07                             \\
11                   & 74.46              & 74.86              & 71.27              & 73.53                & 58.51              & 60.33              & 57.59              & 58.81                & 66.17                             \\
12                   & 74.50              & 74.82              & 71.05              & 73.46                & 58.73              & 60.35              & 57.81              & 58.96                & 66.21                             \\
13                   & 74.51              & 74.86              & 71.35              & 73.57                & 58.74              & 60.58              & 58.00              & 59.11                & 66.34                             \\
14                   & 74.37              & 74.84              & 71.33              & 73.51                & 58.96              & 60.60              & 57.95              & 59.17                & 66.34                             \\
15                   & 74.48              & 74.76              & 71.47              & 73.57                & 58.92              & 60.41              & 57.95              & 59.09                & 66.33                             \\
16                   & 74.37              & 74.81              & 71.35              & 73.51                & 59.03              & 60.63              & 57.93              & 59.19                & 66.35                             \\
17                   & 74.54              & 74.80              & 71.66              & 73.67                & 59.10              & 60.53              & 57.94              & 59.19                & 66.43                             \\
18                   & 74.57              & 74.72              & 71.50              & 73.60                & 59.08              & 60.80              & 58.14              & 59.34                & 66.47                             \\
19                   & 74.67              & 74.90              & 71.62              & 73.73                & 59.19              & 60.64              & 58.20              & 59.34                & 66.53                             \\
20                   & 74.62              & 74.82              & 71.56              & 73.67                & 59.28              & 60.71              & 58.23              & 59.41                & 66.54                             \\
21                   & 74.62              & 74.94              & 71.62              & 73.72                & 59.32              & 60.65              & 58.31              & 59.43                & 66.58                             \\
22                   & 74.71              & 74.85              & 71.53              & 73.70                & 59.24              & 60.74              & 58.35              & 59.44                & 66.57                             \\
23                   & 74.68              & 74.92              & 71.60              & 73.73                & 59.30              & 60.67              & 58.22              & 59.40                & 66.56                             \\
\rowcolor{cyan} 24                   & 74.74     & 74.87    & 71.60   & 73.73                & 59.32& 60.78    & 58.32    & 59.47                & 66.60                             \\
25                   & 74.73              & 75.00              & 71.72              & 73.81                & 59.17              & 60.67              & 58.33              & 59.39                & 66.60                             \\
26                   & 74.73              & 74.83              & 71.70              & 73.76                & 58.95              & 60.74              & 58.16              & 59.28                & 66.52                             \\
27                   & 74.78              & 74.98              & 71.78              & 73.85                & 59.04              & 60.72              & 58.14              & 59.30                & 66.57                             \\
28                   & 74.67              & 74.96              & 71.69              & 73.77                & 59.08              & 60.69              & 58.09              & 59.29                & 66.53                             \\
29                   & 74.74              & 74.98              & 71.74              & 73.82                & 59.10              & 60.79              & 58.04              & 59.31                & 66.57                             \\
30                   & 74.56              & 74.94              & 71.59              & 73.70                & 59.18              & 60.76              & 58.04              & 59.33                & 66.51                             \\
31                   & 74.60              & 75.04              & 71.67              & 73.77                & 59.16              & 60.73              & 58.10              & 59.33                & 66.55                             \\
32                   & 74.57              & 75.00              & 71.52              & 73.70                & 59.19              & 60.78              & 58.04              & 59.33                & 66.52                             \\
33                   & 74.56              & 75.00              & 71.69              & 73.75                & 59.23              & 60.67              & 58.04              & 59.32                & 66.53                             \\
34                   & 74.64              & 74.90              & 71.68              & 73.74                & 59.07              & 60.64              & 57.98              & 59.23                & 66.49                             \\
35                   & 74.74              & 74.92              & 71.67              & 73.78                & 59.17              & 60.55              & 57.97              & 59.23                & 66.50      \\                      

\bottomrule

\end{tabular}
\caption{Scores with different cluster size $N$ for the \texttt{[Resemble (top cluster)]} baseline.}
\label{tab:top_cluster_breakdown}
\end{table*}

%% file: acl_latex.bbl
\begin{thebibliography}{60}
\providecommand{\natexlab}[1]{#1}

\bibitem[{Bai et~al.(2022)Bai, Jones, Ndousse, Askell, Chen, DasSarma, Drain, Fort, Ganguli, Henighan, Joseph, Kadavath, Kernion, Conerly, El-Showk, Elhage, Hatfield-Dodds, Hernandez, Hume, Johnston, Kravec, Lovitt, Nanda, Olsson, Amodei, Brown, Clark, McCandlish, Olah, Mann, and Kaplan}]{bai2022hhrlhf}
Yuntao Bai, Andy Jones, Kamal Ndousse, Amanda Askell, Anna Chen, Nova DasSarma, Dawn Drain, Stanislav Fort, Deep Ganguli, Tom Henighan, Nicholas Joseph, Saurav Kadavath, Jackson Kernion, Tom Conerly, Sheer El-Showk, Nelson Elhage, Zac Hatfield-Dodds, Danny Hernandez, Tristan Hume, Scott Johnston, Shauna Kravec, Liane Lovitt, Neel Nanda, Catherine Olsson, Dario Amodei, Tom Brown, Jack Clark, Sam McCandlish, Chris Olah, Ben Mann, and Jared Kaplan. 2022.
\newblock \href {https://arxiv.org/abs/2204.05862} {Training a helpful and harmless assistant with reinforcement learning from human feedback}.
\newblock \emph{Preprint}, arXiv:2204.05862.

\bibitem[{Castricato et~al.(2024)Castricato, Lile, Rafailov, Fränken, and Finn}]{PERSONA2024}
Louis Castricato, Nathan Lile, Rafael Rafailov, Jan-Philipp Fränken, and Chelsea Finn. 2024.
\newblock \href {https://arxiv.org/abs/2407.17387} {Persona: A reproducible testbed for pluralistic alignment}.
\newblock \emph{Preprint}, arXiv:2407.17387.

\bibitem[{Chen et~al.(2024{\natexlab{a}})Chen, Chen, Rege, and Vinayak}]{chen2024palpluralisticalignmentframework}
Daiwei Chen, Yi~Chen, Aniket Rege, and Ramya~Korlakai Vinayak. 2024{\natexlab{a}}.
\newblock \href {https://arxiv.org/abs/2406.08469} {Pal: Pluralistic alignment framework for learning from heterogeneous preferences}.
\newblock \emph{Preprint}, arXiv:2406.08469.

\bibitem[{Chen et~al.(2024{\natexlab{b}})Chen, Saha, and Bansal}]{chen2024reconcile}
Justin Chih-Yao Chen, Swarnadeep Saha, and Mohit Bansal. 2024{\natexlab{b}}.
\newblock \href {https://arxiv.org/abs/2309.13007} {Reconcile: Round-table conference improves reasoning via consensus among diverse llms}.
\newblock \emph{Preprint}, arXiv:2309.13007.

\bibitem[{Chiu et~al.(2024{\natexlab{a}})Chiu, Jiang, Antoniak, Park, Li, Bhatia, Ravi, Tsvetkov, Shwartz, and Choi}]{chiu2024culturalteaming}
Yu~Ying Chiu, Liwei Jiang, Maria Antoniak, Chan~Young Park, Shuyue~Stella Li, Mehar Bhatia, Sahithya Ravi, Yulia Tsvetkov, Vered Shwartz, and Yejin Choi. 2024{\natexlab{a}}.
\newblock \href {https://arxiv.org/abs/2404.06664} {Culturalteaming: Ai-assisted interactive red-teaming for challenging llms' (lack of) multicultural knowledge}.
\newblock \emph{Preprint}, arXiv:2404.06664.

\bibitem[{Chiu et~al.(2024{\natexlab{b}})Chiu, Jiang, Lin, Park, Li, Ravi, Bhatia, Antoniak, Tsvetkov, Shwartz, and Choi}]{chiu2024culturalbench}
Yu~Ying Chiu, Liwei Jiang, Bill~Yuchen Lin, Chan~Young Park, Shuyue~Stella Li, Sahithya Ravi, Mehar Bhatia, Maria Antoniak, Yulia Tsvetkov, Vered Shwartz, and Yejin Choi. 2024{\natexlab{b}}.
\newblock \href {https://arxiv.org/abs/2410.02677} {Culturalbench: a robust, diverse and challenging benchmark on measuring the (lack of) cultural knowledge of llms}.
\newblock \emph{Preprint}, arXiv:2410.02677.

\bibitem[{Curry et~al.(2019{\natexlab{a}})Curry, Chesters, and Van~Lissa}]{curry2019mapping}
Oliver~Scott Curry, Matthew~Jones Chesters, and Caspar~J Van~Lissa. 2019{\natexlab{a}}.
\newblock Mapping morality with a compass: Testing the theory of ‘morality-as-cooperation’with a new questionnaire.
\newblock \emph{Journal of Research in Personality}, 78:106--124.

\bibitem[{Curry et~al.(2019{\natexlab{b}})Curry, Mullins, and Whitehouse}]{curry2019good}
Oliver~Scott Curry, Daniel~Austin Mullins, and Harvey Whitehouse. 2019{\natexlab{b}}.
\newblock Is it good to cooperate? testing the theory of morality-as-cooperation in 60 societies.
\newblock \emph{Current anthropology}, 60(1):47--69.

\bibitem[{Durmus et~al.(2024)Durmus, Nguyen, Liao, Schiefer, Askell, Bakhtin, Chen, Hatfield-Dodds, Hernandez, Joseph, Lovitt, McCandlish, Sikder, Tamkin, Thamkul, Kaplan, Clark, and Ganguli}]{GlobalOpinionQA2024}
Esin Durmus, Karina Nguyen, Thomas~I. Liao, Nicholas Schiefer, Amanda Askell, Anton Bakhtin, Carol Chen, Zac Hatfield-Dodds, Danny Hernandez, Nicholas Joseph, Liane Lovitt, Sam McCandlish, Orowa Sikder, Alex Tamkin, Janel Thamkul, Jared Kaplan, Jack Clark, and Deep Ganguli. 2024.
\newblock \href {https://arxiv.org/abs/2306.16388} {Towards measuring the representation of subjective global opinions in language models}.
\newblock \emph{Preprint}, arXiv:2306.16388.

\bibitem[{Feng et~al.(2024)Feng, Sorensen, Liu, Fisher, Park, Choi, and Tsvetkov}]{modularpluralism2024}
Shangbin Feng, Taylor Sorensen, Yuhan Liu, Jillian Fisher, Chan~Young Park, Yejin Choi, and Yulia Tsvetkov. 2024.
\newblock \href {https://arxiv.org/abs/2406.15951} {Modular pluralism: Pluralistic alignment via multi-llm collaboration}.
\newblock \emph{Preprint}, arXiv:2406.15951.

\bibitem[{Fung et~al.(2024)Fung, Zhao, Doo, Sun, and Ji}]{Fung2024MassivelyMK}
Yi~Ren Fung, Ruining Zhao, Jae Doo, Chenkai Sun, and Heng Ji. 2024.
\newblock \href {https://api.semanticscholar.org/CorpusID:267657749} {Massively multi-cultural knowledge acquisition \& lm benchmarking}.
\newblock \emph{ArXiv}, abs/2402.09369.

\bibitem[{Haerpfer et~al.(2020{\natexlab{a}})Haerpfer, Inglehart, Moreno, Welzel, Kizilova, Diez-Medrano, Lagos, Norris, Ponarin, and Puranen}]{wvswave72020}
Christian Haerpfer, Ronald Inglehart, Alejandro Moreno, Christian Welzel, Kseniya Kizilova, José Diez-Medrano, Marta Lagos, Pippa Norris, Eduard Ponarin, and Björn Puranen, editors. 2020{\natexlab{a}}.
\newblock \href {https://doi.org/10.14281/18241.1} {\emph{World Values Survey: Round Seven – Country-Pooled Datafile}}.
\newblock JD Systems Institute and WVSA Secretariat, Madrid, Spain and Vienna, Austria.
\newblock World Values Survey: Round Seven.

\bibitem[{Haerpfer et~al.(2020{\natexlab{b}})Haerpfer, Inglehart, Moreno, Welzel, Kizilova, Diez-Medrano, Lagos, Norris, Ponarin, and Puranen}]{worldvaluesurveywave7}
Christian Haerpfer, Ronald Inglehart, Alejandro Moreno, Christian Welzel, Kseniya Kizilova, Juan Diez-Medrano, Marta Lagos, Pippa Norris, Eduard Ponarin, and Björn Puranen, editors. 2020{\natexlab{b}}.
\newblock \href {https://doi.org/10.14281/18241.1} {\emph{World Values Survey: Round Seven -- Country-Pooled Datafile}}.
\newblock JD Systems Institute \& WVSA Secretariat, Madrid, Spain \& Vienna, Austria.

\bibitem[{Han et~al.(2024)Han, Shenfeld, Srivastava, Kim, and Agrawal}]{han2024valueaugmentedsamplinglanguage}
Seungwook Han, Idan Shenfeld, Akash Srivastava, Yoon Kim, and Pulkit Agrawal. 2024.
\newblock \href {https://arxiv.org/abs/2405.06639} {Value augmented sampling for language model alignment and personalization}.
\newblock \emph{Preprint}, arXiv:2405.06639.

\bibitem[{Hsieh and Andersson(2021)}]{sep-value-incommensurable}
Nien-hê Hsieh and Henrik Andersson. 2021.
\newblock {Incommensurable Values}.
\newblock In Edward~N. Zalta, editor, \emph{The {Stanford} Encyclopedia of Philosophy}, {F}all 2021 edition. Metaphysics Research Lab, Stanford University.

\bibitem[{Jang et~al.(2023)Jang, Kim, Lin, Wang, Hessel, Zettlemoyer, Hajishirzi, Choi, and Ammanabrolu}]{personalizedsoup2023}
Joel Jang, Seungone Kim, Bill~Yuchen Lin, Yizhong Wang, Jack Hessel, Luke Zettlemoyer, Hannaneh Hajishirzi, Yejin Choi, and Prithviraj Ammanabrolu. 2023.
\newblock \href {https://arxiv.org/abs/2310.11564} {Personalized soups: Personalized large language model alignment via post-hoc parameter merging}.
\newblock \emph{Preprint}, arXiv:2310.11564.

\bibitem[{Ji et~al.(2024)Ji, He, and Gu}]{ji2024activequeries}
Kaixuan Ji, Jiafan He, and Quanquan Gu. 2024.
\newblock \href {https://arxiv.org/abs/2402.09401} {Reinforcement learning from human feedback with active queries}.
\newblock \emph{Preprint}, arXiv:2402.09401.

\bibitem[{Jiang et~al.(2023)Jiang, Xu, Zhu, Han, Zhang, and Zhu}]{jiang2023evaluating}
Guangyuan Jiang, Manjie Xu, Song-Chun Zhu, Wenjuan Han, Chi Zhang, and Yixin Zhu. 2023.
\newblock \href {https://openreview.net/forum?id=I9xE1Jsjfx} {Evaluating and inducing personality in pre-trained language models}.
\newblock In \emph{Thirty-seventh Conference on Neural Information Processing Systems}.

\bibitem[{Kalimeri et~al.(2019)Kalimeri, Beiró, Delfino, Raleigh, and Cattuto}]{KALIMERI2019428}
Kyriaki Kalimeri, Mariano~G. Beiró, Matteo Delfino, Robert Raleigh, and Ciro Cattuto. 2019.
\newblock \href {https://doi.org/10.1016/j.chb.2018.11.024} {Predicting demographics, moral foundations, and human values from digital behaviours}.
\newblock \emph{Computers in Human Behavior}, 92:428--445.

\bibitem[{Keswani et~al.(2024)Keswani, Conitzer, Heidari, Borg, and Sinnott-Armstrong}]{keswani2024prosconsactivelearning}
Vijay Keswani, Vincent Conitzer, Hoda Heidari, Jana~Schaich Borg, and Walter Sinnott-Armstrong. 2024.
\newblock \href {https://arxiv.org/abs/2407.18889} {On the pros and cons of active learning for moral preference elicitation}.
\newblock \emph{Preprint}, arXiv:2407.18889.

\bibitem[{Kirk et~al.(2024{\natexlab{a}})Kirk, Whitefield, Röttger, Bean, Margatina, Ciro, Mosquera, Bartolo, Williams, He, Vidgen, and Hale}]{kirk2024prismalignment}
Hannah~Rose Kirk, Alexander Whitefield, Paul Röttger, Andrew Bean, Katerina Margatina, Juan Ciro, Rafael Mosquera, Max Bartolo, Adina Williams, He~He, Bertie Vidgen, and Scott~A. Hale. 2024{\natexlab{a}}.
\newblock \href {https://arxiv.org/abs/2404.16019} {The prism alignment project: What participatory, representative and individualised human feedback reveals about the subjective and multicultural alignment of large language models}.
\newblock \emph{Preprint}, arXiv:2404.16019.

\bibitem[{Kirk et~al.(2024{\natexlab{b}})Kirk, Vidgen, R{\"o}ttger et~al.}]{kirk2024personalization}
H.R. Kirk, B.~Vidgen, P.~R{\"o}ttger, et~al. 2024{\natexlab{b}}.
\newblock \href {https://doi.org/10.1038/s42256-024-00820-y} {The benefits, risks and bounds of personalizing the alignment of large language models to individuals}.
\newblock \emph{Nature Machine Intelligence}, 6:383--392.

\bibitem[{Kwok et~al.(2024)Kwok, Bravansky, and Griffin}]{kwok2024syntheticpersonas}
Louis Kwok, Michal Bravansky, and Lewis Griffin. 2024.
\newblock \href {https://openreview.net/forum?id=S4ZOkV1AHl} {Evaluating cultural adaptability of a large language model via simulation of synthetic personas}.
\newblock In \emph{First Conference on Language Modeling}.

\bibitem[{Lake et~al.(2024)Lake, Choi, and Durrett}]{lake2024distributionalovertonpluralisminvestigating}
Thom Lake, Eunsol Choi, and Greg Durrett. 2024.
\newblock \href {https://arxiv.org/abs/2406.17692} {From distributional to overton pluralism: Investigating large language model alignment}.
\newblock \emph{Preprint}, arXiv:2406.17692.

\bibitem[{Lee et~al.(2024)Lee, Park, Kim, and Seo}]{lee2024aligningthousandspreferencesmessage}
Seongyun Lee, Sue~Hyun Park, Seungone Kim, and Minjoon Seo. 2024.
\newblock \href {https://arxiv.org/abs/2405.17977} {Aligning to thousands of preferences via system message generalization}.
\newblock \emph{Preprint}, arXiv:2405.17977.

\bibitem[{Li et~al.(2024{\natexlab{a}})Li, Chen, Wang, Sitaram, and Xie}]{Li2024CultureLLMIC}
Cheng Li, Mengzhou Chen, Jindong Wang, Sunayana Sitaram, and Xing Xie. 2024{\natexlab{a}}.
\newblock \href {https://api.semanticscholar.org/CorpusID:267750997} {Culturellm: Incorporating cultural differences into large language models}.
\newblock \emph{ArXiv}, abs/2402.10946.

\bibitem[{Li et~al.(2024{\natexlab{b}})Li, Peris, Mehrabi, Goyal, Chang, Galstyan, Zemel, and Gupta}]{li-etal-2024-steerability}
Junyi Li, Charith Peris, Ninareh Mehrabi, Palash Goyal, Kai-Wei Chang, Aram Galstyan, Richard Zemel, and Rahul Gupta. 2024{\natexlab{b}}.
\newblock \href {https://doi.org/10.18653/v1/2024.naacl-long.405} {The steerability of large language models toward data-driven personas}.
\newblock In \emph{Proceedings of the 2024 Conference of the North American Chapter of the Association for Computational Linguistics: Human Language Technologies (Volume 1: Long Papers)}, pages 7290--7305, Mexico City, Mexico. Association for Computational Linguistics.

\bibitem[{Liu et~al.(2024)Liu, Zhu, Wang, Wei, Min, Lu, Wang, Yin, and Dou}]{liu2024llmspersonaplug}
Jiongnan Liu, Yutao Zhu, Shuting Wang, Xiaochi Wei, Erxue Min, Yu~Lu, Shuaiqiang Wang, Dawei Yin, and Zhicheng Dou. 2024.
\newblock \href {https://arxiv.org/abs/2409.11901} {Llms + persona-plug = personalized llms}.
\newblock \emph{Preprint}, arXiv:2409.11901.

\bibitem[{Maio(2010)}]{MAIO20101}
Gregory~R. Maio. 2010.
\newblock \href {https://doi.org/10.1016/S0065-2601(10)42001-8} {Chapter 1 - mental representations of social values}.
\newblock In \emph{Advances in Experimental Social Psychology}, volume~42 of \emph{Advances in Experimental Social Psychology}, pages 1--43. Academic Press.

\bibitem[{Mason(2023)}]{sep-value-pluralism}
Elinor Mason. 2023.
\newblock {Value Pluralism}.
\newblock In Edward~N. Zalta and Uri Nodelman, editors, \emph{The {Stanford} Encyclopedia of Philosophy}, {S}ummer 2023 edition. Metaphysics Research Lab, Stanford University.

\bibitem[{Mehta et~al.(2023)Mehta, Das, Neopane, Dai, Bogunovic, Schneider, and Neiswanger}]{mehta2023sampleefficientreinforcementlearning}
Viraj Mehta, Vikramjeet Das, Ojash Neopane, Yijia Dai, Ilija Bogunovic, Jeff Schneider, and Willie Neiswanger. 2023.
\newblock \href {https://arxiv.org/abs/2312.00267} {Sample efficient reinforcement learning from human feedback via active exploration}.
\newblock \emph{Preprint}, arXiv:2312.00267.

\bibitem[{Moon et~al.(2024)Moon, Abdulhai, Kang, Suh, Soedarmadji, Behar, and Chan}]{moon2024personabackstory}
Suhong Moon, Marwa Abdulhai, Minwoo Kang, Joseph Suh, Widyadewi Soedarmadji, Eran~Kohen Behar, and David~M. Chan. 2024.
\newblock \href {https://arxiv.org/abs/2407.06576} {Virtual personas for language models via an anthology of backstories}.
\newblock \emph{Preprint}, arXiv:2407.06576.

\bibitem[{Muldrew et~al.(2024)Muldrew, Hayes, Zhang, and Barber}]{muldrew2024activepreferencelearning}
William Muldrew, Peter Hayes, Mingtian Zhang, and David Barber. 2024.
\newblock \href {https://arxiv.org/abs/2402.08114} {Active preference learning for large language models}.
\newblock \emph{Preprint}, arXiv:2402.08114.

\bibitem[{Mysore et~al.(2023)Mysore, Lu, Wan, Yang, Menezes, Baghaee, Gonzalez, Neville, and Safavi}]{mysore2023pearlpersonalizinglargelanguage}
Sheshera Mysore, Zhuoran Lu, Mengting Wan, Longqi Yang, Steve Menezes, Tina Baghaee, Emmanuel~Barajas Gonzalez, Jennifer Neville, and Tara Safavi. 2023.
\newblock \href {https://arxiv.org/abs/2311.09180} {Pearl: Personalizing large language model writing assistants with generation-calibrated retrievers}.
\newblock \emph{Preprint}, arXiv:2311.09180.

\bibitem[{Myung et~al.(2024)Myung, Lee, Zhou, Jin, Putri, Antypas, Borkakoty, Kim, P{\'e}rez-Almendros, Ayele, Guti'errez-Basulto, Ib'anez-Garc'ia, Lee, Muhammad, Park, Rzayev, White, Yimam, Pilehvar, Ousidhoum, Camacho-Collados, and Oh}]{Myung2024BLEnDAB}
Jun-Hee Myung, Nayeon Lee, Yi~Zhou, Jiho Jin, Rifki~Afina Putri, Dimosthenis Antypas, Hsuvas Borkakoty, Eunsu Kim, Carla P{\'e}rez-Almendros, Abinew~Ali Ayele, V'ictor Guti'errez-Basulto, Yazm'in Ib'anez-Garc'ia, Hwaran Lee, Shamsuddeen~Hassan Muhammad, Kiwoong Park, Anar Rzayev, Nina White, Seid~Muhie Yimam, Mohammad~Taher Pilehvar, Nedjma~Djouhra Ousidhoum, Jos{\'e} Camacho-Collados, and Alice Oh. 2024.
\newblock \href {https://api.semanticscholar.org/CorpusID:270521296} {Blend: A benchmark for llms on everyday knowledge in diverse cultures and languages}.
\newblock \emph{ArXiv}, abs/2406.09948.

\bibitem[{Ouyang et~al.(2022)Ouyang, Wu, Jiang, Almeida, Wainwright, Mishkin, Zhang, Agarwal, Slama, Ray, Schulman, Hilton, Kelton, Miller, Simens, Askell, Welinder, Christiano, Leike, and Lowe}]{ouyang2022rlhf}
Long Ouyang, Jeff Wu, Xu~Jiang, Diogo Almeida, Carroll~L. Wainwright, Pamela Mishkin, Chong Zhang, Sandhini Agarwal, Katarina Slama, Alex Ray, John Schulman, Jacob Hilton, Fraser Kelton, Luke Miller, Maddie Simens, Amanda Askell, Peter Welinder, Paul Christiano, Jan Leike, and Ryan Lowe. 2022.
\newblock \href {https://arxiv.org/abs/2203.02155} {Training language models to follow instructions with human feedback}.
\newblock \emph{Preprint}, arXiv:2203.02155.

\bibitem[{Park et~al.(2022)Park, Popowski, Cai, Morris, Liang, and Bernstein}]{social_simulacra}
Joon~Sung Park, Lindsay Popowski, Carrie Cai, Meredith~Ringel Morris, Percy Liang, and Michael~S. Bernstein. 2022.
\newblock \href {https://doi.org/10.1145/3526113.3545616} {Social simulacra: Creating populated prototypes for social computing systems}.
\newblock In \emph{Proceedings of the 35th Annual ACM Symposium on User Interface Software and Technology}, UIST '22, New York, NY, USA. Association for Computing Machinery.

\bibitem[{{Pew Research Center}(n.d.)}]{pew_research_center}
{Pew Research Center}. n.d.
\newblock Pew research center.
\newblock \url{https://www.pewresearch.org}.
\newblock Accessed: 2024-09-30.

\bibitem[{Piriyakulkij et~al.(2024)Piriyakulkij, Kuleshov, and Ellis}]{piriyakulkij2024activepreference}
Wasu~Top Piriyakulkij, Volodymyr Kuleshov, and Kevin Ellis. 2024.
\newblock \href {https://arxiv.org/abs/2312.12009} {Active preference inference using language models and probabilistic reasoning}.
\newblock \emph{Preprint}, arXiv:2312.12009.

\bibitem[{Poddar et~al.(2024)Poddar, Wan, Ivison, Gupta, and Jaques}]{poddar2024personalizing}
Sriyash Poddar, Yanming Wan, Hamish Ivison, Abhishek Gupta, and Natasha Jaques. 2024.
\newblock \href {https://arxiv.org/abs/2408.10075} {Personalizing reinforcement learning from human feedback with variational preference learning}.
\newblock \emph{Preprint}, arXiv:2408.10075.

\bibitem[{Rafailov et~al.(2024)Rafailov, Sharma, Mitchell, Ermon, Manning, and Finn}]{rafailov2024dpo}
Rafael Rafailov, Archit Sharma, Eric Mitchell, Stefano Ermon, Christopher~D. Manning, and Chelsea Finn. 2024.
\newblock \href {https://arxiv.org/abs/2305.18290} {Direct preference optimization: Your language model is secretly a reward model}.
\newblock \emph{Preprint}, arXiv:2305.18290.

\bibitem[{Rao et~al.(2024)Rao, Yerukola, Shah, Reinecke, and Sap}]{Rao2024NORMADAB}
Abhinav Rao, Akhila Yerukola, Vishwa Shah, Katharina Reinecke, and Maarten Sap. 2024.
\newblock \href {https://api.semanticscholar.org/CorpusID:269282746} {Normad: A benchmark for measuring the cultural adaptability of large language models}.
\newblock \emph{ArXiv}, abs/2404.12464.

\bibitem[{Ryan et~al.(2024)Ryan, Held, and Yang}]{ryan-etal-2024-unintended}
Michael Ryan, William Held, and Diyi Yang. 2024.
\newblock \href {https://doi.org/10.18653/v1/2024.acl-long.853} {Unintended impacts of {LLM} alignment on global representation}.
\newblock In \emph{Proceedings of the 62nd Annual Meeting of the Association for Computational Linguistics (Volume 1: Long Papers)}, pages 16121--16140, Bangkok, Thailand. Association for Computational Linguistics.

\bibitem[{Santurkar et~al.(2023)Santurkar, Durmus, Ladhak, Lee, Liang, and Hashimoto}]{lmopinion2023}
Shibani Santurkar, Esin Durmus, Faisal Ladhak, Cinoo Lee, Percy Liang, and Tatsunori Hashimoto. 2023.
\newblock Whose opinions do language models reflect?
\newblock In \emph{Proceedings of the 40th International Conference on Machine Learning}, ICML'23. JMLR.org.

\bibitem[{Schaffer(2018)}]{sep-monism}
Jonathan Schaffer. 2018.
\newblock {Monism}.
\newblock In Edward~N. Zalta, editor, \emph{The {Stanford} Encyclopedia of Philosophy}, {W}inter 2018 edition. Metaphysics Research Lab, Stanford University.

\bibitem[{Schulman et~al.(2017)Schulman, Wolski, Dhariwal, Radford, and Klimov}]{schulman2017ppo}
John Schulman, Filip Wolski, Prafulla Dhariwal, Alec Radford, and Oleg Klimov. 2017.
\newblock \href {https://arxiv.org/abs/1707.06347} {Proximal policy optimization algorithms}.
\newblock \emph{Preprint}, arXiv:1707.06347.

\bibitem[{Schwartz(2012)}]{schwartz2012overview}
Shalom~H. Schwartz. 2012.
\newblock \href {https://doi.org/10.9707/2307-0919.1116} {An overview of the schwartz theory of basic values}.
\newblock \emph{Online Readings in Psychology and Culture}, 2(1).

\bibitem[{Serapio-García et~al.(2023)Serapio-García, Safdari, Crepy, Sun, Fitz, Romero, Abdulhai, Faust, and Matarić}]{serapiogarcía2023personalitytraits}
Greg Serapio-García, Mustafa Safdari, Clément Crepy, Luning Sun, Stephen Fitz, Peter Romero, Marwa Abdulhai, Aleksandra Faust, and Maja Matarić. 2023.
\newblock \href {https://arxiv.org/abs/2307.00184} {Personality traits in large language models}.
\newblock \emph{Preprint}, arXiv:2307.00184.

\bibitem[{Shi et~al.(2024)Shi, Li, Zhang, Ziems, yu, Horesh, de~Paula, and Yang}]{shi2024culturebank}
Weiyan Shi, Ryan Li, Yutong Zhang, Caleb Ziems, Chunhua yu, Raya Horesh, Rogério~Abreu de~Paula, and Diyi Yang. 2024.
\newblock \href {https://arxiv.org/abs/2404.15238} {Culturebank: An online community-driven knowledge base towards culturally aware language technologies}.
\newblock \emph{Preprint}, arXiv:2404.15238.

\bibitem[{Sorensen et~al.(2024)Sorensen, Moore, Fisher, Gordon, Mireshghallah, Rytting, Ye, Jiang, Lu, Dziri, Althoff, and Choi}]{sorensen2024roadmappluralisticalignment}
Taylor Sorensen, Jared Moore, Jillian Fisher, Mitchell Gordon, Niloofar Mireshghallah, Christopher~Michael Rytting, Andre Ye, Liwei Jiang, Ximing Lu, Nouha Dziri, Tim Althoff, and Yejin Choi. 2024.
\newblock \href {https://arxiv.org/abs/2402.05070} {A roadmap to pluralistic alignment}.
\newblock \emph{Preprint}, arXiv:2402.05070.

\bibitem[{Stenner et~al.(2008)Stenner, Watts, and Worrell}]{1a35f96dcc55415ca791831dced159cd}
Paul Stenner, S.~Watts, and M.~Worrell. 2008.
\newblock \href {https://doi.org/10.4135/9781848607927} {\emph{Q Methodology}}, pages 215--239.
\newblock Sage Research Methods.

\bibitem[{Sun et~al.(2024)Sun, Yang, Reddy, Fung, Chan, Small, Zhai, and Ji}]{sun2024personadb}
Chenkai Sun, Ke~Yang, Revanth~Gangi Reddy, Yi~R. Fung, Hou~Pong Chan, Kevin Small, ChengXiang Zhai, and Heng Ji. 2024.
\newblock \href {https://arxiv.org/abs/2402.11060} {Persona-db: Efficient large language model personalization for response prediction with collaborative data refinement}.
\newblock \emph{Preprint}, arXiv:2402.11060.

\bibitem[{Verga et~al.(2024)Verga, Hofstatter, Althammer, Su, Piktus, Arkhangorodsky, Xu, White, and Lewis}]{verga2024replacingjudgesjuries}
Pat Verga, Sebastian Hofstatter, Sophia Althammer, Yixuan Su, Aleksandra Piktus, Arkady Arkhangorodsky, Minjie Xu, Naomi White, and Patrick Lewis. 2024.
\newblock \href {https://arxiv.org/abs/2404.18796} {Replacing judges with juries: Evaluating llm generations with a panel of diverse models}.
\newblock \emph{Preprint}, arXiv:2404.18796.

\bibitem[{Xu et~al.(2022)Xu, Szlam, and Weston}]{xu-etal-2022-beyond}
Jing Xu, Arthur Szlam, and Jason Weston. 2022.
\newblock \href {https://doi.org/10.18653/v1/2022.acl-long.356} {Beyond goldfish memory: Long-term open-domain conversation}.
\newblock In \emph{Proceedings of the 60th Annual Meeting of the Association for Computational Linguistics (Volume 1: Long Papers)}, pages 5180--5197, Dublin, Ireland. Association for Computational Linguistics.

\bibitem[{Zhang et~al.(2024)Zhang, Yu, Sharma, Yang, Wang, Hassan, and Wang}]{zhang2024selfexploring}
Shenao Zhang, Donghan Yu, Hiteshi Sharma, Ziyi Yang, Shuohang Wang, Hany Hassan, and Zhaoran Wang. 2024.
\newblock \href {https://arxiv.org/abs/2405.19332} {Self-exploring language models: Active preference elicitation for online alignment}.
\newblock \emph{Preprint}, arXiv:2405.19332.

\bibitem[{Zhao et~al.(2024)Zhao, Mondal, Tandon, Dillion, Gray, and Gu}]{worldvaluesbench2024}
Wenlong Zhao, Debanjan Mondal, Niket Tandon, Danica Dillion, Kurt Gray, and Yuling Gu. 2024.
\newblock \href {https://aclanthology.org/2024.lrec-main.1539} {{W}orld{V}alues{B}ench: A large-scale benchmark dataset for multi-cultural value awareness of language models}.
\newblock In \emph{Proceedings of the 2024 Joint International Conference on Computational Linguistics, Language Resources and Evaluation (LREC-COLING 2024)}, pages 17696--17706, Torino, Italia. ELRA and ICCL.

\bibitem[{Zhou et~al.(2024)Zhou, Zhu, Mathur, Zhang, Yu, Qi, Morency, Bisk, Fried, Neubig, and Sap}]{zhou2024sotopia}
Xuhui Zhou, Hao Zhu, Leena Mathur, Ruohong Zhang, Haofei Yu, Zhengyang Qi, Louis-Philippe Morency, Yonatan Bisk, Daniel Fried, Graham Neubig, and Maarten Sap. 2024.
\newblock \href {https://arxiv.org/abs/2310.11667} {Sotopia: Interactive evaluation for social intelligence in language agents}.
\newblock \emph{Preprint}, arXiv:2310.11667.

\bibitem[{Zhu et~al.(2024)Zhu, Yang, and Zhang}]{zhu2024personalityalignment}
Minjun Zhu, Linyi Yang, and Yue Zhang. 2024.
\newblock \href {https://arxiv.org/abs/2408.11779} {Personality alignment of large language models}.
\newblock \emph{Preprint}, arXiv:2408.11779.

\bibitem[{Ziems et~al.(2024)Ziems, Held, Shaikh, Chen, Zhang, and Yang}]{ziems-etal-2024-large}
Caleb Ziems, William Held, Omar Shaikh, Jiaao Chen, Zhehao Zhang, and Diyi Yang. 2024.
\newblock \href {https://doi.org/10.1162/coli_a_00502} {Can large language models transform computational social science?}
\newblock \emph{Computational Linguistics}, 50(1):237--291.

\bibitem[{Zimmerman and Bradley(2019)}]{sep-value-intrinsic-extrinsic}
Michael~J. Zimmerman and Ben Bradley. 2019.
\newblock {Intrinsic vs. Extrinsic Value}.
\newblock In Edward~N. Zalta, editor, \emph{The {Stanford} Encyclopedia of Philosophy}, {S}pring 2019 edition. Metaphysics Research Lab, Stanford University.

\end{thebibliography}
